# LEPOR: An Augmented Machine Translation Evaluation Metric

by

**Lifeng Han (Aaron)**

**Master of Science in Software Engineering**

**2014**

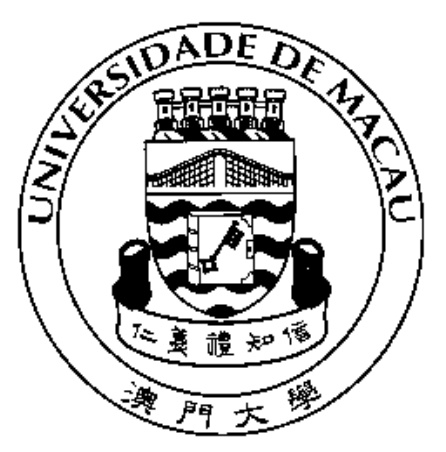

**Faculty of Science and Technology**
**University of Macau**

# LEPOR: AN AUGMENTED MACHINE TRANSLATION EVALUATION METRIC

by

LiFeng Han, Aaron

A thesis submitted in partial fulfillment of the
requirements for the degree of

Master of Science in Software Engineering

Faculty of Science and Technology
University of Macau

2014

Approved by:
Examination Committee
Chairperson : Prof. Zhiguo Gong, Dept. Head of CIS, FST
Supervisor : Dr. Fai Wong, Assistant Professor of CIS, FST
Co-supervisor: Dr. Sam Chao, Assistant Professor of CIS, FST'
Member : Prof. Chi Man Pun, Associate Professor of CIS, FST

Date: July 10th, 2014

In presenting this thesis in partial fulfillment of the requirements for a Master's degree at the University of Macau, I agree that the Library and the Faculty of Science and Technology shall make its copies freely available for inspection. However, reproduction of this thesis for any purposes or by any means shall not be allowed without my written permission. Authorization is sought by contacting the author at

Address:FST Building 2023, University of Macau, Macau S.A.R.

Telephone: +(00)853-63567608 (expired)

E-mail: hanlifengaaron@gmail.com / mb15488@umac.mo

Student no. M-B1-5488-7
Name: Lifeng Han

Signature ______

Date 2014.07.10th

University of Macau

Abstract

LEPOR: AN AUGMENTED MACHINE TRANSLATION EVALUATION METRIC

by LiFeng Han (Aaron)

Thesis Supervisors: Dr. Lidia S. Chao and Dr. Derek F. Wong
Master of Science in Software Engineering

Machine translation (MT) was developed as one of the hottest research topics in the natural language processing (NLP) literature. One important issue in MT is that how to evaluate the MT system reasonably and tell us whether the translation system makes an improvement or not. The traditional manual judgment methods are expensive, time-consuming, unrepeatable, and sometimes with low agreement. On the other hand, the popular automatic MT evaluation methods have some weaknesses. Firstly, they tend to perform well on the language pairs with English as the target language, but weak when English is used as source. Secondly, some methods rely on many additional linguistic features to achieve good performance, which makes the metric unable to replicate and apply to other language pairs easily. Thirdly, some popular metrics utilize incomprehensive factors, which result in low performance on some practical tasks.

In this thesis, to address the existing problems, we design novel MT evaluation methods and investigate their performances on different languages. Firstly, we design augmented factors to yield highly accurate evaluation. Secondly, we design a tunable evaluation model where weighting of factors can be optimized according to the characteristics of languages. Thirdly, in the enhanced version of our methods, we design concise linguistic feature using part-of-speech (POS) to show that our methods can yield even higher performance when using some external linguistic resources. Finally, we introduce the practical performance of our metrics in the ACL-WMT workshop shared tasks, which show that the proposed methods are robust across different languages. In addition, we also present some novel work on quality estimation of MT without using reference translations including the usage of probability models of Naïve Bayes (NB), support vector machine (SVM) classification algorithms, and a discriminative undirected graphical model conditional random field (CRF), in addition to feature engineering.

LISTOF FIGURES

# LIST OF ABBREVIATIONS

| | |
|---|---|
| ACL | Association for Computational Linguistics |
| HMM | Hidden Markov Models |
| ITG | Inverse Transducer Grammar |
| LM | Language Model |
| ML | Machine Learning |
| MT | Machine Translation |
| MTE | Machine Translation Evaluation |
| NP | Noun Phrase |
| NLP | Natural Language Processing |
| POS | Part-of-Speech |
| PP | Preposition Phrase |
| SIG | Special Interest Group |
| SMT | Statistical Machine Translation |
| VP | Verb Phrase |
| WMT | International Workshop of SMT |

## ACKNOWLEDGEMENTS

I would like to thank my thesis supervisors, Dr. Derek F. Wong and Dr. Lidia S. Chao, who gave me many valuable advices and suggestions during my research periods. They always have encouraged and supported me to go further. This work would not have been possible without their guidance.

I also give my thanks to my colleagues in our NLP$^{2}$CT-lab as following.

Thanks to Mr. Liang Tian & Mr. Xiaodong Zeng (Samuel), from whom I learnt a lot during the three years research in lab. Thanks to Mr. Liangye He (Yervant), Mr. Yi Lu, and Mr. Junwen Xing (Anson), who gave me a lot of technical and programming help in our lab during my research.

Thanks to Mr. Longyue Wang (Vincent) who inspired me a lot in the lab. He is a hard-working guy, and we worked in our lab all through the night for the paper submission before the deadline of some international conferences, and this kind of experiment occurred several times. It was a really tiring work such during all the nights we spent in the lab; however, it was also such a happy experience after we submitted the paper on time.

Thanks to our lab members Mr. Fan Sun (Jeff), Miss. Qiuping Huang (Michelle), Miss. Ling Zhu (Lynn), Mr. Shuo Li (Shawn), Mr. Hao Zong (Peter), Mr. Yiming Wang, Mr. Jiaji Zhou, Mr. Chunhui Lu (Olivio), Mr. Yuchu Lin, Miss. Ian, Mr. Ban, Mr. Kin, Miss. Wendy, and Mr. Jacky, et al. who accompanied my master stage and made the life in the lab wonderful.

Thanks a lot to Mr. Kevin Tang and Mr. Francisco Oliveira for their technical help in our lab. I must have made a lot of trouble to them when my computer went wrong in the lab. It is very kind of them to deal with the problems with a lot of patience.

I would like to give my thanks to many kind people outside of the NLP$^2$CT lab as follows.

My thanks go to Prof. Deng Ding from the Department of Mathematics, who helped me a lot in the national postgraduate mathematical modelling competition, and our team finally gained the national second prize.

Thanks to Mr. Vincent Li from the ICTO of UM, who gave me a lot of help including financial support before I joined the NLP$^2$CT lab and I worked in ICTO for three months under his guidance.

In addition, I want to give my thanks to my friends from other departments in UM. Thanks to Mr. Zhibo Wang, Miss. Shulin Lv, Mr. Jianhao Lv, and Mr. Zhi Li et al. from the Department of Mathematics; Mr. Gary Wong (Enlong Huang) from the Department of E-commerce; Mr. Hongqiang Zhu, Mr. Yunsheng Xie (Shawn), Miss. Liyu Deng, Miss. Hong Zhang, Miss. Selin et al. from the Department of English, Mr. Xingqiang Peng from the Department of Electrical and Computer Engineering (ECE), Mr. Ziqian Ma, Mr. Dong Li et al. from the Department of Electromechanical Engineering (EME), Mr. Shuang Lin, Miss. Le Dong et al. from the Department of Civil Engineering, Miss. Xiaonan Bai, Mr. Li Li, Mr. Xiaozhou Shen et al. from the FSH, Mr. Haipeng Lei, Mr. Hefeng Zhou, Mr. Haitao Li, Miss. Xiaohui Huang et al. from the Institute of Chinese Medical Sciences (ICMS). Thanks for their fulfilling the good memory of my master stage in UM.

My special thanks to my roommate, Mr. Wei Hong, for his kind tolerance and staying with me throughout the master stage. I receive many good suggestions from him.

Thanks to Mr. Gust (Augustine), my swimming coach, who teaches me to swim with kind patience. When I got tired in my lab, I went to the Olympic swimming pool in Taipa of Macau and he taught me how to float and relax in water. Swimming became almost my only hobby in the master stage, and kept my body in health.

Swimming will become a life-long hobby for me. He likestraveling and thinking and told me a lot of knowledge about the people all around the world.

Thanks to Mr. Tristan and Ms. Nadya from Germany, who kindly invited me to stay at their place during the international conference of GSCL, and they kindly cooked the delicious Germany-style breakfast and dinner for me during my stay. I enjoyed the stay at their place. They went to the station to pick me up when I reached Germany and sent me to the station when I left Germany. I also remember the two lovely pets Pjofur & Pora, the nice movie in the night before I left Darmstadt.

Furthermore, I would like to give my thanks to Prof. Xiuwen Xu from the Hebei Normal University (HNU) who cared me a lot in the past days. Thanks to Prof. Jianguo Lei, Prof. Lixia Liu, and Prof. Wenming Li from HNU, who kindly wrote the recommendation letters for me when I was applying for UM.

It will be my great honour if you read this thesis in one day. Please kindly forgive me if I lost your name here, because there are really so many people I should give my thanks to.

Finally, I give my deepest gratitude to my family. Thanks a lot to my parents for giving me a life such that I have the chance to explore the world. Thanks for their end-less loving and support to me. Thanks to my sister Miss. Litao Han (Grace), who encouraged me a lot. They are one most important motivation of me to work hard. There is still a long way in ahead of us to go together and I have confidence that it will be a *beautiful* long way.

# DEDICATION

I wish to dedicate this thesis to my parents and my sister.

# 1. INTRODUCTION

The machine translation (MT) began as early as in the 1950s (Weaver, 1955), and gained a quick development since the 1990s due to the development of computer technology, e.g. augmented storage capacity and the computational power, and the enlarged bilingual corpora (Mariño et al., 2006). We will first introduce several MT events that promote the MT technology, and then it is the importance of MT evaluation (MTE). Subsequently, we give a brief introduction of each chapter in the thesis.

## 1.1. MT Events

There are several events that promote the development of MT and MT evaluation research.

One of which is the Open machine translation (OpenMT) Evaluation series of National Institute of Standards and Technology (NIST) that are very prestigious evaluation campaigns, the corpora including Arabic-English and Chinese-English language pairs from 2001 to 2009. The OpenMT evaluation series aim to help advance the state of the art MT technologies (NIST, 2002, 2009; Li, 2005) and make the system output to be an adequate and fluent translation of the original text that includes all forms.

The innovation of MT and the evaluation methods is also promoted by the annual Workshop on Statistical Machine Translation (WMT) (Koehn and Monz, 2006; Callison-Burch et al., 2007, 2008, 2009, 2010, 2011, 2012) organized by the special interest group in machine translation (SIGMT) of the Association for Computational Linguistics (ACL) since 2006. The evaluation campaigns focus on European languages. There are roughly two tracks in the annual WMT workshop including the translation task and evaluation task. From 2012, they added a new task of Quality Estimation of MT without given reference translations (unsupervised evaluation). The tested language pairs are clearly divided into two parts, English-to-other and other-to-English, relating to French, German, Spanish (Koehn and Monz, 2006), Czech (Callison-Burch et al., 2007, 2010, 2011, 2012), Hungarian (Callison-Burch et al., 2008, 2009), and Haitian Creole, featured task translating Haitian Creole SMS messages that were sent to an emergency response hotline, due to the Haitian earthquake (Callison-Burch et al., 2011).

Another promotion is the International Workshop of Spoken Language Translation (IWSLT) that is organized annually since 2004 (Eck and Hori, 2005; Paul, 2008, 2009; Paul, et al., 2010; Federico et al., 2011). This campaign has a stronger focus on speech translation including the English and Asian languages, e.g. Chinese, Japanese and Korean.

## 1.2. Importance of MT Evaluation

Due to the wide-spread development of MT systems, the MT evaluation becomes more and more important to tell us how well the MT systems perform and whether they make some progress. However, the MT evaluation is difficult because multiple reasons. The natural languages are highly ambiguous and different languages do not always express the same content in the same way (Arnold, 2003), and language variability results in no single correct translation.

The earliest human assessment methods for machine translation include the intelligibility and fidelity. They were used by the Automatic Language Processing Advisory Committee (ALPAC) around 1966 (Carroll, 1966a and 1966b). The afterwards proposed human assessment methods include adequacy, fluency, and comprehension (improved intelligibility) by Defense Advanced Research Projects Agency (DARPA) of US (White et al., 1994; White, 1995). However, the human judgments are usually expensive, time-consuming, and un-repeatable.

To overcome the weekness of manual judgments, the early automatic evaluation metrics include the word error rate (WER) (Su et al., 1992), and position independent word error rate (PER) (Tillmann et al., 1997). WER and PER are based on the Levenshtein distance that is the number of editing steps of insertions, deletions, and substitutions to match two sequences. The nowadays commonly used automatic metrics include BLEU (Papineni et al., 2002), TER (Snover et al., 2006), and METEOR (Banerjee and Lavie, 2005), etc. However, there remain some weaknesses in the existing automatic MT evaluation metrics, such as lower performances on the language pairs with English as source language, highly relying on large amount of linguistic features for good performance, and incomprehensive factors, etc.

As discussed in the works of Liu et al. (2011) and Wang and Manning (2012b), the accurate and robust evaluation metrics are very important for the development of MT technology. To address some of the existing problems in the automatic MT evaluation metrics, we first design reinforced evaluation factors to achieve robustness; then we design tunable parameters to address the language bias performance; finally, we investigate some concise linguistic features to enhance the performance our metrics. The practical performances in the ACL-WMT shared task show that our metrics were robust and achieved some improvements on the language pairs with English as source language.

## 1.3. Guidance of the Thesis Layout

This thesis is constructed as the following describes.

Chapter One is the introduction of MT development and importance of MT evaluation. The weekness of existing automatic MT evaluation metrics is briefly mentioned and we give a brief introduction about how we will address the problem.

Chapter Two is the background and related work. It contains the knowledge of manual judgment methods, automatic evaluation methods, and the evaluation methods for automatic evaluation metrics.

Chapter Three proposes the designed automatic evaluation method LEPOR of this thesis, including the introduction of each factor in the metric. The main factors in the LEPOR metric include enhanced sentence length penalty, n-gram position difference penalty, and the harmonic mean of precision and recall. We designed several different strategies to group the factors together.

Chapter Four is the improved models of the designed LEPOR metric. This chapter contains the new factors and linguistic features developed in the metric and several variants of LEPOR.

Chapter Five is the experimental evaluation of the designed LEPOR metric. It introduces the used corpora, evaluation criteria, experimental results, and comparisons with existing metrics.

Chapter Six introduces the participation in the annual international workshop of statistical MT (WMT). This chapter contains the submitted metrics and the official results in the shared tasks.

Chapter Seven is the latest development of quality estimation (QE) for MT. It introduces the difference of QE and traditional MT evaluations. This chapter also mentions our designed methods in the QE tasks.

Chapter Eight draws the conclusion and future work of this thesis.

# 2. MACHINE TRANSLATION EVALUATIONS

In this chapter, we first introduce the human judgment methods for MT; then we introduce the existing automatic evaluation metrics; finally, it is the evaluation criteria for MT evaluation.White (1995) proposes the concept of Black box evaluation. Black Box evaluation measures the quality of a system based solely upon its output, without respect to the internal mechanisms of the translation system. The coordinate methodology with it is the Glass Box evaluation, which measures the quality of a system based upon internal system properties. In this work, we mainly focus on the black box MT evaluation.

## 2.1. Human Assessment Methods for MT

This section introduces the human evaluation methods for MT, sometimes called as the manual judgments. We begin with the traditional human assessment methods and end with the advanced human assessment methods.

### 2.1.1. Traditional Manual Judgment Methods

The traditional human assessments include intelligibility, fidelity, fluency, adequacy, and comprehension, etc. There are also some further developments of these methods.

The earliest human assessment methods for MT can be traced back to around 1966, which includethe intelligibility, measuring how understandable the sentence is, and fidelity, measuring how much information the translated sentence retains as compared to the original, used by the automatic language processing advisory committee (ALPAC) (Carroll, 1966a and 1966b). ALPAC was established in 1964 by the US government to evaluate the progress in computational linguistics in general and machine translation.The requirement that a translation be intelligible means that as far as possible the translation should be read as normal, well-edited prose and be readily understandable in the same way that such a sentence would be understandable if originally composed in the translation language. The requirement that a translation be of high fidelity or accuracy includes that the translation should as little as possible twist, distort, or controvert the meaning intended by the original.

On the other hand, fidelity is measured indirectly, "informativeness" of the original relative to the translation. The translated sentence is presented, and after reading it and absorbing the content, the original sentence is presented. The judges are asked to rate the original sentence on informativeness.The fidelity is measured on a scale of 0-to-9 spanning from less information, not informative at all, no really new meaning added, a slightly different "twist" to the meaning on the word level, a certain amount of information added about the sentence structure and syntactical relationships, clearly informative, very informative, to extremely informative.

Thanks to the development of computer technology in 1990s, the machine translation developed fast and the human assessment methods also did around the 1990s. As part of the Human Language Technologies Program, the Advanced Research Projects Agency (ARPA) created the methodology to evaluate machine translation systems using the adequacy, fluency and comprehension (Church et al., 1991) as the evaluation criteria in MT evaluation campaigns for the full automatic MT systems (FAMT) (White et al., 1994; White, 1995). All the three components are plotted between 0 and 1 according to the formulas (White, 1995):

$$Comprehension = \frac{\#Correct}{6} \tag{2-1}$$

$$Fluency = \frac{\frac{Judgment\ point - 1}{5-1}}{\#Sentences\ in\ passage} \tag{2-2}$$

$$Adequacy = \frac{\frac{Judgment\ point - 1}{5-1}}{\#Fragments\ in\ passage} \tag{2-3}$$

The evaluator is asked to look at each fragment, usually less than a sentence in length, delimited by syntactic constituent and containing sufficient information, and judge the adequacy on a scale 1-to-5, and the results are computed by averaging the judgments over all of the decisions in the translation set and mapped onto a 0-to-1 scale. Adequacy is similar to the fidelity assessment used by ALPAC.

The fluency evaluation is compiled with the same manner as for the adequacy except for that the evaluator is to make intuitive judgments on a sentence by sentence basis for each translation.The evaluators are asked to determine whether the translation is good English without reference to the correct translation. The fluency evaluation is to determine whether the sentence is well-formed and fluent in context.

The modified comprehension develops into the "Informativeness", whose objective is to measure a system's ability to produce a translation that conveys sufficient information, such that people can gain necessary information from it. Developed from the reference set of expert translations, six questions have six possible answers respectively including "none of above" and "cannot be determined". The results are computed as the number of right answers for each translation, averaged for all outputs of each system and mapped into a 0-to-1 scale.

There are some further developments of the above manual judgments with some examples as below.

Linguistics Data Consortium (LDC, 2005) develops two five-point-scales representing fluency and adequacy for the annual NIST Machine Translation Evaluation Workshop, which become the widely used methodology when manually evaluating MT is to assign values, e.g. utilized in the WMT workshops (Koehn and Monz, 2006; Callison-Burch et al., 2007) and IWSLP evaluation campaigns (Eck and Hori, 2005; Paul et al., 2010).

**Table 21: Fluency and Adequacy Criteria.**

| Fluency | | Adequacy | |
|---|---|---|---|
| Score | Meaning | Score | Meaning |
| 1 | Incomprehensible | 1 | None |
| 2 | Disfluent English | 2 | Little information |
| 3 | Non-Native English | 3 | Much information |
| 4 | Good English | 4 | Most information |
| 5 | Flawless English | 5 | All information |

The five point scale for adequacy indicates how much of the meaning expressed in the reference translation is also expressed in a hypothesis translation; the second five point scale indicates how fluent the translation is, involving both grammatical correctness and idiomatic word choices. When translating into English the values correspond to the Table 2-1.

Other related works include Bangalore et al. (2000) and Reeder (2004), Callison-Burch et al. (2007), Przybocki et al. (2008), Specia et al. (2011), Roturier and Bensadoun (2011), etc.

### 2.1.2. Advances of Human Assessment Methods

The advanced manual assessments include task oriented method, extended criteria, binary system comparison, utilization of post-editing and linguistic tools, and online manual evaluation, etc.

White and Taylor (1998) develop a task-oriented evaluation methodology for Japanese-to-English translation to measure MT systems in light of the tasks for which their output might be used. They seek to associate the diagnostic scores assigned to the output used in the DARPA evaluation with a scale of language-dependent tasks such as scanning, sorting, and topic identification.

King et al. (2003) extend a large range of manual evaluation methods for MT systems, which, in addition to the early talked accuracy, include suitability, whether even accurate results are suitable in the particular context in which the system is to be used; interoperability, whether with other software or with hardware platforms; reliability, i.e. don't break down all the time or take a long time to run again after breaking down; usability, easy to get the interfaces, easy to learn and operate, and looks pretty; efficiency, when needed, keep up with the flow of dealt documents; maintainability, being able to modify the system in order to adapt it to particular users; and portability, one version of a system can be replaced by a new one.

Based on the ideas that the final goal of most evaluations is to rank the different systems and human judge can normally choose the best one out of two translations, Vilar et al. (2007) design a novel human evaluation scheme by the direct comparison of pairs of translation candidates. The proposed method has lower dependencies on extensive evaluation guidelines and typically focuses on the ranking of different MT systems.

A measure of quality is to compare translation from scratch and post-editing the result of an automatic translation. This type of evaluation is however time consuming and depends on the skills of the translator and post-editor. One example of a metric that is designed in such a manner is the human translation error rate (HTER) (Snover et al., 2006), based on the number of editing steps, computing the editing steps between an automatic translation and a reference translation. Here, a human annotator has to find the minimum number of insertions, deletions, substitutions, and shifts to convert the system output into an acceptable translation. HTER is defined as the number of editing steps divided by the number of words in the acceptable translation.

Naskar et al. (2011) describe an evaluation approach DELiC4MT, diagnostic evaluation using linguistic checkpoints for MT, to conduct the MT evaluation with a flexible framework, which is experienced with three language pairs from German, Italian and Dutch into English. It makes use of many available component and representation standards, e.g. the GIZA++ POS taggers and word aligner (Och and Ney, 2003), public linguistic parsing tool, the KYOTO Annotation Format (Bosma et al., 2009) to represent textual analysis, and the Kybots (Vossen et al.,2010) to define the evaluation targets (linguistic checkpoint). The diagnostic evaluation scores reveal that the rule-based systems Systran and FreeTranslation are not far behind the SMT systems Google Translate and Bing Translator, and show some crucial knowledge to the MT developers in determining which linguistic phenomena their systems are good at dealing with.

Federmann (2012) describes Appraise, an open-source toolkit supporting manual evaluation of machine translation output. The system allows collecting human judgments on translation output and implementing annotation tasks such as 1) quality checking, 2) translation ranking, 3) error classification, and 4) manual post-editing. It features an extensible, XML-based format for import/export and can be easily adapted to new annotation tasks. Appraise also includes automatic computation of inter-annotator agreements allowing quick access to evaluation results.

Other related works include Miller and Vanni (2005), Bentivogli et al. (2011), Paul et al. (2012), etc.

## 2.2. Automatic Evaluation Metrics for MT

Manual evaluation suffers some disadvantages such as time-consuming, expensive, not tunable, and not reproducible. Some researchers also find that the manual judgments sometimes result in low agreement (Callison-Burch et al., 2011). For instance, in the WMT 2011 English-Czech task, multi-annotator agreement kappa value is very low, and even the same strings produced by two systems are ranked differently each time by the same annotator. Due to the weaknesses in human judgments, automatic evaluation metrics have been widely used for machine translation. Typically, they compare the output of machine translation systems against human translations but there are also some metrics that do not use the reference translation. Common metrics measure the overlap in words and word sequences, as well as word order and edit distance (Cherry, 2010). Advanced metrics also take linguistic features into account such as syntax, semantics, e.g. POS, sentence structure, textual entailment, paraphrase, synonyms and named entities, and language models.

### 2.2.1. Metrics Based on Lexical Similarity

This section discusses the automatic MT evaluation metrics employing the lexical similarity including the factors of edit distance, precision, recall, and word order. Some of the metrics also employ the n-gram co-occurrence (Doddington, 2002) information and others use the unigram matching only. Some metrics mentioned in this section also utilize the linguistic features.

#### 2.2.1.1. Edit Distance Metrics

By calculating the minimum number of editing steps to transform output to reference, Su et al. (1992) introduce the word error rate (WER) metric from speech recognition into MT evaluation. This metric takes word order into account, and the operations include insertion (adding word), deletion (dropping word) and replacement (or substitution, replace one word with another) using the Levenshtein distance, the minimum number of editing steps needed to match two sequences.The measuring formula is shown as below whose value ranges from 0 (the best) to 1 (the worst).

$$WER = \frac{substitution + insertion + deletion}{reference_{length}} \tag{2-4}$$

Because WER is to compare the raw translation output of a MT system with the final revised version that is used directly as a reference, this method can reflect real quality gap between the system performance and customer expectation. The Multi-reference WER is later defined in (Nießen et al., 2000). They compute an "enhanced" WER as follows: a translation is compared to all translations that have been judged "perfect" and the most similar sentence is used for the computation of the edit distance.

One of the weak points of the WER is the fact that word ordering is not taken into account appropriately.The WER scores very low when the word order of system output translation is "wrong" according to the reference. In the Levenshtein distance, the mismatches in word order require the deletion and re-insertion of the misplaced words.However, due to the diversity of language expression, some so-called "wrong" order sentences by WER also prove to be good translations. To address this problem in WER, the position-independent word error rate (PER) (Tillmann et al., 1997) ignores word order when matching output and reference.Without taking into account of the word order, PER counts the number of times that identical words appear in both sentences. Depending on whether the translated sentence is longer or shorter than the reference translation, the rest of the words are either insertions or deletions. PER is guaranteed to be less than or equal to the WER.

$$PER = 1 - \frac{correct - max(0, output_{length} - reference_{length})}{reference_{length}} \tag{2-5}$$

$$PER = \frac{lost_{number} + redundancy_{number}}{reference_{length}} \tag{2-6}$$

where $lost_{number}$ means the number of words that appear in the reference but not in the output, and $redundancy_{number}$ means the difference value of the $output_{length}$ and $reference_{length}$ when the output is longer.

Another way to overcome the unconscionable penalty on word order in the Levenshtein distance is adding a novel editing step that allows the movement of word sequences from one part of the output to another. This is something a human post-editor would do with the cut-and-paste function of a word processor. In this light, Snover et al. (2006) design the translation edit rate (TER) metric that is also based on Levenshtein distance but adds block movement (jumping action) as an editing step. The weakness is that finding the shortest sequence of editing steps is a computationally hard problem.

Other related researches using the edit distances as features include (Akiba, et al., 2001), (Akiba, et al., 2006), (Leusch et al., 2006), TERp (Snover et al., 2009), Dreyer and Marcu (2012), and (Wang and Manning, 2012a), etc.

#### 2.2.1.2. Precision Based Metrics

Precision is a widely used criterion in the MT evaluation tasks. For instance, if we use the $\#correct$ to specify the number of correct words in the output sentence and the $\#output$ as the total number of the output sentence, then the precision score of this sentence can be calculated by their quotient value.

$$Precision = \frac{\#correct}{\#output} \tag{2-7}$$

The commonly used evaluation metric BLEU (**bil**ingual **e**valuation **u**nderstudy) (Papineni et al., 2002) is based on the degree of n-gram overlapping between the strings of words produced by the machine and the human translation references at the corpus level. BLEU computes the precision for n-gram of size 1-to-4 with the coefficient of brevity penalty. The theory under this design is that if most of the outputs are right but with too short output (e.g. many meanings of the source sentences lost), then the precision value may be very high but this is not a good translation; the brevity-penalty coefficient will decrease the final score to balance this phenomenon.

$$BLEU = Brevity_{penalty} \times exp \sum_{n=1}^{N} \lambda_n logPrecision_n \tag{2-8}$$

$$Brevity_{penalty} = \begin{cases} 1 & if\, c > r \\ e^{(1-\frac{r}{c})} & if\, c \leq r \end{cases} \tag{2-9}$$

where $c$ is the total length of candidate translation corpus (the sum of sentences' length), and $r$ refers to the sum of effective reference sentence length in the corpus. The effective sentence means that if there are multi-references for each candidate sentence, then the nearest length as compared to the candidate sentence is selected as the effective one.

The n-gram matching of candidate translation and reference translation is first performed at sentence level. The unigram matching is designed to capture the adequacy and the n-gram matching is to achieve the fluency evaluation. Then the successfully matched n-gram numbers are added sentence by sentence and the n-gram precisions and brevity penalty values in the formula are calculated at the corpus level instead of sentence level.

BLEU is now still one of the commonly used metrics by researchers to show their improvements in their researches including the translation quality and evaluation metrics. For example, Nakov and Ng (2012) show their improved language model on machine translation quality for resource-poor language by the gaining of up to several points of BLEU scores; Sanchez-Martınez and Forcada (2009) describe a method for the automatic inference of structural transfer rules to be used in a shallow-transfer MT system from small parallel corpora with the verifying metrics TER and BLEU; Li et al. (2011) propose a feedback selecting algorithm for manually acquired rules employed in a Chinese to English MT system stating the improvement of SMT quality by 17.12% and by 5.23 in terms of case-insensitive BLEU-4 score over baseline. Actually, BLEU has a wider applicability than just MT. Alqudsi et al. (2012) extend its use to evaluate the generation of natural language and the summarization of systems.

In the BLEU metric, the n-gram precision weight $\lambda_n$ is usually selected as uniform weight $1/N$. However, the 4-gram precision value is usually very low or even zero when the test corpus is small. Furthermore, the geometric average results in 0 score whenever one of the component n-grams scores is 0. To weight more heavily those n-grams that are more informative, Doddington (2002) proposes the NIST metric with the information weight added.

$$Info = log_2\left(\frac{\#the\ occurrence\ of\ w_1 \ldots w_{n-1}}{\#the\ occurrence\ of\ w_1 \ldots w_n}\right) \quad (2\text{-}10)$$

Furthermore, he replace the geometric mean of co-occurrences with the arithmetic average of n-gram counts, extend the n-gram into 5-gram (N=5), and select the average length of reference translations instead of the nearest length. The arithmetic mean ensures that the co-occurrence for different n-gram can be weighted.

$$NIST = \sum_{n=1}^{N} \left\{ \frac{\sum_{\substack{all\ w_1 \ldots w_n \\ that\ cooccur}} Info(w_1 \ldots w_n)}{\sum_{\substack{all\ w_1 \ldots w_n \\ in\ sysoutput}} (1)} \right\} \times exp\left\{ \beta log^2 \left[ min\left( \frac{L_{sys}}{\bar{L}_{ref}}, 1 \right) \right] \right\}$$

(2-11)

where $w_1 \ldots w_n$ means the words sequence, $\bar{L}_{ref}$ is the average number of words in a reference translation, averaged over all reference translations. The experiments show that NIST provides improvement in score stability and reliability, and higher correlation with human adequacy judgments than BLEU on four languages, Chinese, French, Japanese and Spanish. However, NIST correlates lower with the human Fluency judgments than BLEU on the other three corpora except for Chinese.

Combining BLEU with weights of statistical salience from vector space model (Babych et al., 2003), which is similar to TF.IDF score (SaltonandLesk, 1968), (Babych, 2004) and (Babych and Hartley, 2004a and 2004b) describe an automated MT evaluation toolkit weighted N-gram model (WNM) and implement a rough model of legitimate translation variation (LTV). The method has been tested by correlation with human scores on DARPA 94 MT evaluation corpus (White et al, 1994).

Other research works based precision include Liu and Gildea (2007 and 2006), etc.

#### 2.2.1.3. Recall Based Metrics

Recall is another crucial criterion in the MT evaluation. For instance, if the $\#reference$ means the number of words in the reference sentence, and the $\#correct$ also specify the number of correct words in the output sentence, then the recall value is calculated as:

$$Recall = \frac{\#correct}{\#reference} \quad (2\text{-}12)$$

Different with precision criterion, which reflects the accuracy of the system output, recall value reflects the loyalty of the output to the reference (or input) (Melamed et al., 2003).

ROUGE (Lin and Hovy 2003; Lin 2004a) is a recall-oriented automated evaluation metric, which is initially developed for summaries. Automated text summarization has drawn a lot of interest in the natural language processing and information retrieval communities. A series of workshops on automatic text summarization (WAS, 2002) are held as special topic sessions in ACL. Following the adoption by the machine translation community of automatic evaluation using the BLEU/NIST scoring process, Lin (2004a) conducts a study of a similar idea for evaluating summaries. The experiments show that automatic evaluation using unigram co-occurrences, i.e. ROUGE, between summary pairs correlates surprising well with human evaluations, based on various statistical metrics; on the other hand, direct application of the BLEU evaluation procedure does not always give good results. They also explore the effect of sample size in (Lin, 2004b) and apply the ROUGE into automatic machine translation evaluation in the work (Lin and Och, 2004a and 2004b). Furthermore, Lin and Och (2004) introduce a family of ROUGE including three measures, of which ROUGE-S is a skip bigram F-measure, ROUGE-L and ROUGE-W are measures based on the length of the longest common subsequence of the sentences. ROUGE-S has a similar structure to the bigram PER and they expect ROUGE-L and ROUGE-W to have similar properties to WER.

Other related works include (Leusch et al., 2006) and (Lavie et al., 2004) that talk about the significance of recall values in automatic evaluation of machine translation.

#### 2.2.1.4. Combination of Precision and Recall

As mentioned in the precious section of this paper, the precision value reflects the accuracy, how much proportion of the output is correct, of the automatic MT system and the recall value reflects the loyalty, how much meaning is lost or remained, of the output to the inputs, both of which are the crucial criteria to judge the quality of the translations. To evaluate the MT quality more reasonable, it is not difficult to think of the combination of these two factors.

F-measure is the combination of precision (P) and recall (R), which is firstly employed in the information retrieval and latterly has been adopted by the information extraction, MT evaluation and other tasks. Let's see a set of formula first.

$$E = 1 - \left(\frac{\alpha}{P} + \frac{1-\alpha}{R}\right)^{-1}, \ let\, \alpha = \frac{1}{1+\beta^2}\ then$$

$$E = 1 - \left(1+\beta^2\right)\frac{PR}{R+\beta^2 P}, \ let\, F_\beta = 1 - E\ then$$

$$F_\beta = \left(1+\beta^2\right)\frac{PR}{R+\beta^2 P} \quad (2\text{-}13)$$

The variable $F_\beta$ measures the effectiveness of retrieval with respect to a user who attaches $\beta$ times as much importance to recall as precision. The effectiveness measure ($E$ value) is defined in (van Rijsbergen. 1979). In the effectiveness measure, there is a parameter $\alpha$ which sets the trade-off between Precision and Recall. When an equal trade-off is desired, $\alpha$ is set to 0.5. The first full definition of the F-measure ($F_\beta$) to evaluation tasks of information extraction technology was given by (Chinchor, 1992) in the fourth message understanding conference (MUC-4).

Traditional F-measure or balanced F-score ($F_1$ score) is exactly the harmonic mean of precision and recall (put the same trade-off on precision and recall, $\beta = 1$) (Sasaki and Fellow, 2007).

$$F_1 = \frac{2PR}{R+P} \quad (2\text{-}14)$$

If we bring the precision and recall formula introduced in the precious sections into the F-measure, we can get the following inferred formula.

$$F_1 = \frac{2PR}{R+P} = \frac{2 \times \#correct}{\#reference + \#output} \quad (2\text{-}15)$$

$$F_{(\alpha P,\, \beta R)} = \frac{(\alpha+\beta)PR}{\alpha R + \beta P} = \frac{(\alpha+\beta) \times \#correct}{\beta\#reference + \alpha\#output} \quad (2\text{-}16)$$

where $F_{(\alpha P, \beta R)}$ means assign the weight $\alpha$ and $\beta$ respectively to Precision and Recall. We should note that the unigram precision, recall and F-score do not take word order into consideration.

Riezler and Maxwell III(2005) investigate some pitfalls regarding the discriminatory power of MT evaluation metrics and the accuracy of statistical significance tests. In a discriminative re-ranking experiment for phrase-based SMT, they show that the NIST metric is more sensitive than BLEU or F-score despite their incorporation of aspects of fluency or meaning adequacy into MT evaluation.Pointing out a well-known problem of randomly assessing significance in multiple pairwise comparisons, they recommend for multiple comparisons of subtle differences to combine the NIST score for evaluation with the approximate randomization test for significance testing, at stringent rejection levels.

F-measure is based on the unigram matching and two sentences containing the same words always get the same F-measure rating regardless of the correct order of the words in the sentence. To eliminate this drawbacks, BLEU and NIST reward the correct word order by double-counting all sub-runs, where the factor $run$ is the contiguous sequence of matching words in the matching M (M is usually a sentence). On the other hand, GTM (general text matching) that is proposed in (Turian et al., 2003) rewards the contiguous sequences of correctly translated words by the assigned weight to the $run$. GTM is based on the F-measure but adds the maximum match size (MMS) information in the calculation of precision and recall.

$$size(M) = \sqrt[e]{\sum_{run \in M} length(run)^e} \qquad (2\text{-}17)$$

They first define the weight of a $run$ to be the $length(run)^e$, then they generalize the definition of match size as $size(M)$. The reward is controlled by parameter $e$. The contiguous sequences of words are rewarded and penalized respectively when $e \in (0,1)$ and $e \in (1,\infty)$. When $e = 1$, the GTM score achieves the same performance with the original F-measure. GTM calculates word overlap between a reference and a solution, without double counting duplicate words. Furthermore, BLEU and NIST are difficult to gain insight from the experiment scores, whereas GTM performs the measures that have an intuitive graphical interpretation and can facilitate insights into how MT systems might be improved.

BLEU is an n-gram precision based metric and performs the exact words matching. However, Banerjee and Lavie (2005) find that the recall value plays a more important role than precision to obtain higher correlation with human judgments and design a novel evaluation metric METEOR. METEOR is based on general concept of flexible unigram matching, unigram precision and unigram recall, e.g. unigram F-measure, including the match of words that are simple morphological variants of each other by the identical stem and words that are synonyms of each other. METEOR assigns 9 times as importance of recall as precision value, i.e., $\alpha = 1, \beta = 9$ in the F-measure $F_{(\alpha P, \beta R)}$.

To measure how well-ordered the matched words in the candidate translation are in relation to the human reference, METEOR introduces a novel penalty coefficient by employing the number of matched chunks.

$$Penalty = 0.5 \times \left( \frac{\#chunks}{\#unigrams_mathed} \right)^3 \tag{2-18}$$

$$METEOR = \frac{10PR}{R + 9P} \times (1 - Penalty) \tag{2-19}$$

When there are no bigram or longer matches between the candidate translation and the reference, there are as many chunks as there are unigram matches. Experiments tested on LDC TIDES 2003 Arabic-to-English and Chinese-to-English show that all of the individual factors included within METEOR contribute to improved correlation with human judgments, which means that METEOR achieve higher correlation score than the unigram precision, unigram recall and unigram F-measure, in addition to the BLEU and NIST metrics. The enhanced version of METEOR (Lavie and Agarwal, 2007) also employs the paraphrases, using WordNet a popular ontology of English words, into the matching period. The weakness of METEOR is the computationally expensive word alignment.

Other related works using the combination of precision and recall include Lita et al. (2005), Chen and Kuhn (2011), Chen et al. (2012a), etc.

#### 2.2.1.5. Word Order Utilization

The right word order places an important role to ensure a high quality translation output. However, the language diversity also allows different appearances or structures of the sentence. How to successfully achieve the penalty on really wrong word order (wrongly structured sentence) instead of on the "correctly" different order, the candidate sentence that has different word order with the reference is well structured, compared with the reference translation, attracts a lot of interests from researchers in the NLP literature. In fact, the Levenshtein distance and n-gram based measures contain the word order information. Here, we introduce severale valuation metrics that are not based on Levenshtein distance but also take the word order in to consideration.

Featuring the explicit assessment of word order and word choice, Wong and Kit (2008 and 2009) develop the evaluation metric ATEC, assessment of text essential characteristics, which is also based on precision and recall criteria but with the designed position difference penalty coefficient attached. The word choice is assessed by matching word forms at various linguistic levels, including surface form, stem, sound and sense, and further by weighing the informativeness of each word. The word order is quantified in term of the discordance of word position and word sequence between the translation candidate and its reference. The evaluation on the Metrics MATR08, the LDC MTC2 and MTC4 corpora demonstrates an impressive positive correlation to human judgments at the segment level.The parameter-optimized version of ATEC and its performance is described in (Wong and Kit,2010).

Other related works include Zhou et al. (2008), Isozaki et al.(2010), Chen et al. (2012b), Popovic (2012), etc.

### 2.2.2. Combination with Linguistic Features

Although some of the previous mentioned metrics employ the linguistic information into consideration, e.g. the semantic information synonyms and stemming in METEOR, the lexical similarity mainly focus on the exact matches of the surface words in the output translation. The advantages of the metrics based on lexical similarity are that they perform well in capturing the translation fluency as mentioned in (Lo et al., 2012), and they are very fast and low cost. On the other hand, there are also some weaknesses, for instance, the syntactic information is rarely considered and the underlying assumption that a good translation is one that shares the same lexical choices as the reference translations is not justified semantically. Lexical similarity does not adequately reflect similarity in meaning. Translation evaluation metric that reflects meaning similarity needs to be based on similarity of semantic structure not merely flat lexical similarity.

In this section we focus on the introduction of linguistic features utilized into the evaluation including the syntactic features and semantic information.

#### 2.2.2.1. Syntactic Similarity

The lexical similarity metrics tend to perform on the local level without considering the overall grammaticality of the sentence or sentence meaning. To address this problem, the syntax information should be considered. Syntactic similarity methods usually employ the features of morphological part-of-speech information, phrase categories or sentence structure generated by the linguistic tools such as language parser or chunker.

**POS information**

In grammar, a part of speech, also called a lexical category, is a linguistic category of words or lexical items, which is generally defined by the syntactic or morphological behaviour of the lexical item. Common linguistic categories of lexical items include noun, verb, adjective, adverb, and preposition, etc. To reflect the syntactic quality of automatically translated sentences, some researchers employ the POS information into their evaluation.

Using the IBM model one, Popovic et al. (2011) evaluate the translation quality by calculating the similarity scores of source and target (translated) sentence without using reference translation based on the morphemes, 4-gram POS and lexicon probabilities. This evaluation metric MP4IBM1 relies on the large parallel bilingual corpus to extract the lexicon probability, precise POS tagger to gain the details about verb tenses, cases, number, gender, etc., and other linguistic tool to split words into morphemes. The experiments show good performance with English as the source language but very weak performance when English is the target language. For instance, the correlation score with human judgments is 0.12 and 0.08 respectively on Spanish-to-English and French-to-English WMT corpus (Popovic et al., 2011), which means very low correlation. Other similar works using POS information include the Giménez and Márquez (2007), Popovic and Ney (2007), Dahlmeier et al. (2011), Liu et al. (2010), and our own metric hLEPOR (Han et al., 2013f), etc.

**Phrase information**

To measure a MT system's performance in translating new text-types, such as in what ways the system itself could be extended to deal with new text-types, Povlsen, et al. (1998) perform a research work focusing on the study of English-to-Danish machine-translation system PaTrans (Bech 1997), which covers the domain of petro-chemical and mechanical patent documents. The overall evaluation and quality criterion is defined in terms of how much effort it takes to post-edit the text after having been translated by the MT system. A structured questionnaire rating different error types is given to the post-editors. In addition to the lexical analysis such as the identifying of *deverbal nouns*, *adjectives*, *homograghs* (with different target translations) and *words* (translate into nexus adverbs), the syntactic constructions and semantic features are explored with more complex linguistic knowledge, such as the identifying of *valency* and *non-valency bond prepositions*, *fronted adverbial subordinate clauses, prepositional phrases*, and *part-of-speech disambiguation* with constraint-grammar parser ENGCG (Voutilainen et al., 1992). In order to achieve consistency and reliability, the analysis of thenew text-types is automated as far as possible. In the experiments, a reference text, known as a good text, is first analysed using the procedure in order to provide a benchmark against which to assess the results from analysing the new text-types.After running the evaluation, a representative subset of the new text-types is then selected and translated by a slightly revised version of the MT system and assessed by the post-editors using the same questionnaire.Their evaluation is a first stage in an iterative process, in which the suite of programs is extended to account for the newly identified gaps in coverage and the evaluation of the text type carried out again.

Assuming that the similar grammatical structures should occur on both source and translations, Avramidis et al. (2011) perform the evaluation on source (German) and target (English) sentence employing the features of sentence length ratio, unknown words, phrase numbers including noun phrase, verb phrase and prepositional phrase. The used tools include Probabilistic Context Free Grammar (PCFG) parser (Petrov et al., 2006) and statistical classifiers using Naive Bayes and K-Nearest Neighbour algorithm. However, the experiment shows lower performance than the BLEU score. The further development of this metric (DFKI) is introduced in (Avramidis, 2012) where the features are expanded with verbs, nouns, sentences, subordinate clauses and punctuation occurrences to derive the adequacy information.

Han et al. (2013d) develop a phrase tagset mapping between English and French treebanks, and perform the MT evaluation work on the developed universal phrase tagset instead of the surface words of the sentences. The experiments on ACL-WMT (2011 and 2012) corpora, without using reference translations, yield promising correlation scores as compared to the traditional evaluation metrics BLEU and TER.

Other similar works using the phrase similarity include the (Li et al., 2012) that uses noun phrase and verb phrase from chunking and (Echizen-ya and Araki, 2010) that only uses the noun phrase chunking in automatic evaluation.

**Sentence structure information**

To address the overall goodness of the translated sentence's structure, Liu and Gildea (2005) employ constituent labels and head-modifier dependencies from language parser as syntactic features for MT evaluation. They compute the similarity of dependency trees between the candidate translation and the reference translations using their designed methods HWCM (Headword Chain Based Metric), STM (sub-tree metric), DSTM (dependency sub-tree metric), TKM (kernel-based sub-tree metric) and DTKM (dependency tree kernel metric). The overall experiments prove that adding syntactic information can improve the evaluation performance especially for predicting the fluency of hypothesis translations.

Featuring that valid syntactic variations in the translation can avoid the unfairly penalize, Owczarzak et al. (2007) develop a MT evaluation method using labelled dependencies that are produced by a lexical-functional grammar parser in contrast to the string-based methods. Similarly, the dependency structure relation is also employed in the feature set by the work (Ye et al., 2007), which performs the MT evaluation as a ranking problem.

Other works that using syntactic information into the evaluation are shown in (Lo and Wu, 2011a) and (Lo et al., 2012) that use an automatic shallow parser, (Mutton et al., 2007) that focuses on the fluency criterion, and (Fishel et al., 2012; Avramidis, 2012; Felice and Specia, 2012) that use different linguistic features.

#### 2.2.2.2. Semantic Similarity

As contrast to the syntactic information, which captures the overall grammaticality or sentence structure similarity, the semantic similarity of the automatic translations and the source sentences (or references) can be measured by the employing of some semantic features, such as the named entity, synonyms, semantic roles, paraphrase and textual entailment.

**Named entity information**

To capture the semantic equivalence of sentences or text fragments, the named entity knowledge is brought from the literature of named-entity recognition, also called as entity extraction or entity identification, which is aiming to identify and classify atomic elements in the text into different entity categories (Marsh and Perzanowski, 1998; Guo et al., 2009). The commonly used entity categories include the names of persons, locations, organizations and times (Han et al., 2013e; 2013g), etc.

In the MEDAR, an international cooperation between the EU and the Mediterranean region on Speech and Language Technologies for Arabic, 2011 evaluation campaign, two SMT systems based on Moses (Koehn et al., 2007) are used as baselines respectively for English-to-Arabic and Arabic-to-English directions. The Baseline-1 system adapts SRILM (Stockle, 2002), GIZA++ (Och & Ney, 2003) and a morphological analyzer to Arabic, whereas Baseline-2 system also utilizes OpenNLP[1] toolkit to perform named entity detection, in addition to other packages that provides tokenizing, POS tagging and base phrase chunking for Arabic text (Hamon and Choukri, 2011). The experiments show that the low performances from the perspective of named entities, many entities are either not translated or not well translated, cause a drop in fluency and adequacy.

[1]http://opennlp.apache.org/index.html

Other such related works include Buck (2012), Raybaud et al. (2011), and Finkel et al. (2005), etc.

**Synonym information**

Synonyms are used to specify the words that have the same or close meanings. One of the widely used synonym database in NLP literature is the WordNet (Miller et al., 1990; Fellbaum, 1998), which is an English lexical database grouping English words into sets of synonyms. WordNet classifies the words mainly into four kinds of part-of-speech (POS) categories including Noun, Verb, Adjective, and Adverb without prepositions, determiners, etc. Synonymous words or phrases are organized using the unit of synset. Each synset is a hierarchical structure with the words in different levels according to their semantic relations. For instance, the words in upper level belong to the words (hypernym) in lower level.

Utilizing the WordNet and the semantic distance designed by (Wu and Palmer,1994) to identify near-synonyms, Wong and Kit (2012) develop a document level evaluation metric with lexical cohesion device information. They define the lexical cohesion as the content words of synonym and near-synonym that appear in a document. In their experiments, the employed lexical cohesion has a weak demand for language resource as compared to the other discourse features such as the grammatical cohesion, so it is much unaffected by grammatical errors that usually appear in translation outputs. The performances on the corpora of MetricsMATR 2008 (Przybockiet al., 2009) and MTC-4 (Ma, 2006) show high correlation rate with manually adequacy judgments. Furthermore, the metrics BLEU and TER also achieve improved scores through the incorporating of the designed document-level lexical cohesion features.

Other works employing the synonym features include Chan and Ng (2008), Agarwal and Lavie (2008), and Liu and Ng (2012), etc.

**Semantic roles**

The semantic roles are employed by some researchers as linguistic features in the MT evaluation. To utilize the semantic roles, the sentences are usually first shallow parsed and entity tagged. Then the semantic roles used to specify the arguments and adjuncts that occur in both the candidate translation and reference translation. For instance, the semantic roles introduced by Giménez and Márquez (2007, 2008) include causative agent, adverbial adjunct, directional adjunct, negation marker, and predication adjunct, etc. In the further development, Lo and Wu (2011a and 2011b) design the metric MEANT to capture the predicate-argument relations as the structural relations in semantic frames, which is not reflected by the flat semantic role label features in the work of (Giménez and Márquez, 2007). Furthermore, instead of using uniform weights, Lo, Tumuluru and Wu (2012) weight the different types of semantic roles according to their relative importance to the adequate preservation of meaning, which is empirically determined. Generally, the semantic roles account for the semantic structure of a segment and have proved effective to assess adequacy in the above papers.

**Textual entailment**

Textual entailment is usually used as a directive relation between text fragments.If the truth of one text fragment TA follows another text fragment TB, then there is a directional relation between TA and TB (TB=>TA). Instead of the pure logical or mathematical entailment, the textual entailment in natural language processing (NLP) is usually performed with a relaxed or loose definition (Dagan et al., 2005; 2006). For instance, according to text fragment TB, if it can be inferred that the text fragment TA is most likely to be true then the relationship TB=>TA also establishes. That the relation is directive also means that the inverse inference (TA=>TB) is not ensured to be true (Dagan and Glickman, 2004).

To address the task of handling unknown terms in SMT, Mirkin et al. (2009) propose a Source-Language entailment model. Firstly they utilize the source-language monolingual models and resources to paraphrase the source text prior to translation. They further present a conceptual extension to prior work by allowing translations of entailed texts rather than paraphrases only. This method is experimented on some 2500 sentences with unknown terms and substantially increases the number of properly translated texts.

Other works utilizing the textual entailment can be referred to Pado et al. (2009a and 2009b), Lo and Wu (2011a), Lo et al. (2012), Aziz et al. (2010), and Castillo and Estrella (2012), etc.

**Paraphrase features**

Paraphrase is to restatement the meaning of a passage or text utilizing other words, which can be seen as bidirectional textual entailment (Androutsopoulos and Malakasiotis, 2010). Instead of the literal translation, word by word and line by line, used by metaphrase, paraphrase represents a dynamic equivalent. For instance, "He is a great man" may be paraphrased as "He has ever done a lot of great things". Further knowledge of paraphrase from the aspect of linguistics is introduced in the works of (McKeown, 1979; Meteer and Shaked, 1988; Dras, 1999; Barzilay and Lee, 2003). As an example, let's see the usage of paraphrase and other linguistic features in the improvement of TER evaluation metric.

While Translation Edit Rate (TER) metric (Snover 2006) has been shown to correlate well with human judgments of translation quality, it has several flaws, including the use of only a single reference translation and the measuring of similarity only by exact word matches between the hypothesis and the reference (Snover et al., 2011). These flaws are addressed through the use of Human-Mediated TER (HTER), but are not captured by the automatic metric.To address this problem, Snover et al. (2009a; 2009b) describea new evaluation metric TER-Plus (TERp). TERp uses all the edit operations of TER, Matches, Insertions, Deletions, Substitutions and Shifts, as well as three new edit operations, Stem Matches, Synonym Matches (Banerjee and Lavie 2005) and Phrase Substitutions (Zhou et al., 2006; Kauchak 2006). TERp identifies words in the hypothesis and reference that share the same stem using the Porter stemming algorithm (Porter, 1980). Two words are determined to be synonyms if they share the same synonym set according to WordNet (Fellbaum, 1998). Sequences of words in the reference are considered to be paraphrases of a sequence of words in the hypothesis if that phrase pair occurs in TERp phrase table. They present a correlation study comparing TERp to BLEU, METEOR and TER, and illustrate that TERp can better evaluate translation adequacy.

Other works using the paraphrase information can be seen in (Owczarzak et al., 2006), (Zhou et al., 2006), and (Kauchak and Barzilay, 2006), etc.There are also many researchers who combine the syntactic and semantic features together in the MT evaluation, such as Peral and Ferrandez (2003), Gimenez and Marquez (2008), Wang and Manning (2012b), de Souza et al. (2012), etc.

#### 2.2.2.3. Language Model Utilization

The language models are also utilized by the MT and MT evaluation researchers. A statistical language model usually assigns a probability to a sequence of words by means of a probability distribution.

Gamon et al. (2005) propose LM-SVM (language-model, support vector machine) method investigating the possibility of evaluating MT quality and fluency at the sentence level in the absence of reference translations. They measure the correlation between automatically-generated scores and human judgments, and evaluate the performance of the system when used as a classifier for identifying highly dysfluent and illformed sentences. They show that they can substantially improve the correlation between language model perplexity scores and human judgment by combining these perplexity scores with class probabilities from a machine-learned classifier. The classifier uses linguistic features and is trained to distinguish human translations from machine translations.

There are also some other research works that use the linguistic features, such as Wong and Kit (2011), Aikawa and Rarrick (2011),Reeder F. (2006a), etc.

Generally, the linguistic features mentioned above, including both syntactic and semantic features, are usually combined in two ways, either by following a machine learning approach (Albrecht and Hwa, 2007; Leusch and Ney, 2009), or trying to combine a wide variety of metrics in a more simple and straight forward way, such as Giménez and Márquez (2008), Specia and Giménez (2010), and Comelles et al. (2012), etc.

### 2.2.3. Combination of Different Metrics

This sub-section introduces some metrics that are designed by the combination, or offering a framework for the combination of existing metrics.

Adequacy-oriented metrics, such as BLEU, measure n-gram overlap of MT outputs and their references, but do not represent sentence-level information. In contrast, fluency-oriented metrics, such as ROUGE-W, compute longest common sub-sequences, but ignore words not aligned by the longest common sub-sequence. To address these problems, Liu and Gildea (2006) describea new metric based on stochastic iterative string alignment (SIA) for MT evaluation, which achieves good performance by combining the advantages of n-gram-based metrics and loose-sequence-based metrics. SIA uses stochastic word mapping to allow soft or partial matches between the MT hypotheses and the references. It works especially well in fluency evaluation. This stochastic component is shown to be better than PORTER-STEM and WordNet in their experiments. They also analyse the effect of other components in SIA and speculate that they can also be used in other metrics to improve their performance.

Other related works include Gimenez and Amigo (2006), Parton et al. (2011), Popovic (2011), etc.

## 2.3. Evaluation Methods of MT Evaluation

This section introduces the evaluation methods for the MT evaluation. They include the statistical significance, evaluation of manual judgments, correlation with manual judgments, etc.

### 2.3.1. Statistical Significance

If different MT systems produce translations with different qualities on a data set, how can we ensure that they indeed own different system quality? To explore this problem, Koehn (2004) performs a research work on the statistical significance test for machine translation evaluation. The bootstrap resampling method is used to compute the statistical significance intervals for evaluation metrics on small test sets. Statistical significance usually refers to two separate notions, of which one is the p-value, the probability that the observed data will occur by chance in a given single null hypothesis, and another is the Type I error, false positive, rate of a statistical hypothesis test, the probability of incorrectly rejecting a given null hypothesis in favour of a second alternative hypothesis (Hald, 1998). The fixed number 0.05 is usually referred to as the significance level, i.e. the level of significance.

### 2.3.2. Evaluating the Human Judgments

Since the human judgments are usually trusted as the golden standards that the automatic evaluation metrics should try to approach, the reliability and coherence of human judgments is very important. Cohen's kappa agreement coefficient is one of the commonly used evaluation methods (Cohen, 1960). For the problem in nominal scale agreement between two judges, there are two relevant quantities:

$$p_0 = the\ proportion\ of\ units\ in\ which\ the\ judges\ agreed \qquad (2\text{-}20)$$

$$p_c = the\ proportion\ of\ units\ for\ which\ agreement\ is\ expected\ by\ chance$$
(2-21)

The test of agreement comes then with regard to the $1 - p_c$ of the units of which the hypothesis of no association would predict disagreement between the judges. The coefficient $k$ is simply the proportion of chance-expected disagreements which do not occur, or alternatively, it is the proportion of agreement after chance agreement is removed from consideration:

$$k = \frac{p_0 - p_c}{1 - p_c} \qquad (2\text{-}22)$$

where $p_0 - p_c$ represents the proportion of the cases in which beyond-chance agreement occurs and is the numerator of the coefficient. The interval of 0-to-0.2 is slight, 0.2-to-0.4is fair, 0.4-to-0.6 is moderate, 0.6-to-0.8 is substantial, and 0.8-to-1.0 is almost perfect (Landisand Koch, 1977).

In the annual ACL-WMT workshop (Callison-Burch et al., 2007, 2008, 2009, 2010, 2011, 2012), they also use this agreement formula to calculate the Inter- and Intra-annotator agreement in the ranking task to ensure their process as a valid evaluation setup. To ensure they have enough data to measure agreement, they occasionally show annotator items that are repeated from previously completed items. These repeated items are drawn from ones completed by the same annotator and from different annotators. They measure pairwise agreement among annotators using following formula:

$$k = \frac{P(A) - P(E)}{1 - P(E)} \qquad (2\text{-}23)$$

where P(A) is the proportion of times that the annotators agree, and P(E) is the proportion of times that they will agree by chance. The agreement value k has a value of at most 1, higher rates of agreement resulting in higher k value.

### 2.3.3. Correlating the Manual and Automatic Evaluation

This section introduces the correlation criteria that are usually utilized to measure the closeness between manual judgments and automatic evaluations. They cover the Pearson correlation coefficient, Spearman correlation coefficient, and Kendall's $\tau$.

#### 2.3.3.1. Pearson Correlation Coefficient

Pearson correlation coefficient (Pearson, 1900) is usually used as the formula to calculate the system-level correlations between automatic evaluation results and human judgments.

Pearson's correlation coefficient when applied to a population is commonly represented by the Greek letter $\rho$ (rho). The correlation between random variables X and Y denoted as $\rho_{XY}$ is measured as follow (Montgomery and Runger, 1994; Montgomery and Runger, 2003).

$$\rho_{XY} = \frac{cov(X,Y)}{\sqrt{V(x)V(Y)}} = \frac{\sigma_{XY}}{\sigma_X \sigma_Y} \qquad (2\text{-}24)$$

Because the standard deviations of variable X and Y are higher than 0 ($\sigma_X > 0$ and $\sigma_Y > 0$), if the covariance $\sigma_{XY}$ between X and Y is positive, negative or zero, the correlation score $\rho_{XY}$ between X and Y will correspondingly result in positive, negative or zero, respectively. For any two random variables, the correlation score of them varies in the following interval:

$$-1 \leq \rho_{XY} \leq 1 \qquad (2\text{-}25)$$

The correlation just scales the covariance by the standard deviation of each variable. Consequently the correlation is a dimensionless quantity that can be used to compare the linear relationships between pairs of variables in different units. In the above formula, the covariance between the random variables X and Y, denoted as cov(X, Y) or $\sigma_{XY}$, is defined as:

$$\sigma_{XY} = E\left[\left(X - \mu_X\right)\left(Y - \mu_Y\right)\right] = E(XY) - \mu_X \mu_Y \quad (2\text{-}26)$$

To learn the above formula, let's first see a definition: a discrete random variable is a random variable with a finite (or countably infinite) range; a continuous random variable is a random variable with an interval (either finite or infinite) of real numbers for its range. The mean or expected value of the discrete random variable X, denoted as $\mu_X$ or E(X), is

$$\mu_X = E(X) = \sum_x x f(x) \quad (2\text{-}27)$$

A random variable X has a discrete uniform distribution if each of the n values in its range, say $x_1, x_2, \ldots, x_n$, has equal probability. Then,

$$f\left(x_i\right) = \frac{1}{n},\ \mu = \frac{1}{n}\sum\nolimits_{i=1}^{n} x_i \quad (2\text{-}28)$$

The variance of discrete random variable X, denoted as ${\sigma_X}^2$ or V(X), is

$$\sigma_X{}^2 = V(X) = E\left(X - \mu_X\right)^2 = \sum_x x^2 f(x) - \mu_X{}^2 \quad (2\text{-}29)$$

$$\sigma_X{}^2 = \frac{1}{n}\sum\nolimits_{i=1}^{n}\left(x_i - \mu_X\right)^2 = \frac{1}{n}\sum\nolimits_{i=1}^{n} x_i^2 - \mu_X{}^2 \quad (2\text{-}30)$$

The standard deviation of X is $\sigma_X = \sqrt{\sigma_X{}^2}$.

Finally, based on a sample of paired data (X, Y) as $(x_i,\ y_i)$, $i = 1\ to\ n$, the Pearson correlation coefficient is:

$$\rho_{XY} = \frac{\sum_{i=1}^{n}(x_i - \mu_x)(y_i - \mu_y)}{\sqrt{\sum_{i=1}^{n}\left(x_i - \mu_x\right)^2}\sqrt{\sum_{i=1}^{n}\left(y_i - \mu_y\right)^2}} \quad (2\text{-}31)$$

where $\mu_x$ and $\mu_y$ specify the means of discrete random variable X and Y respectively.

As the supplementary knowledge, we also list the mean and variance formula for the continuous random variable X. The mean or expected value of continuous variable X is

$$\mu = E(X) = \int_{-\infty}^{\infty} xf(x)dx \quad (2\text{-}32)$$

The variance of continuous variableX is

$$\sigma^2 = V(X) = \int_{-\infty}^{\infty} (x-\mu)^2 f(x)dx = \int_{-\infty}^{\infty} x^2 f(x)dx - \mu^2 \quad (2\text{-}33)$$

2.3.3.2. Spearman Correlation Coefficient

In order to distinguish the reliability of different MT evaluation metrics, Spearman rank correlation coefficient (a simplified version of Pearson correlation coefficient) ρ is also commonly used to calculate the system level correlation, especially for recent years WMT task (Callison-Burch et al., 2011, 2010, 2009, 2008). When there are no ties, Spearman rank correlation coefficient, which is sometimes specified as (*rs*) is calculated as:

$$\rho_{\varphi(XY)} = 1 - \frac{6\sum_{i=1}^{n} d_i^2}{n(n^2-1)} \quad (2\text{-}34)$$

where $d_i$ is the difference-value (D-value) between the two corresponding rank variables $(x_i - y_i)$ in $\vec{X} = \{x_1, x_2, \ldots, x_n\}$ and $\vec{Y} = \{y_1, y_2, \ldots, y_n\}$ describing the system $\varphi$, and $n$ is the number of variables in the system.

In the MT evaluation task, the Spearman rank correlation coefficient method is usually used by the authoritative ACL WMT to evaluate the correlation of MT evaluation metrics with the human judgments. There are some problems existing in this method. For instance, let two MT evaluation metrics MA and MB with their evaluation scores $\vec{MA} = \{0.50, 0.95, 0.45\}$ and $\vec{MB} = \{0.75, 0.77, 0.74\}$ respectively reflecting the MT systems $\vec{M} = \{M_1, M_2, M_3\}$.

Before the calculation of correlation with human judgments, they will be changed as $\breve{\vec{M}}A = \{2, 1, 3\}$ and $\breve{\vec{M}}B = \{2, 1, 3\}$ with the same rank sequence using Spearman method. Thus, the two evaluation systems will get the same correlation score with human judgments. But the two metrics reflect different results indeed: MA gives the outstanding score (0.95) to $M_2$ system and puts very low scores (0.50 and 0.45) on other two systems $M_1$ and $M_3$; on the other hand, MB thinks the three MT systems have similar performances (scores from 0.74 to 0.77). This information is lost using the Spearman rank correlation methodology.

#### 2.3.3.3. Kendall's τ

Kendall's $\tau$ (Kendall, 1938) has been used in recent years for the correlation between automatic order and reference order (Callison-Burch et al., 2012, 2011, 2010). It is defined as:

$$\tau = \frac{num\ concordant\ pairs - num\ discordant\ pairs}{total\ pairs} \quad (2\text{-}35)$$

The latest version of Kendall's $\tau$ is introduced in (Kendall and Gibbons, 1990). Lebanon and Lafferty (2002) give an overview work for Kendall's $\tau$ showing its application in calculating how much the system orders differ from the reference order. More concretely, Lapata (2003) proposes the use of Kendall's $\tau$, a measure of rank correlation, estimating the distance between a system-generated and a human-generated gold-standard order.

Kendall's $\tau$ is less widely used than Spearman's rank correlation coefficient ($rs$). The two measures have different underlying scales, and, numerically, they are not directly comparable to each other. Siegel and Castellan (1988) express the relationship of the two measures in terms of the inequality:

$$-1 \leq 3\tau - 2rs \leq 1 \quad (2\text{-}36)$$

More importantly, Kendall's $\tau$ and Spearman's rank correlation coefficient $rs$ have different interpretations. Kendall's $\tau$ can be interpreted as a simple function of the probability of observing concordant and discordant pairs (Kerridge 1975). In other words, it is the difference between the probability, that in the observed data two variables are in the same order, versus the probability, that they are in different orders. On the other hand, no simple meaning can be attributed to Spearman's rank correlation coefficient $rs$. The latter is similar to the Pearson correlation coefficient computed for values consisting of ranks. It is difficult to draw any meaningful conclusions with regard to information ordering based on the variance of ranks. In practice, while both correlations frequently provide similar answers, there are situations where they diverge. For example, the statistical distribution of $\tau$ approaches the normal distribution faster than $rs$ (Kendall and Gibbons, 1990), thus offering an advantage for small to moderate sample studies with fewer data points. This is crucial when experiments are conducted with a small number of subjects or test items. Another related issue concerns sample size. Spearman's rank correlation coefficient is a biased statistic (Kendall and Gibbons, 1990). The smaller the sample, the more $rs$ diverges from the true population value, usually underestimating it. In contrast, Kendall's $\tau$ does not provide a biased estimate of the true correlation.

# 3. LEPOR – PROPOSED MODEL

The weaknesses of Manual judgments are apparent, such as time consuming, expensive, unrepeatable, and low agreement sometimes (Callison-Burch et al., 2011). On the other hand, there are also some weaknesses of existing automatic MT evaluation methods. Firstly, they usually show good performance on certain language pairs (e.g. EN as target) and weak on others (e.g. EN as source). This is partly due to the rich English resource people can utilize to aid the evaluation, such as dictionary, synonym, paraphrase, etc. it may be also due to the different characteristics of the language pairs and they need different strategies. For instance, TER metric (Snover et al., 2006) achieved 0.89 (ES-EN) vs 0.33 (DE-EN) correlation score with human judgments on WMT-2011 tasks. Secondly, some metric rely on many linguistic features for good performances. This makes it not easy to repeat the experiment by other researchers, and it also makes the metric difficult to achieve generalization for other languages. For instance, MP4IBM1 metric (Popovic et al., 2011) utilizes large bilingual corpus, POS taggers, linguistic tools for morphemes/ POS / lexicon probabilities, etc. This metric can show good performance on its focused language pairs (English-German), but low performance on others. Thirdly, some metrics utilize incomprehensive factors. For example, the state-of-the-art BLEU metric (Papineni et al., 2002) is based on n-gram precision score only. Some researchers also held the opinion that the higher BLEU score is not necessarily indicative of better translation (Callison-Burch et al., 2006).

To address some of the existing problems in the automatic MT evaluation metrics, in this chapter, we introduce our designed models, including the augmented factors and the metrics (Han et al., 2012).

### 3.1. Enhanced Factors

In this section, we introduce the three enhanced factors of our methods including enhanced length penalty, n-gram position difference penalty, and n-gram precision and recall.

#### 3.1.1. Length Penalty

In the widely used metric BLEU (Papineni et al., 2002), it utilizes a brevity penalty for shorter sentence; however, the redundant (or longer) sentences are not penalized properly. To achieve a penalty score for the MT system, which tends to yield redundant information, we design a new version of the sentence length penalty factor, the enhanced length penalty $LP$.

In the Equation, $LP$ means Length penalty, which is defined to embrace the penalty for both longer and shorter system outputs compared with the reference translations, and it is calculated as:

$$LP = \begin{cases} e^{1-\frac{r}{c}} & if\ c < r \\ 1 & if\ c = r \\ e^{1-\frac{c}{r}} & if\ c > r \end{cases} \tag{3-1}$$

where $c$and $r$ mean the sentence length of output candidate translation and reference translation respectively. When the output length of sentence is equal to that of the reference one, $LP$ will be one which means no penalty. However, when the output length $c$ is larger or smaller than that of the reference one, $LP$ will be little than one which means a penalty on the evaluation value of LEPOR. And according to the characteristics of exponential function mathematically, the larger of numerical difference between $c$ and$r$, the smaller the value of $LP$ will be.

BLEU is measured at corpus level, which means the penalty score is directly the system level score, with $c$ and$r$ referring to the length of corpus level output and reference translations. Our length penalty score is measured in different way, first by sentence level.

### 3.1.2. N-gram Position Difference Penalty

The word order information is introduced in the research work of (Wong and Kit, 2008); however, they utilized the traditional nearest matching strategy and did not give a formulized measuring function.

We design a different way to measure the position different, i.e. the n-gram position difference penalty factor, which means our matching is based on n-gram alignment considering neighbour information of the candidate matches. Furthermore, we give clear formulized measuring function.To measure this factor, there are mainly two stages, which include n-gram word alignment and score measuring. Let's see it step by step as below.

The n-gram position difference penalty, $NPosPenal$,is defined as:

$$NPosPenal = e^{-NPD} \tag{3-2}$$

where $NPD$ means $n$-gram position difference penalty. The $NPosPenal$ value is designed to compare the words order in the sentences between reference translation and output translation. The $NPosPenal$ value is normalized. Thus we can take all MT systems into account whose effective$NPD$ value varies between 0 and 1, and when $N$equals 0, the $NPosPenal$ will be 1 which represents no penalty and is quite reasonable. When the $NPD$ increases from 0 to 1, the $NPosPenal$value decreases from 1 to $1/e$, which is based on the mathematical analysis. Consequently, the final LEPOR value will be smaller. According to this thought, the $NPD$ is defined as:

$$NPD = \frac{1}{Length_{output}} \sum_{i=1}^{Length_{output}} |PD_i| \tag{3-3}$$

**Output Sentence**: $W = \{w_1 w_2 w_3 \ldots w_{m_1} | \ m_1 \in (1,\infty)\}$

**Reference Sentence**: $W^r = \{w_1^r w_2^r w_3^r \ldots w_{m_2}^r | \ m_2 \in (1,\infty)\}$

$\forall x \in (1,\infty)$, The Alignment of word $w_x$:

***if*** $\forall y \in (1,\infty): w_x \neq w_y^r$ *// ∀ means for each, ∃ means there is/are*

$(w_x \rightarrow \emptyset)$; *// → shows the alignment*

***elseif*** $\exists!\, y \in (1,\infty): w_x = w_y^r$ *// ∃! means there exists exactly one*

$(w_x \rightarrow w_y^r)$;

***elseif*** $\exists y_1, y_2 \in (1,\infty): (w_x = w_{y_1}^r) \wedge (w_x = w_{y_2}^r)$ *// ∧ is logical conjunction, and*

***foreach*** $k \in (-n,-1) \cup (1,n)$

***foreach*** $j \in (-n,-1) \cup (1,n)$

***if*** $\exists k_1, k_2, j_1, j_2: (w_{x+k_1} = w_{y_1+j_1}^r) \wedge (w_{x+k_2} = w_{y_2+j_2}^r)$

***if*** $Distance(w_x, w_{y_1}^r) \leq Distance(w_x, w_{y_2}^r)$

$(w_x \rightarrow w_{y_1}^r)$;

***else***

$(w_x \rightarrow w_{y_2}^r)$;

***elseif*** $\exists k_1, j_1: (w_{x+k_1} = w_{y_1+j_1}^r) \wedge (\forall k_2, j_2: (w_{x+k_2} \neq w_{y_2+j_2}^r))$

$(w_x \rightarrow w_{y_1}^r)$;

***else*** *// i.e.* $\forall k_1, k_2, j_1, j_2: (w_{x+k_1} \neq w_{y_1+j_1}^r) \wedge (w_{x+k_2} \neq w_{y_2+j_2}^r)$

***if*** $Distance(w_x, w_{y_1}^r) \leq Distance(w_x, w_{y_2}^r)$

$(w_x \rightarrow w_{y_1}^r)$;

***else***

$(w_x \rightarrow w_{y_2}^r)$;

***else*** *// when more than two candidates, the selection steps are similar as above*

**Figure 31: N-gram Word Alignment Algorithm.**

where $Length_{output}$ represents the length of system output sentence and $PD_i$ means the *n*-gram position *D*-value (difference value) of aligned words between output and reference sentences. Every word from both output translation and reference should be aligned only once (one-to-one alignment). Case (upper or lower) is irrelevant. When there is no match, the value of $PD_i$ will be zero as default for this output translation word.

To calculate the NPD value, there are two steps: aligning and calculating. To begin with, the Context-dependent n-gram Word Alignment task: we take the context-dependent factor into consideration and assign higher priority on it, which means we take into account the surrounding context (neighbouring words) of the potential word to select a better matching pairs between the output and the reference. If there are both nearby matching or there is no matched context around the potential words pairs, then we consider the nearest matching to align as a backup choice. The alignment direction is from output sentence to the reference translations. Assuming that $w_x$ represents the current word in output sentence and $w_{x+k}$ (or $w_{x+k_i}$) means the $k$th word to the previous ($k < 0$) or following ($k > 0$). While $w_y^r$ (or $w_{y_i}^r$) means the words matching$w_x$in the references, and $w_{y+j}^r$ (or $w_{y_i+j_i}^r$) has the similar meaning as $w_{x+k}$ but in reference sentence. $Distance$ is the position difference value between the matching words in outputs and references. The operation process and pseudo code of the context-dependent *n*-gram word alignment algorithm are shown in Figure 1 (with "→" as the alignment). Taking 2-gram (*n* = 2) as an example, let's see explanation in Figure 3-1. We label each word with its absolute position, then according to the context-dependent *n*-gram method, the first word "A" in the output sentence has no nearby matching with the beginning word "A" in reference, so it is aligned to the fifth word "a" due to their matched neighbor words "stone"and "on" within one ( $\leq 2$) and two ( $\leq 2$) steps respectively away from current position. Then the fourth word "a" in the output will align the first word "A" of the reference due to the one-to-one alignment.The alignments of other words in the output are obvious.

In the second step (calculating step), we label each word with its position number divided by the corresponding sentence length for normalization, and then using the Eq. (4) to finish the calculation. We also use the example in Figure 6-2 for the $NPD$ introduction:

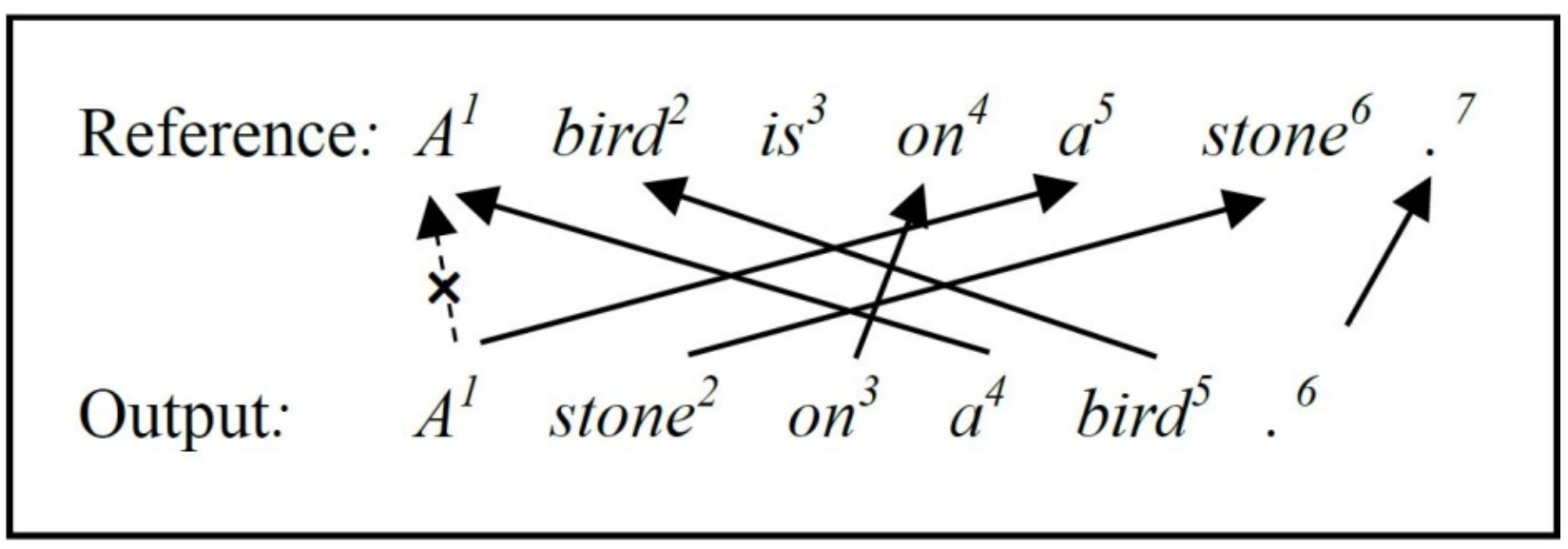

**Figure 32: N-gram Word Alignment Example.**

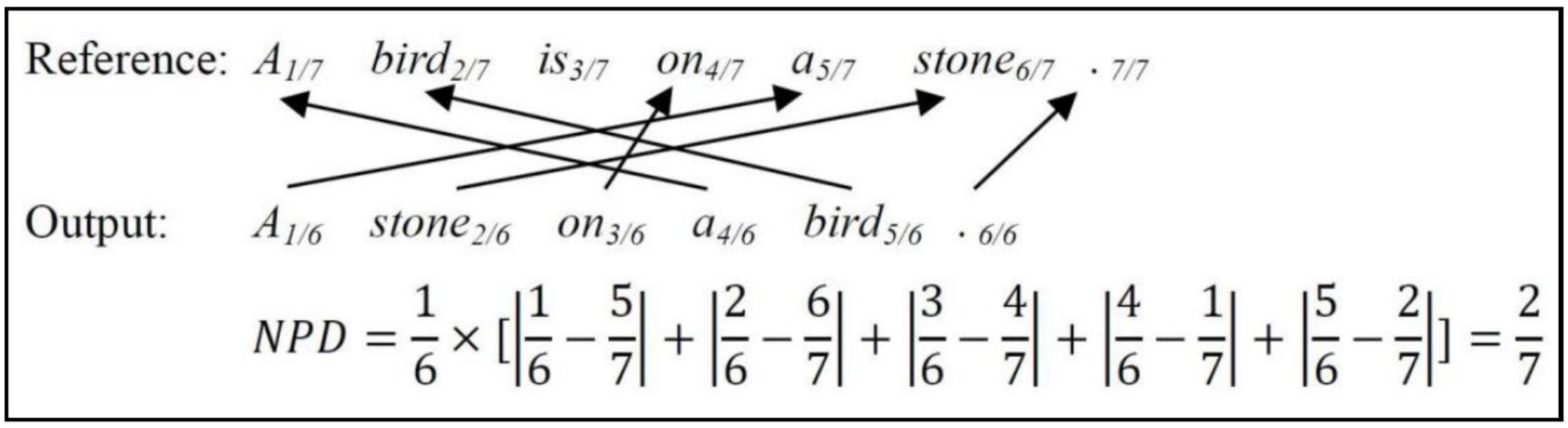

**Figure 33: NPD Calculation Example.**

In the example, when we label the word position of output sentence we divide the numerical position (from 1 to 6) of the current word by the reference sentence length 6. Similar way is applied in labelling the reference sentence. After we get the $NPD$ value, using the Eq. (3), the values of $NPosPenal$ can be calculated.

When there is multi-references (more than one reference sentence), for instance 2 references, we take the similar approach but with a minor change. The alignment direction is reminded the same (from output to reference), and the candidate alignments that have nearby matching words also embrace higher priority. If the matching words from Reference-1 and Reference-2 both have the nearby matching with the output word, then we select the candidate alignment that makes the final $NPD$ value smaller. See below (also 2-gram) for explanation:

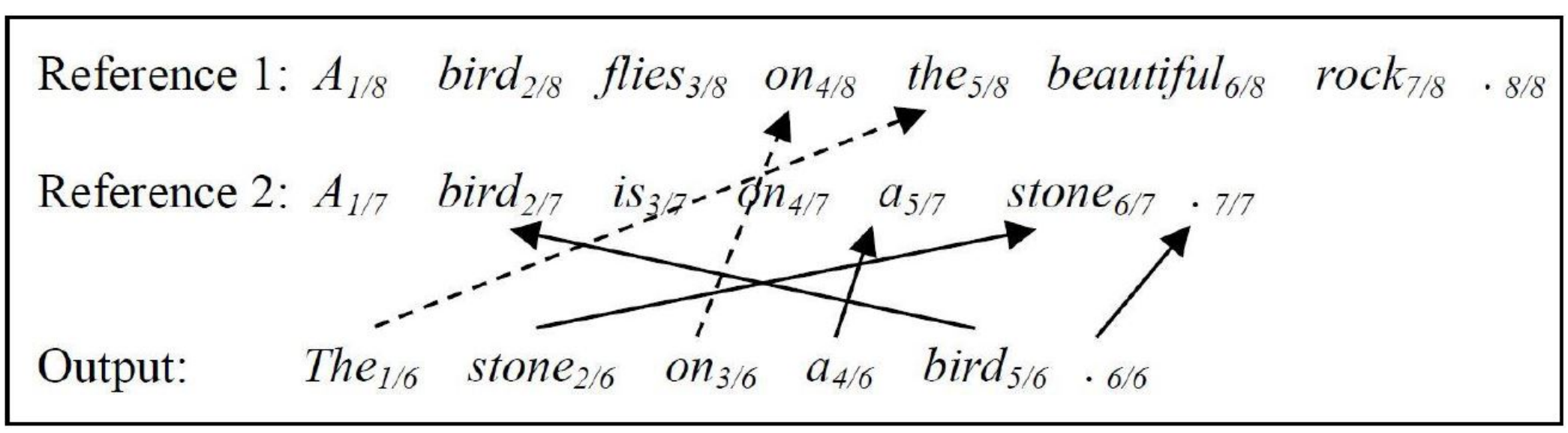

**Figure 34: N-gram Word Alignment Example with Multi-references.**

The beginning output words “the” and “stone” are aligned simply for the single matching. The output word “on” has nearby matching with the word “on” both in Reference-1 and Reference-2, due to the words “the” (second to previous) and “a” (first in the following) respectively. Then we should select its alignment to the word “on” in Reference-1, not Reference-2 for the further reason $\left|\frac{3}{6}-\frac{4}{8}\right| < \left|\frac{3}{6}-\frac{4}{7}\right|$ and this selection will obtain a smaller $NPD$ value. The remaining two words "a" and "bird" in output sentence are aligned using the same principle.

### 3.1.3. Harmonic Mean of Precision and Recall

In the BLEU metric, there is only precision factor without recall. Generally, precision and recall are two parallel factors. Precision reflects the probability of how much the output content is correct, while recall reflects the probability of how much of the answer is included by the output. So, both of the two aspects are important in evaluation. On the other hand, METEOR (Banerjee and Lavie, 2005) puts fixed higher weight on recall as compared with precision score. For different language pairs, the importance of precision and recall differ. To make a generalized factor for wide spread language pairs, we design the tunable parameters for precision and recall, i.e., the weighted harmonic mean of precision and recall.

The weighted harmonic mean of precision and recall ($\alpha R$ and $\beta P$), $Harmonic(\alpha R, \beta P)$ in the equation, is calculated as:

$$Harmonic\left(\alpha R, \beta P\right) = (\alpha + \beta)/(\frac{\alpha}{R} + \frac{\beta}{P}) \tag{3-4}$$

where $\alpha$ and $\beta$ are two parameters we designed to adjust the weight of $R$ (recall) and $P$ (precision). The two metrics are calculated by:

$$P = \frac{common_num}{system_length} \tag{3-5}$$

$$R = \frac{common_num}{reference_length} \tag{3-6}$$

where $common_num$ represents the number of aligned (matching) words and marks appearing both in translations and references, $system_length$ and $reference_length$ specify the sentence length of system output and reference respectively (Melamed et al., 2003).

## 3.2. Metrics Scores of Designed Methods

We name our metric as **LEPOR**, automatic machine translation evaluation metric considering the enhanced **Le**ngth Penalty, **P**recision, n-gram **Po**sition difference Penalty and **R**ecall (Han et al., 2012). To begin with, the sentence level score is the simple product value of each factor.

$$LEPOR = \prod_{i=1}^{3} Factor_i \tag{3-7}$$

Then, we design two strategies to measure the system level (document level) scores. One is the arithmetic mean of each sentence level score, called as $LE\bar{P}OR_A$. The other one $LE\bar{P}OR_B$ is the product value of system level factors score, which means that we first measure the system level factor scores as the arithmetic mean of sentence level factor scores, and then the system level LEPOR metric score is the product value of system level factors.

$$LE\bar{P}OR_A = \frac{1}{SentNum} \sum_{i=1}^{SentNum} LEPOR_{ithSent} \tag{3-8}$$

$$LE\bar{P}OR_B = \prod_{i=1}^{3} Fa\bar{c}tor_i \tag{3-9}$$

$$Fa\bar{c}tor_i = \frac{1}{SentNum} \sum_{j=1}^{SentNum} Factor_{i,\ jthSent} \tag{3-10}$$

In this initial version, the designed system level $LE\bar{P}OR_A$ reflects the system level metric score, and $LE\bar{P}OR_B$ reflects the system level factors score.

# 4. IMPROVED MODELS WITH LINGUISTIC FEATURES

This chapter introduces our improved model of our metrics, such as the new factors, variants of LEPOR, and utilization of concise linguistic features (Han et al., 2013a).

## 4.1. New Factors

To consider more about the content information, we design the new factors including n-gram precision and n-gram recall. These two factors are also measured first at sentence level, which is different with BLEU.

Let's see the n-gram scores. Here, n is the number of words in the block matching.

$$P_n = \frac{\#ngram_{matched}}{\#ngram\ chunks\ in\ system\ output} \quad (4\text{-}1)$$

$$R_n = \frac{\#ngram_{matched}}{\#ngram\ chunks\ in\ reference} \quad (4\text{-}2)$$

$$HPR = Harmonic(\alpha R_n, \beta P_n) = \frac{\alpha + \beta}{\frac{\alpha}{R_n} + \frac{\beta}{P_n}} \quad (4\text{-}3)$$

Let's see an example of bigram matching, and it is the similar strategies for the block matching with $n > 2$.

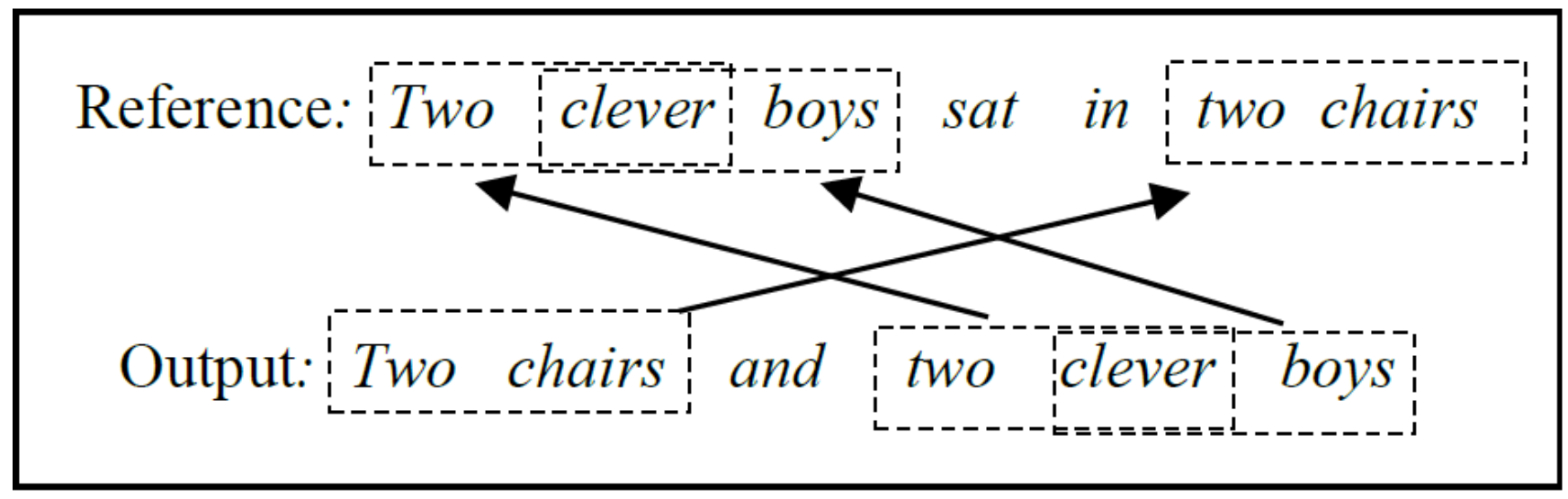

**Figure 41: N-gram Block Alignment Example.**

## 4.2. Variants of LEPOR

This section introduces two variants of LEPOR metric. The first one is based on tunable parameters designed for factor level, and the second one is based on n-gram metric score (Han et al., 2013a; Han et al., 2014).

### 4.2.1. hLEPOR

To achieve higher correlation with manual judgments when dealing with special language pairs, we design tunable parameters to tune the weights of factors. It is achieved by using the weighted harmonic mean again (Han et al. 2013a; 2013e). In this way, we try to seize the important characteristics of focused languages.

Let's see the formula below. The parameters $w_{LP}$, $w_{NPosPenal}$ and $w_{HPR}$ are the weights of three factors respectively.

$$hLEPOR = Harmonic\left(w_{LP}LP, w_{NPosPenal}NPosPenal,\ w_{HPR}HPR\right) \quad (4\text{-}4)$$

$$= \frac{\sum_{i=1}^{n} w_i}{\sum_{i=1}^{n} \frac{w_i}{Factor_i}} = \frac{w_{LP} + w_{NPosPenal} + w_{HPR}}{\frac{w_{LP}}{LP} + \frac{w_{NPosPenal}}{NPosPenal} + \frac{w_{HPR}}{HPR}} \quad (4\text{-}5)$$

In this version, the system-level scores are also measured by the two strategies introduced above. The corresponding formulas are:

$$\overline{hLEPOR}_A = \frac{1}{SentNum} \sum_{i=1}^{SentNum} hLEPOR_{ithSent} \quad (4\text{-}6)$$

$$\overline{hLEPOR}_B = Harmonic(w_{LP}\overline{LP},\ w_{NPosPenal}\overline{NPosPenal},\ w_{HPR}\overline{HPR}) \quad (4\text{-}7)$$

### 4.2.2. nLEPOR

The n-gram metric score is based on the utilization of weighted n-gram precision and n-gram recall factors. Let's see the designed formula, where HPR is measured using weighted n-gram precision and recall formula introduced previously. This variant is designed for the languages that request high fluency.

$$nLEPOR = LP \times NPosPenal \times exp(\sum_{n=1}^{N} w_n logHPR) \quad (4\text{-}8)$$

$$\overline{nLEPOR}_A = \frac{1}{SentNum} \sum_{i=1}^{SentNum} nLEPOR_{ithSent} \quad (4\text{-}9)$$

$$\overline{nLEPOR}_B = \overline{LP} \times \overline{PosPenalty} \times exp(\sum_{n=1}^{N} w_n log\overline{HPR}) \quad (4\text{-}10)$$

## 4.3. Utilization of Linguistic Feature

The linguistic features have been shown very helpful in many previous researches. However, many linguistic features relying on large external data may result in low repeatable. In this section, we investigate the part of speech (POS) information in our model. We first attach the POS tags of each word or token of the sentences by POS tagger or parsing tools, and extract the POS sequence from both the MT output and the reference translations. Then, we apply our algorithms on the POS sequence as the same way on words sequence. In this way, we gain two different kinds of similarity scores, the word level and the POS level. The final metric score will be the weighted combination of these two scores.

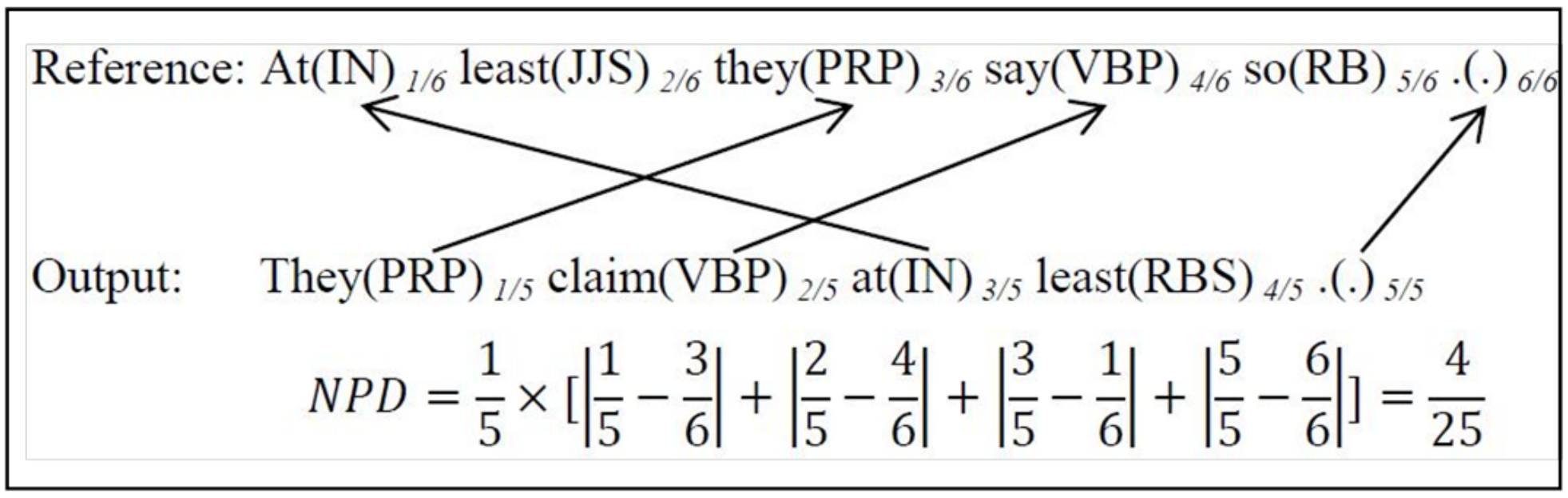

**Figure 42: N-gram POS Sequence Alignment Example.**

Let's see an example with the algorithm applied on POS sequences in the figure.

In this way, some words can be aligned by POS. It sometimes performs as synonym information, e.g. the words "say" and "claim" in the example are successful aligned.

The final scores of our methods using the linguistic features are the combination of word level and linguistic level scores:

$$LEPOR_{liguistic} = \frac{1}{w_{hw} + w_{hp}}(w_{hw}LEPOR_{word} + w_{hp}LEPOR_{POS}) \quad (4\text{-}11)$$

$$LEPO\bar{R}_{linguistic} = \frac{1}{w_{hw} + w_{hp}}(w_{hw}LEP\bar{O}R_{word} + w_{hp}LEP\bar{O}R_{POS}) \quad (4\text{-}12)$$

The $LEPOR_{POS}$ and $LEPOR_{word}$ are measured using the same algorithm on POS sequence and word sequence respectively.

# 5. EVALUATION

This chapter introduces the experimental performance of our designed MT evaluation methods, including the results of initial metric LEPOR and the enhanced model hLEPOR.

## 5.1. Experimental Setting

We first introduce the corpora preparation, the selected state-of-the-art metrics to compare, and the evaluation criteria.

### 5.1.1. Corpora

We utilize the standard WMT shared task corpora in our experiments. For the development set, we use the WMT 2008 corpora. The development corpora are to tune the parameters in our metrics to achieve a higher correlation with manual judgments. For the testing corpora, we use the WMT 2011 corpora.

Both the WMT 2008[2] and WMT 2011[3] corpora contain the language pairs of English (EN) to other (ES: Spanish, DE: German, FR: French and CS: Czech) and the inverse translation direction, i.e. other to English.

There are 2,028 and 3,003 sentences respectively for each language document in the WMT 2008 and WMT 2011 MT testing corpora. The effective number of participated MT systems in WMT 2011 for each language pair is shown in the Table 5-1.

**Table 51: Participated MT Systems in WMT 2011.**

| English-to-Other | | Other-to-English | |
|---|---|---|---|
| Language pair | MT systems | Language pair | MT systems |
| EN-ES | 15 | ES-EN | 15 |

[2]http://www.statmt.org/wmt08/

[3]http://www.statmt.org/wmt11/

| EN-FR | 17 | FR-EN | 18 |
|---|---|---|---|
| EN-DE | 22 | DE-EN | 20 |
| EN-CS | 10 | CS-EN | 8 |

### 5.1.2. Existing Metrics for Comparison

To compare with our initial version metric, the LEPOR, we selected three state-of-the-art metrics as comparisons, including precision based metric BLEU(Papineni et al., 2002), edit distance based metric TER (Snover et al., 2006), and METEOR (version 1.3) (Denkowski and Lavie, 2011), which used synonym and stemming as external linguistic features. See Section 2.2 for detailed introduction of the metrics. Furthermore, we also selected two latest metrics the AMBER and MP4IBM1 as comparisons. AMBER (Chen and Kuhn, 2011) is a modified version of BLEU, attaching more kinds of penalty coefficients and combining the n-gram precision and recall. MP4IBM1 (Popovic et al., 2011) is based on morphemes, POS (4-grams) and lexicon probabilities, etc.

To investigate the performance of our improved metric hLEPOR, i.e. the metric with more tunable parameters and concise POS as linguistic feature, we added two more metrics in the comparison list including ROSE and MPF. This is due to the fact that both ROSE (Song and Cohn, 2011) and MPF (Popovic, 2011) metrics also utilized the POS information as linguistic feature and they are very related with our work.

### 5.1.3. Evaluation Criteria

Each year, there is a manual judgment task in the WMT after the participants submitted their MT outputs, and we regard the manual judgments as the golden one.

In this light, we measure the correlation score between manual judgments and automatic evaluations. The evaluation criterion we utilized is the widely used system level spearman correlation score. See Section 2.3, the evaluation criteria for MT evaluation, for detailed formula.

## 5.2. Experimental Results

This section introduces the experiments results of the MT evaluation, including the results and initial metric and the results of improved metric.

### 5.2.1. Initial MT Evaluation Results

The MT evaluation results using our initial metric LEPOR are demonstrated in Table 5-2, the correlation score with human judgments. The $LEPOR_A$ is the arithmetic mean of sentence level score, and $LEPOR_B$ is the product value of system level factor scores. Please see Section 3.2, i.e. the metric scores, for the detailed metric formula. The parameters $\alpha$ and $\beta$ are tuned to be the value 9 and 1 respectively for all the language pairs, except for Czech-to-English with the value 1 and 9.

**Table 52: Spearman Correlation Scores of LEPOR and Others.**

| **Evaluation system** | **Correlation Score with Human Judgment** | | | | | | | | |
|---|---|---|---|---|---|---|---|---|---|
| | other-to-English | | | | English-to-other | | | | Mean Score |
| | CZ-EN | DE-EN | ES-EN | FR-EN | EN-CZ | EN-DE | EN-ES | EN-FR | |
| LEPOR-B | 0.93 | 0.62 | **0.96** | 0.89 | 0.71 | 0.36 | 0.88 | 0.84 | **0.77** |
| LEPOR-A | **0.95** | 0.61 | 0.96 | 0.88 | 0.68 | 0.35 | **0.89** | 0.83 | **0.77** |
| AMBER | 0.88 | 0.59 | 0.86 | **0.95** | 0.56 | 0.53 | 0.87 | 0.84 | 0.76 |
| Meteor-1.3-RANK | 0.91 | **0.71** | 0.88 | 0.93 | 0.65 | 0.30 | 0.74 | 0.85 | 0.75 |
| BLEU | 0.88 | 0.48 | 0.90 | 0.85 | 0.65 | 0.44 | 0.87 | **0.86** | 0.74 |
| TER | 0.83 | 0.33 | 0.89 | 0.77 | 0.50 | 0.12 | 0.81 | 0.84 | 0.64 |
| MP4IBM1 | 0.91 | 0.56 | 0.12 | 0.08 | **0.76** | **0.91** | 0.71 | 0.61 | 0.58 |

The metrics are ranked by their mean (hybrid) performance on the eight corpora from the best to the worst. It shows that BLEU, AMBER (modified version of BLEU) and Meteor-1.3 perform unsteady with better correlation on some translation languages and worse on others, resulting in medium level generally. TER and MP4IBM1 get the worst scores by the mean correlation.

The evaluation results also demonstrate that the first simplified version of our metric without using external resources yielded three top-one correlation scores on CZ-EN / ES-EN / EN-ES language pairs. Furthermore, LEPOR showed robust performance across languages, resulting in top one Mean-score 0.77.

It also releases the information that although the test metrics yield high system-level correlations with human judgments on certain language pairs, e.g. all correlations above 0.83 on Czech-to-English, they are far from satisfactory by synthetically mean scores on total eight corpora, spanning from 0.58 to 0.77 only, and there is clearly a potential for further improvement.

### 5.2.2. MT Evaluation Results with Improved Methods

In this improved version hLEPOR, the metric based on factor level weighted harmonic mean, with concise linguistic feature, the tuned values of many parameters on the development set are shown in Table 5-3.In the parameters table, the token "(W)" and "(POS)" mean this set of parameters are on the word level and extracted POS level respectively. The ratio "HPR:ELP:NPP" represents the different weights of three main factors in our metric, i.e. the harmonic mean of precision and recall, the enhanced length penalty, and the n-gram position difference penalty. The ratio "$\alpha:\beta$" means the weights of recall and precision. The ratio "$w_{hw}:w_{hp}$" represents the different weights of word level score and the POS level score. The token "N/A" means the POS information was not utilized on that language pair, so there is only word level score. The testing results, correlation score with manual judgments, are demonstrated in Table 5-4.

The evaluation results using correlation score with manual judgments demonstrate that our enhanced model hLEPOR yielded the highest score on the language pair German-to-English and higher scores on other language pairs. Our initial metric LEPOR remains the highest score on Spanish-to-English and Czech-to-English language pairs. The MPF and ROSE metrics achieved the highest scores on English-to-French and English-to-Spanish respectively. However, in the overall performance, our improved model hLEPOR reached the top one level with the mean score 0.83, which is much higher than the initial version LEPOR with 0.77 score.

**Table 53: Tuned Parameters of hLEPOR Metric.**

| **Ratio** | **Other-to-English** | | | | **English-to-Other** | | | |
|---|---|---|---|---|---|---|---|---|
| | CZ-EN | DE-EN | ES-EN | FR-EN | EN-CZ | EN-DE | EN-ES | EN-FR |
| HPR:ELP:NPP(W) | 7:2:1 | 3:2:1 | 7:2:1 | 3:2:1 | 3:2:1 | 1:3:7 | 3:2:1 | 3:2:1 |
| HPR:ELP:NPP(POS) | N/A | 3:2:1 | N/A | 3:2:1 | N/A | 7:2:1 | N/A | 3:2:1 |
| $\alpha:\beta$(W) | 1:9 | 9:1 | 1:9 | 9:1 | 9:1 | 9:1 | 9:1 | 9:1 |
| $\alpha:\beta$(POS) | N/A | 9:1 | N/A | 9:1 | N/A | 9:1 | N/A | 9:1 |
| $w_{hw}:w_{hp}$ | N/A | 1:9 | N/A | 9:1 | N/A | 1:9 | N/A | 9:1 |

**Table 54: Spearman Correlation Scores of hLEPOR and Others.**

| **Evaluation system** | **Correlation Score with Human Judgment** | | | | | | | | |
|---|---|---|---|---|---|---|---|---|---|
| | other-to-English | | | | English-to-other | | | | Mean Score |
| | CZ-EN | DE-EN | ES-EN | FR-EN | EN-CZ | EN-DE | EN-ES | EN-FR | |
| hLEPOR | 0.93 | **0.86** | 0.88 | 0.92 | 0.56 | 0.82 | 0.85 | 0.83 | **0.83** |
| MPF | 0.95 | 0.69 | 0.83 | 0.87 | 0.72 | 0.63 | 0.87 | **0.89** | 0.81 |
| LEPOR-B | 0.93 | 0.62 | **0.96** | 0.89 | 0.71 | 0.36 | 0.88 | 0.84 | 0.77 |
| LEPOR-A | **0.95** | 0.61 | **0.96** | 0.88 | 0.68 | 0.35 | 0.89 | 0.83 | 0.77 |
| ROSE | 0.88 | 0.59 | 0.92 | 0.86 | 0.65 | 0.41 | **0.90** | 0.86 | 0.76 |
| AMBER | 0.88 | 0.59 | 0.86 | **0.95** | 0.56 | 0.53 | 0.87 | 0.84 | 0.76 |
| Meteor-1.3-RANK | 0.91 | 0.71 | 0.88 | 0.93 | 0.65 | 0.30 | 0.74 | 0.85 | 0.75 |
| BLEU | 0.88 | 0.48 | 0.90 | 0.85 | 0.65 | 0.44 | 0.87 | 0.86 | 0.74 |
| TER | 0.83 | 0.33 | 0.89 | 0.77 | 0.50 | 0.12 | 0.81 | 0.84 | 0.64 |
| MP4IBM1 | 0.91 | 0.56 | 0.12 | 0.08 | **0.76** | **0.91** | 0.71 | 0.61 | 0.58 |

# 6. EVALUATION ON ACL-WMT SHARED TASK

This chapter introduces our participation in the shared tasks of WMT 2013, the Eighth International Workshop of Statistical Machine Translation accompanied with ACL conference, including our submitted metrics and the official evaluation results (Han et al., 2013b).

## 6.1. Task Introduction in WMT 2013

In the WMT 2013, there are mainly three kinds of shared tasks, i.e. the MT task, the MT evaluation task, and the quality estimation (QE) task. This section introduces our participation in the MT evaluation task. The MT evaluation task is to evaluate the translation qualities of submitted MT systems, including manual judgments and automatic MT evaluation.

**Table 6-1: Participated MT Systems in WMT 2013.**

| English-to-Other | | Other-to-English | |
|---|---|---|---|
| Language pair | MT systems | Language pair | MT systems |
| EN-ES | 13 | ES-EN | 12 |
| EN-FR | 17 | FR-EN | 13 |
| EN-DE | 15 | DE-EN | 17 |
| EN-CS | 12 | CS-EN | 11 |
| EN-RU | 14 | RU-EN | 19 |

In addition to the traditional language corpora, i.e. English, Spanish, German, French, and Czech, there is one new language Russian participated in the WMT 2013; thus, there are two new language pairs in translation, the English-to-Russian and Russian-to-English. In these two newly added language pairs, there are not very many training or developing data.

For each language pair, the evaluation task is to evaluate the translation quality of one single document that contains 3,000 sentences. The participated MT systems in each language pair are shown in Table 6-1 (Bojar et al., 2013). The evaluation criteria for the automatic MT emulation metrics are the correlation score with human judgements, including Spearman, Pearson and Kendall's tau. See Section 2.3 for the detailed evaluation criteria.

## 6.2. Submitted Methods

We submitted two versions of our methods to the shared tasks in WMT-2013. The submitted metrics are LEPOR_v3.1 and nLEPOR_baseline. The LEPOR_v3.1 is actually the $h\bar{LEPOR}_{linguistic}$ metric and the nLEPOR_baseline is the n-gram based $n\bar{LEPOR}$ metric with default parameter values. See Section 4.2 "variants of LEPOR" for detailed formula introduction. The parameters in the hLEPOR metric utilized are the same set as in the last section, testing for WMT-2011 corpora. For the nLEPOR_baseline metric, we utilized the unigram harmonic mean in the factor HPR.The parameters $\alpha$ and $\beta$ in nLEPOR are tuned to be the value 9 and 1 respectively for all the language pairs, except for Czech-to-English with the value 1 and 9.

## 6.3. Official Evaluation Results

There are 18 and 20 effective automatic MT evaluation metrics for the English-to-Other and Other-to-English translation directions respectively. Some of the metrics only performed on single direction, such as English-to-other or other-to-English. For instance, the UMEANT and Deprif metrics only submitted evaluation results for other-to-English direction; and the ACTa only submitted the evaluation results for English-to-French and English-to-German. However, we submitted our evaluation results on both translation directions.

### 6.3.1. System-level Evaluation Results

This section introduces our metrics performances in system level correlation scores with manual judgments. We firstly show the English-to-other direction and then the other-to-English direction.

**Table 6-2: System-level Pearson Correlation Scores.**

| Directions | EN-FR | EN-DE | EN-ES | EN-CS | EN-RU | Av |
|---|---|---|---|---|---|---|
| *LEPOR_v3.1* | *.91* | *.94* | *.91* | *.76* | ***.77*** | ***.86*** |
| *nLEPOR_baseline* | *.92* | *.92* | *.90* | ***.82*** | *.68* | *.85* |
| SIMPBLEU_RECALL | **.95** | .93 | .90 | **.82** | .63 | .84 |
| SIMPBLEU_PREC | .94 | .90 | .89 | **.82** | .65 | .84 |
| NIST-mteval-inter | .91 | .83 | .84 | .79 | .68 | .81 |
| Meteor | .91 | .88 | .88 | **.82** | .55 | .81 |
| BLEU-mteval-inter | .89 | .84 | .88 | .81 | .61 | .80 |
| BLEU-moses | .90 | .82 | .88 | .80 | .62 | .80 |
| BLEU-mteval | .90 | .82 | .87 | .80 | .62 | .80 |
| CDER-moses | .91 | .82 | .88 | .74 | .63 | .80 |
| NIST-mteval | .91 | .79 | .83 | .78 | .68 | .79 |
| PER-moses | .88 | .65 | .88 | .76 | .62 | .76 |
| TER-moses | .91 | .73 | .78 | .70 | .61 | .75 |
| WER-moses | .92 | .69 | .77 | .70 | .61 | .74 |
| TerrorCat | .94 | **.96** | **.95** | na | na | .95 |
| SEMPOS | na | na | na | .72 | na | .72 |
| ACTa | .81 | -.47 | na | na | na | .17 |
| ACTa5+6 | .81 | -.47 | na | na | na | .17 |

#### 6.3.1.1. The Official Results of English-to-other MT Evaluation

Table 6-2 and Table 6-3 demonstrate the official results using Pearson and Spearman correlation scores respectively. Table 6-2 shows LEPOR_v3.1 and nLEPOR_baseline

yield the highest and the second highest average Pearson correlation score 0.86 and 0.85 respectively with human judgments at system-level on five English-to-other language pairs. LEPOR_v3.1 and nLEPOR_baseline also yield the highest Pearson correlation score on English-to-Russian (0.77) and English-to-Czech (0.82) language pairs respectively. The testing results of LEPOR_v3.1 and nLEPOR_baseline show better correlation scores as compared to METEOR (0.81), BLEU (0.80) and TER-moses (0.75) on English-to-other language pairs, which is similar with the training results. On the other hand, using the Spearman rank correlation coefficient, SIMPBLEU_RECALL yields the highest correlation score 0.85 with human judgments. Our metric LEPOR_v3.1 also yields the highest Spearman correlation score on English-to-Russian (0.85) language pair, which is similar with the result using Pearson correlation and shows its robust performance on this language pair.

#### 6.3.1.2. The Official Results of other-to-English MT Evaluation

Table 6-4 and Table 6-5 demonstrate the official results using Pearson and Spearman correlation scores respectively for other-to-English translation direction.

METEOR yields the highest average correlation scores 0.95 and 0.94 respectively using Pearson and Spearman rank correlation methods on other-to-English language pairs. The average performance of nLEPOR_baseline is a little better than LEPOR_v3.1 on the five language pairs of other-to-English even though it is also moder-ate as compared to other metrics. However, using the Pearson correlation method, nLEPOR_baseline yields the average correlation score 0.87 which already wins the BLEU (0.86) and TER (0.80) as shown in Table 6-4.

### 6.3.2. Segment-level MT Evaluation Results

In addition to the system level MT evaluation, our metric can also be utilized for the segment level MT evaluation; because our metrics first measure the sentence score and then the document score. However, many automatic MT evaluation metrics can only perform on system level (document score), thus, the participated automatic MT evaluation metrics for segment level strategy were much fewer than the system level. This section introduces our performance in the segment level MT evaluation. The evaluation criterion is the Kendall's tau.

**Table 6-3: System-level Spearman Correlation Scores.**

| Directions | EN-FR | EN-DE | EN-ES | EN-CS | EN-RU | Av |
|---|---|---|---|---|---|---|
| SIMPBLEU_RECALL | .92 | .93 | .83 | .87 | .71 | **.85** |
| *LEPOR_v3.1* | *.90* | *.9* | *.84* | *.75* | ***.85*** | ***.85*** |
| NIST-mteval-inter | **.93** | .85 | .80 | .90 | .77 | **.85** |
| CDER-moses | .92 | .87 | .86 | .89 | .70 | **.85** |
| *nLEPOR_baseline* | *.92* | *.90* | *.85* | *.82* | *.73* | *.84* |
| NIST-mteval | .91 | .83 | .78 | .92 | .72 | .83 |
| SIMPBLEU_PREC | .91 | .88 | .78 | .88 | .70 | .83 |
| Meteor | .92 | .88 | .78 | **.94** | .57 | .82 |
| BLEU-mteval-inter | .92 | .83 | .76 | .90 | .66 | .81 |
| BLEU-mteval | .89 | .79 | .76 | .90 | .63 | .79 |
| TER-moses | .91 | .85 | .75 | .86 | .54 | .78 |
| BLEU-moses | .90 | .79 | .76 | .90 | .57 | .78 |
| WER-moses | .91 | .83 | .71 | .86 | .55 | .77 |
| PER-moses | .87 | .69 | .77 | .80 | .59 | .74 |
| TerrorCat | **.93** | **.95** | **.91** | na | na | .93 |
| SEMPOS | na | Na | na | .70 | na | .70 |
| ACTa5+6 | .81 | -.53 | na | na | na | .14 |
| ACTa | .81 | -.53 | na | na | na | .14 |

**Table 64: System-level Pearson Correlation on other-to-English Language Pairs.**

| Directions | FR-EN | DE-EN | ES-EN | CS-EN | RU-EN | Av |
|---|---|---|---|---|---|---|
| Meteor | **.98** | .96 | .97 | **.99** | **.84** | **.95** |
| SEMPOS | .95 | .95 | .96 | **.99** | .82 | .93 |
| Depref-align | .97 | .97 | .97 | .98 | .74 | .93 |
| Depref-exact | .97 | .97 | .96 | .98 | .73 | .92 |
| SIMPBLEU_RECALL | .97 | .97 | .96 | .94 | .78 | .92 |
| UMEANT | .96 | .97 | **.99** | .97 | .66 | .91 |
| MEANT | .96 | .96 | **.99** | .96 | .63 | .90 |
| CDER-moses | .96 | .91 | .95 | .90 | .66 | .88 |
| SIMPBLEU_PREC | .95 | .92 | .95 | .91 | .61 | .87 |
| *LEPOR_v3.1* | *.96* | *.96* | *.90* | *.81* | *.71* | *.87* |
| *nLEPOR_baseline* | *.96* | *.94* | *.94* | *.80* | *.69* | *.87* |
| BLEU-mteval-inter | .95 | .92 | .94 | .90 | .61 | .86 |
| NIST-mteval-inter | .94 | .91 | .93 | .84 | .66 | .86 |
| BLEU-moses | .94 | .91 | .94 | .89 | .60 | .86 |
| BLEU-mteval | .95 | .90 | .94 | .88 | .60 | .85 |
| NIST-mteval | .94 | .90 | .93 | .84 | .65 | .85 |
| TER-moses | .93 | .87 | .91 | .77 | .52 | .80 |
| WER-moses | .93 | .84 | .89 | .76 | .50 | .78 |
| PER-moses | .84 | .88 | .87 | .74 | .45 | .76 |
| TerrorCat | **.98** | **.98** | .97 | na | na | .98 |

**Table 6-5: System-level Spearman Correlation on other-to-English Language Pairs.**

| Directions | FR-EN | DE-EN | ES-EN | CS-EN | RU-EN | Av |
|---|---|---|---|---|---|---|
| Meteor | .98 | .96 | **.98** | .96 | .81 | **.94** |
| Depref-align | **.99** | **.97** | .97 | .96 | .79 | **.94** |
| UMEANT | **.99** | .95 | .96 | **.97** | .79 | .93 |
| MEANT | .97 | .93 | .94 | **.97** | .78 | .92 |
| Depref-exact | .98 | .96 | .94 | .94 | .76 | .92 |
| SEMPOS | .94 | .92 | .93 | .95 | **.83** | .91 |
| SIMPBLEU_RECALL | .98 | .94 | .92 | .91 | .81 | .91 |
| BLEU-mteval-inter | **.99** | .90 | .90 | .94 | .72 | .89 |
| BLEU-mteval | **.99** | .89 | .89 | .94 | .69 | .88 |
| BLEU-moses | **.99** | .90 | .88 | .94 | .67 | .88 |
| CDER-moses | **.99** | .88 | .89 | .93 | .69 | .87 |
| SIMPBLEU_PREC | **.99** | .85 | .83 | .92 | .72 | .86 |
| *nLEPOR_baseline* | *.95* | *.95* | *.83* | *.85* | *.72* | *.86* |
| *LEPOR_v3.1* | *.95* | *.93* | *.75* | *0.8* | *.79* | *.84* |
| NIST-mteval | .95 | .88 | .77 | .89 | .66 | .83 |
| NIST-mteval-inter | .95 | .88 | .76 | .88 | .68 | .83 |
| TER-moses | .95 | .83 | .83 | 0.8 | 0.6 | 0.80 |
| WER-moses | .95 | .67 | .80 | .75 | .61 | .76 |
| PER-moses | .85 | .86 | .36 | .70 | .67 | .69 |
| TerrorCat | .98 | .96 | .97 | na | na | .97 |

From the Table 6-6 and Table 6-7, the overall segment-level performance of LEPOR is moderate with the average Kendall's tau correlation score 0.10 and 0.19 respectively on English-to-other and other-to-English directions. This is due to the fact that we trained our metrics at system-level in this shared metrics task. The segment level evaluation scores are actually the bonuses of our participated metrics.

**Table 6-6: Segment-level Kendall's tau Correlation scores on WMT13 English-to-other Language Pairs.**

| Directions | EN-FR | EN-DE | EN-ES | EN-CS | EN-RU | Av |
|---|---|---|---|---|---|---|
| SIMPBLEU_RECALL | **.16** | **.09** | **.23** | **.06** | **.12** | **.13** |
| Meteor | .15 | .05 | .18 | **.06** | .11 | .11 |
| SIMPBLEU_PREC | .14 | .07 | .19 | **.06** | .09 | .11 |
| sentBLEU-moses | .13 | .05 | .17 | .05 | .09 | .10 |
| *LEPOR_v3.1* | *.13* | *.06* | *.18* | *.02* | *.11* | *.10* |
| *nLEPOR_baseline* | *.12* | *.05* | *.16* | *.05* | *.10* | *.10* |
| dfki_logregNorm-411 | na | na | .14 | na | na | .14 |
| TerrorCat | .12 | .07 | .19 | na | na | .13 |
| dfki_logregNormSoft-431 | na | na | .03 | na | na | .03 |

**Table 6-7: Segment-level Kendall's tau Correlation scores on WMT13 other-to-English Language Pairs.**

| Directions | FR-EN | DE-EN | ES-EN | CS-EN | RU-EN | Av |
|---|---|---|---|---|---|---|
| SIMPBLEU_RECALL | **.19** | **.32** | **.28** | .26 | .23 | **.26** |
| Meteor | .18 | .29 | .24 | **.27** | **.24** | .24 |
| Depref-align | .16 | .27 | .23 | .23 | .20 | .22 |
| Depref-exact | .17 | .26 | .23 | .23 | .19 | .22 |
| SIMPBLEU_PREC | .15 | .24 | .21 | .21 | .17 | .20 |
| *nLEPOR_baseline* | *.15* | *.24* | *.20* | *.18* | *.17* | *.19* |
| sentBLEU-moses | .15 | .22 | .20 | .20 | .17 | .19 |
| *LEPOR_v3.1* | *.15* | *.22* | *.16* | *.19* | *.18* | *.18* |
| UMEANT | .10 | .17 | .14 | .16 | .11 | .14 |
| MEANT | .10 | .16 | .14 | .16 | .11 | .14 |
| dfki_logregFSS-33 | Na | .27 | na | na | Na | .27 |
| dfki_logregFSS-24 | Na | .27 | na | na | na | .27 |
| TerrorCat | .16 | .30 | .23 | na | na | .23 |

# 7. QUALITY ESTIMATION OF MT

This chapter introduces the advanced technology in MT evaluation, which is usually called as Quality Estimation (QE) of MT. The Quality Estimation tasks make some differences from the traditional evaluation, such as extracting reference-independent features from input sentences and the translation, obtaining quality score based on models produced from training data, predicting the quality of an unseen translated text at system run-time, filtering out sentences which are not good enough for post processing, and selecting the best translation among multiple systems, etc. In this chapter, we firstly introduce some QE methods without using reference translations. Then, we introduce the latest QE tasks, and finally, it is our proposed methods in the QE research area.

## 7.1. Quality Estimation without using Reference Translations

In recent years, some MT evaluation methods that do not use the manually offered golden reference translations are proposed. The unsupervised MT evaluation is usually called as Quality Estimation. Some of the related works have been mentioned in previous sections. We introduce some works that have not been discussed in previous sections.

Gamon et al. (2005) investigate the possibility of evaluating MT quality and fluency at the sentence level without using the reference translations. Their system can also perform as a classifier to identify the worst-translated (highly dysfluent and ill-formed) sentences. The SVM classifier is used with linguistic features such as the trigram part-of-speech tags, context-free grammar productions (the phrase sequences), semantic analysis features, and semantic modification relations, etc. The experiment on English-to-French corpus show that the described methods achieve lower correlation score with human judgments as compared to the BLEU metric, which uses the reference translations. However, when formulated as a classification task for identifying the worst-translated sentences, the combination of language model and SVM scores outperforms BLEU.

The traditional metrics BLEU and NIST are known to have good correlation with human evaluation at the corpus level, but this is not the case at the segment level. Specia et al. (2010) address the problem of evaluating the quality of MT as a prediction task, where reference-independent features are extracted from the input sentences and their translation, and a quality score is obtained based on models produced from training data. They show that this approach yields better correlation at segment-level with human evaluation as compared to commonly used metrics, even with models trained on different MT systems, language-pairs and text domains.

Suzuki (2011) develops a post-editing system based on phrase-based SMT (Moses) and applies it into a sentence-level automatic quality evaluator for machine translation in the absence of reference translations. The 28 features they used are from the partial least squares regression analysis on translation sentences (without the using of source sentences) such as 2-gram and 3-gram language model probability, 2-gram and 3-gram backward language model probability, POS 2-gram and 3-gram language model probability, noun phrase and verb phrase from grammar parser, etc. The experiment on Japanese-to-English patent translation, using the criteria adequacy and fluency, shows the validity of the designed methods.

Mehdad et al. (2012) treat the MT evaluation as a cross-lingual textual entailment problem, and design the evaluation focusing on adequacy, semantic equivalence between source sentence and target translation, without using reference translations. The designed method is built on the advances in cross-lingual textual entailment recognition. They use support vector machine to learn models for classification and regression with a linear kernel and default parameters setting. The performances are carried out on English-Spanish language pairs. The large feature set is from the part-of-speech tagger, dependency parsers and named entity recognizers, etc. The used features include Surface Form, number of words, punctuation and the ratios, etc.; Shallow Syntactic, ratios of POS tags in source and target; Syntactic, number and ratios of dependency roles; Phrase Table, lexical phrases extracted from bilingual parallel corpus; Dependency Relation, syntactic constraints; and Semantic Phrase Table, named entity, etc. The experiment on WMT 07 corpora shows higher correlation score with human adequacy annotation than METEOR but lower than BLEU and TER.

Other related works about evaluation without using reference translations include (Blatz et al., 2004) and (Quirk, 2004) that perform the early attempt to evaluate the translation quality avoiding reference translations by the utilizing of large number of source, target, and system-dependent features to discriminate the "good" and "bad" translations; Albrecht and Hwa (2007) utilize the regression method and pseudo references; Specia and Gimenez (2010) combine the confidence estimation and reference-based metrics together for the segment-level MT evaluation; Popović et al. (2011) perform the MT evaluation using the IBM model-1 with the information of morphemes, 4-gram POS and lexicon probabilities; Avramidis (2012) performs an automatic sentence-level ranking of multiple machine translation using the features of verbs, nouns, sentences, subordinate clauses and punctuation occurrences to derive the adequacy information. The detailed introduction and descriptions of the MT Quality Estimation tasks can be reached in (Callison-Burch et al., 2012) and (Felice and Specia, 2012).

## 7.2. Latest QE Tasks

The latest quality estimation tasks of MT can be found in WMT12 (Callison-Burch et al., 2012) and WMT13[4]. For the ranking task, they defined a novel task evaluation metric that provides some advantages over the traditional ranking metrics. The designed criterion DeltaAvg assumes that the reference test set has a number associated with each entry that represents its extrinsic value. For instance, using the effort scale, they associate a value between 1 and 5 with each sentence, representing the quality of that sentence. Given these values, their metric does not need an explicit reference ranking, the way the Spearman ranking correlation does. The goal of the DeltaAvg metric is to measure how valuable a proposed ranking (hypothesis ranking) is according to the extrinsic values associated with the test entries.

$$DeltaAvg_V[n] = \frac{\sum_{k=1}^{n-1} V(S_{1,k})}{n-1} - V(S) \quad (7\text{-}1)$$

[4]http://www.statmt.org/wmt13/quality-estimation-task.html

For the scoring task, they usetwo task evaluation metrics that have been traditionally used for measuring performance for regression tasks: Mean Absolute Error (MAE) as a primary metric, and Root of Mean Squared Error (RMSE) as a secondary metric. For a given test set $S$ with entries $s_i$, $1 \leq i \leq |S|$, they denote by $H(s_i)$ the proposed score for entry $s_i$ (hypothesis), and by $V(s_i)$ the reference value for entry $s_i$ (gold-standard value). They formally define the metrics as follows.

$$MAE = \frac{\sum_{i=1}^{N} |H(s_i) - V(s_i)|}{N} \tag{7-2}$$

$$RMSE = \sqrt{\frac{\sum_{i=1}^{N} \left(H(s_i) - V(s_i)\right)^2}{N}} \tag{7-3}$$

where $N = |S|$. Both these metrics are nonparametric, automatic and deterministic (and therefore consistent), and extrinsically interpretable.

There appear to be significant differences between considering the quality estimation task as a ranking problem versus a scoring problem. The ranking based approach appears to be somewhat simpler and more easily amenable to automatic solutions, and at the same time provides immediate benefits when integrated into larger applications, for instance, the post-editing application described in (Specia, 2011). The scoring-based approach is more difficult, as the high error rate even of oracle-based solutions indicates. It is also well-known from human evaluations of MT outputs that human judges also have a difficult time agreeing on absolute-number judgements to translations. The experiences in creating the current datasets confirm that, even with highly-trained professionals, it is difficult to arrive at consistent judgements. The WMT tasks plan to have future investigations on how to achieve more consistent ways of generating absolute-number scores that reflect the quality of automated translations.

## 7.3. Proposed Methods in the Advanced QE

Firstly, we introduce our methods in the WMT-13 QE tasks (Han et al., 2013c). Then, we introduce our research works out side of the task.

### 7.3.1. Proposed Methods in WMT-13 QE Task

In the WMT-2013 QE shared task (Bojar et al., 2013), we participated the tasks of sentence-level English-to-Spanish QE, system selection for English-to-Spanish and English-to-German translation, and the word-level QE for binary and multi-class error classification.

**Table 71: Developed POS Mapping for Spanish and Universal Tagset.**

| ADJ | ADP | ADV | CONJ | DET | NOUN | NUM | PRON | PRT | VERB | X | . |
|---|---|---|---|---|---|---|---|---|---|---|---|
| ADJ | PREP, PREP/DEL | ADV, NEG | CC, CCAD, CCNEG, CQUE, CSUBF, CSUBI, CSUBX | ART | NC, NMEA, NMON, NP, PERCT, UMMX | CARD, CODE, QU | DM, INT, PPC, PPO, PPX, REL | SE | VCLIger, VCLIinf, VCLIfin, VEadj, VEfin, VEger, VEinf, VHadj, VHfin, VHger, VHinf, VLadj, VLfin, VLger, VLinf, VMadj, VMfin, VMger, VMinf, VSadj, VSfin, VSger, VSinf | ACRNM, ALFP, ALFS, FO, ITJN, ORD, PAL, PDEL, PE, PNC, SYM | BACKSLASH, CM, COLON, DASH, DOTS, FS, LP, QT, RP, SEMICOLON, SLASH |

For the sentence-level EN-ES QE task, we designed the English and Spanish POS tagset mapping as shown in Table 7-1. The 75 Spanish POS tags yielded by the Treetagger (Schmid, 1994) are mapped to the 12 universal tags developed in (Petrov et al., 2012). Furthermore, we designed a novel evaluation method, the enhanced BLEU (EBLEU) as bellow:

$$EBLEU = 1 - MLP \times exp\Big(\sum w_n log\big(H(\alpha R_n,\ \beta P_n)\big)\Big) \tag{7-4}$$

$$MLP = \begin{cases} e^{1-\frac{s}{h}} & if\ h < s \\ e^{1-\frac{h}{s}} & if\ h \geq s \end{cases} \tag{7-5}$$

The EBLEU formula is designed with the factors of modified length penalty (MLP), n-gram precision and recall, the $h$ and $s$ representing the lengths of hypothesis (target) sentence and source sentence respectively.

For the system selection task, we investigated the probability model Naïve Bayes (NB) and support vector machine (SVM) classification algorithms in the QE performances using the features of Length penalty, Precision, Recall and Rank values.

For the word-level error classification task, we investigated a discriminative undirected probabilistic graphical model Conditional random field (CRF), in addition to the NB algorithm. The official results show that the NB algorithm can show overall better performance than the CRF for error classification tasks.

### 7.3.2. QE Model using Universal Phrase Category

We have designed a universal phrase tag-set and utilized it into the MT evaluation without relying on reference translations (Han et al., 2013d). The universal tags we refined are 9 commonly used categories, such as NP, VP, AJP, AVP, PP, S, CONJP, COP, and X. The NP tag covers noun phrase, *Wh-* leading noun phrase, quantifier phrase, prenominal modifiers within an NP, head of the NP, classifier phrase, and localizer phrase, Spanish multiword proper name (MPN). The VP tag covers verbal phrase, coordinated verb compound (VCD), verb-resultative and verb-directional compounds (VRD), verb compounds forming a modifier + head relationship (VSB), verb compounds formed by VV+VC (VCP), verb nucleus (VN.fr), coordinated verb phrase (CVP.de), etc. The AJP tag covers adjective phrase, *Wh*-adjective phrase, determiner phrase (DP.cn), coordinated adjective phrase (CAP.de), multi-token adjective (MTA.de), etc. The AVP tag covers adverb phrase, *Wh*-adverb phrase, English particle (PRT), Chinese "XP+地" phrase, coordinated adverbial phrase, etc. The PP tag covers prepositional phrase, *Wh*-prepositional phrase, German coordinated adposition (CAC), German coordinated adpositional phrase (CPP), German coordinated complementiser (CCP), Spanish multi-token preposition (MTP), etc. The S tag covers sentence, sub-sentence, clause introduced by subordinating conjunction, direct question introduced by *Wh*-word or *Wh*-phrase, fragment, reduced relative clause, parenthetical, incomplete sentences, etc. The CONJP tag covers conjunction phrase, multi-token conjunction, etc. The tag COP covers English unlike coordinated phrase, Chinese un-identical coordination phrase, French coordinated phrase (COORD), German coordination (CO), etc. The X tag covers URL, punctuations, list marker, interjection, Chinese chunks of text that are redundant in a sentence, German Negra idiosyncratic unit (ISU), German Negra quasi-language (QL), etc.

These 9 phrase categories are the most frequently appearing ones in the existing treebanks. We design these 9 phrase categories as the universal phrase tag-set, and conduct the mappings from the tags of existing treebanks to the universal ones. The studied 25 treebanks cover 21 languages, i.e., Arabic, Catalan, Chinese, Danish, English, Estonian, French, German, Hebrew, Hindi, Hungarian, Icelandic, Italian, Japanese, Korean, Portuguese, Spanish, Swedish, Thai, Urdu, and Vietnamese. The mapping results are shown in the tables of Appendix A.

To utilize the designed universal phrase tags into QE research. We firstly, parse the source and target (automatic translated) sentences and extract their phrase sequences. Then, we convert the phrase sequences into universal tags using our designed mapping. Finally, we measure the similarity score on the converted source and target phrase sequences. Let's see an example with French-to-English MT evaluation. Figure 7-1 is the parsing for the source French and target English sentences. Figure 7-2 is the extracted phrase sequences of the two sentences and the conversion into universal phrase tag categories.

Finally, we designed the metric HPPR to measure the similarity of the phrase tag sequences. This metric is the harmonic mean of N1-gram position difference penalty $N_1PsDif$, N2-gram precision $N_2Pre$ and N3-gram recall $N_3Rec$.

$$HPPR = Har(w_{Ps}N_1P\bar{s}Dif, w_{Pr}N_2Pre, w_{Rc}N_3Rec) \quad (7\text{-}6)$$

$$N_1P\bar{s}Dif = \frac{1}{n}\sum N_1PsDif_i \quad (7\text{-}7)$$

$$N_2Pre = exp\left(\sum_{N_2} w_n log P_n\right) \quad (7\text{-}8)$$

$$N_3Rec = exp(\sum_{N_3} w_n log R_n) \quad (7\text{-}9)$$

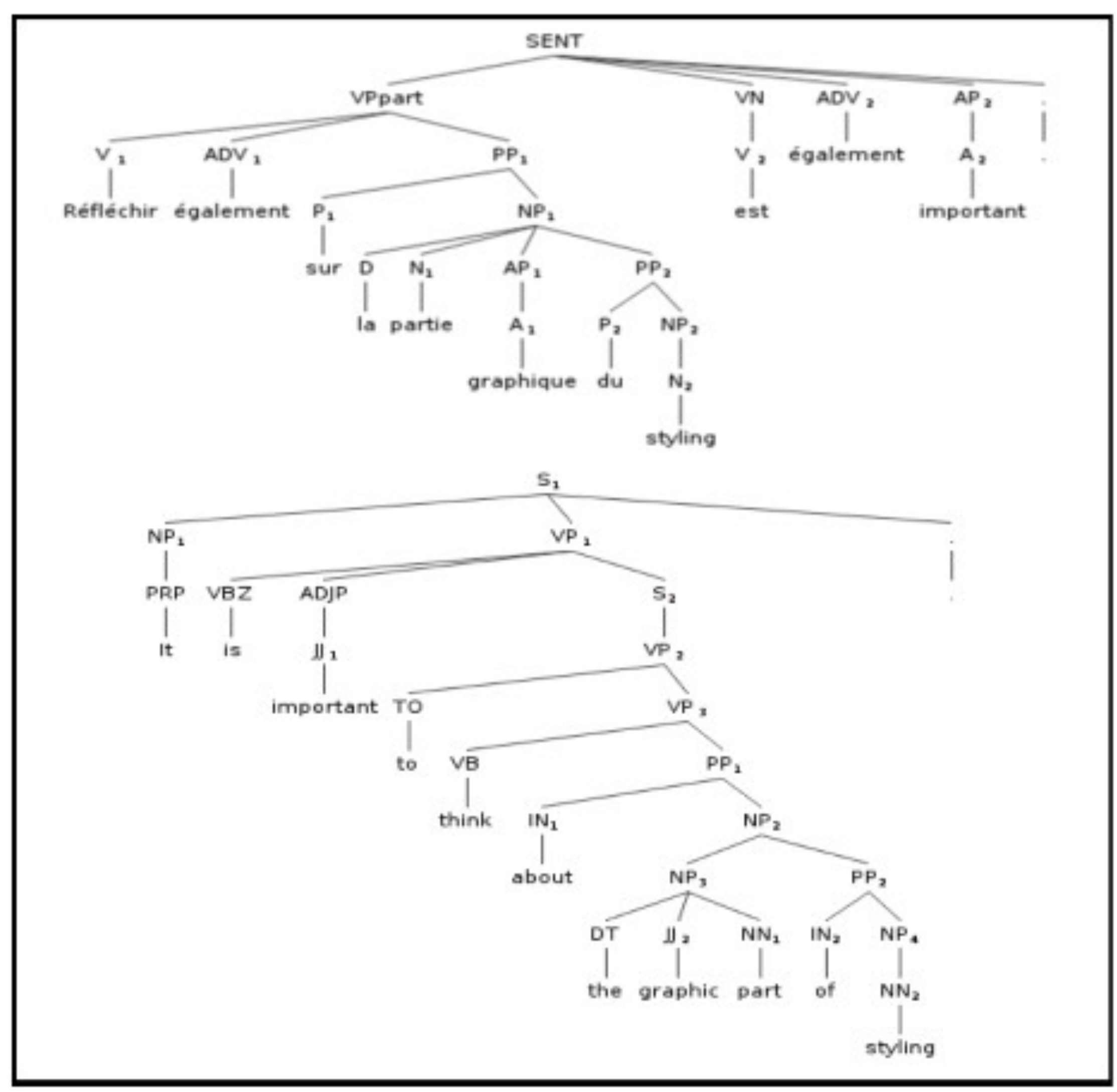

**Figure 7-1: Parsing of the French and English Sentences.**

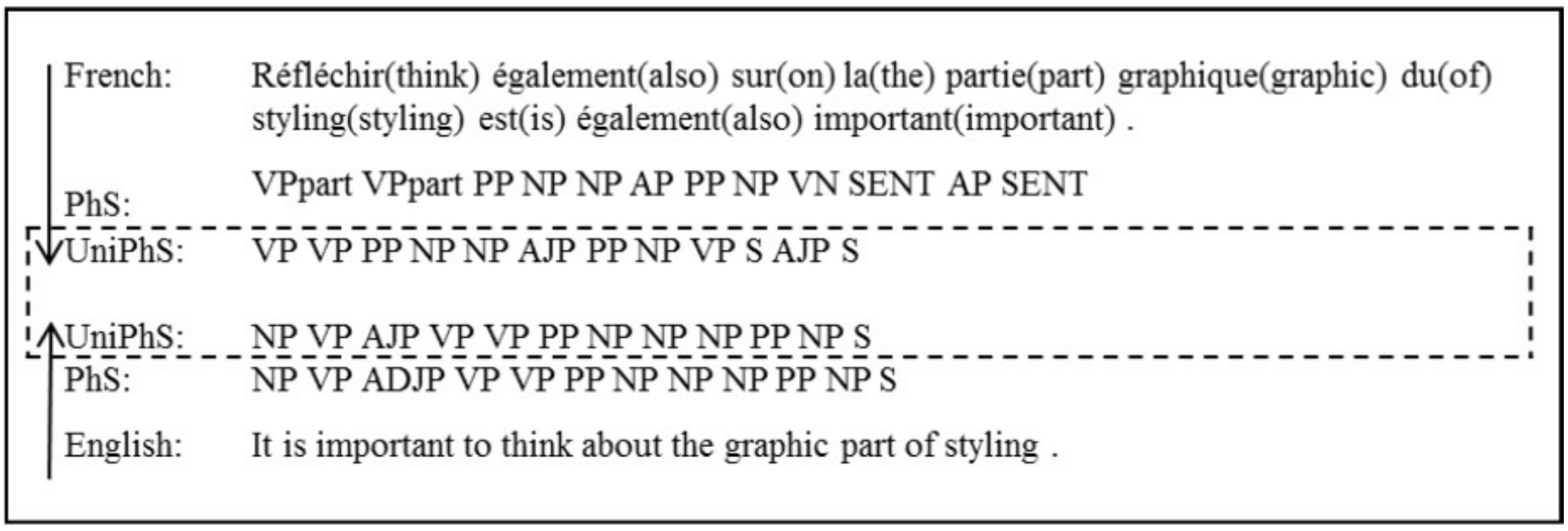

**Figure 7-2: Conversion of the Phrase Tags into Universal Categories.**

We conducted some experiments using our designed HPPR methods. The corpora used in the experiments are from the international workshop of statistical machine translation (WMT). To avoid the overfitting problem, the WMT 2011 corpora are used as the development set to tune the weights of factors in the formula to make the evaluation results close to the human judgments. Then the WMT 2012 corpora are used as the testing set with the formula that has the same parameters tuned in the development stage.

There are 18 and 15 systems respectively in WMT 2011 and WMT 2012 producing the French-to-English translation documents, each document containing 3003 sentences. Each year, there are hundreds of human annotators to evaluate the MT system outputs, and the human judgments task usually costs hundreds of hours of labor. The human judgments are used to validate the automatic metrics. The system-level Spearman correlation coefficient of the different evaluation results will be calculated as compared to the human judgments. The state-of-the-art evaluation metrics BLEU (measuring the closeness between the hypothesis and reference translations) and TER (measuring the editing distance) are selected for the comparison with the designed model HPPR.The values of N2 and N3 are both selected as 3 due to the fact that the4-gram chunk match usually results in 0 score. The tuned factor weights inthe formula are shown in Table 7-2. The experiment results on the testing corpora are shown in Table 7-3, where the phrase "Use Reference?"means whether this metric uses reference translations in the evaluation. The experiment results on the testing corpora show that HPPR without using reference translations has yielded comparable correlation score 0.63 with human judgments even though lower than the reference-aware metrics. This proves to be a promising investigation.

**Table 72: Tuned Parameters of HPPR in the Development Stage.**

| Factors | Parameters | Ratio |
|---|---|---|
| $N_2Pre$ | $w_1 : w_2 : w_3$ | 8:1:1 |
| $N_3Rec$ | $w_1 : w_2 : w_3$ | 1:1:8 |
| $HPPR$ | $w_{Ps} : w_{Pr} : w_{Rc}$ | 1:8:1 |

**Table 73: Evaluation Results of HPPR on WMT 2012 Corpora.**

| Metric | Use reference? | Spearman correlation |
|---|---|---|
| BLEU | Yes | 0.85 |
| TER | Yes | 0.77 |
| HPPR | **No** | 0.63 |

# 8. CONCLUSION AND FUTURE WORK

To facilitate the development of MT itself, it is crucial to tell the MT researchers and developers whether his or her system achieves an improvement after conducting some revisions, such as new algorithms or features. This work introduces our proposed methods for the automatic MT evaluation. To address some of the weaknesses in the existing MT evaluation methods, we designed augmented factors, tunable parameters and concise linguistic features to yield reliable evaluation. Furthermore, our methods can be easily employed to different language pairs, or new language pairs due to the concise external resources utilized. For the existing weakness of MT evaluation, such as the evaluation with non-English in target language and the low resource language pairs, our proposed methods have shown some improvements as compared with the state-of-the-art metrics. We also introduced our designed model for the quality estimation of MT and the experiments show that our proposed methods without using reference translations yielded promising results as compared with the reference-aware metrics. In spite of the many efforts from the MT evaluation researchers, there remain some issues for the future research as bellow.

## 8.1. Issues in Manual Judgments

So far, the human judgment scores of MT results are usually considered as the golden standard that the automatic evaluation metrics should try to approach. However, some improper handlings in the process also yield problems. For instance, in the ACL WMT 2011 English-Czech task, the multi-annotator agreement kappa value $k$ is very low and even the exact same string produced by two systems is ranked differently each time by the same annotator (Bojar et al., 2011). Secondly, the evaluation results are highly affected by the manual reference translations. How to ensure the quality of reference translations and the agreement level of human judgments are two important problems.

## 8.2. Issues in Automatic Evaluation

First, automatic evaluation metrics are indirect measures (Moran and Lewis, 2011)of translation quality, because that they are usually using the various string distance algorithms to measure the closeness between the machine translation system outputs and the manually offered reference translations and they are based on the calculating of correlation score with manual MT evaluation.

Furthermore, the existing automatic evaluation metrics tend to ignore the relevance of words (Koehn, 2010). For instance, the name entities and core concepts are more important than punctuations and determiners but most automatic evaluation metrics put the same weight on each word of the sentences.

Third, existing automatic evaluation metrics usually yield meaningless score, which is very test set specific and the absolute value is not informative. For instance, what is the meaning of -16094 score by the MTeRater metric (Parton et al., 2011) or 1.98 score by ROSE (Song and Cohn, 2011)?

Fourth, some of the existing automatic metrics only use the surface words information without any linguistic features, which makes them result in low correlation with human judgment and receives much criticism from the linguists; on the other hand, some metrics utilize too many language specific linguistic features, which make it difficult to promote them on other language pairs. How to handle the balance between the two aspects is a challenge before researchers.

The automatic evaluation metrics should try to achieve the goals of low cost, tunable, consistent, meaningful, and correct, of which the first three aspects are easily achieved but the rest two goals, i.e. meaningful and correct, and the robustness in different language pairs are usually the challenges in front of us.

## 8.3. Future Work

This work tried to advance the MT evaluation by using augmented factors and concise linguistic features. In our future work, we plan to investigate the MT evaluation performance from some different aspects.

Firstly, we want to utilize our designed universal phrase tagset into the MT evaluation on more language pairs. In this work, we only employ the universal tagset into French-English MT evaluation, and it has shown some promising results.

Secondly, we plan to enhance the performance of the designed LEPOR MT evaluation models with extended linguistic features, especially semantic features, such as synonyms, paraphrasing and text entailments.

Thirdly, we want to investigate some machine learning technologies into MT evaluation. For instance, we plan to utilize the deep learning method to convert the surface words into the vector form. In this way, we can measure the similarity of source and target languages on the vector level instead of word or sentence level. The reference translation can be a waiver in this framework.

# BIBLIOGRAPHY

Agarwal, A. and A. Lavie. 2008. METEOR, M-BLEU and M-TER: Evaluation Metrics for High-Correlation with Human Rankings of Machine Translation Output. In Proceedings of the Third Workshop on Statistical Machine Translation at the 46th Meeting of the Association for Computational Linguistics (ACL-2008), Columbus, OH, June 2008. Pages 115-118.

Aikawa, Takako and Spencer Rarrick. 2011. Are numbers good enough for you? - A linguistically meaningful MT evaluation method,MT Summit XIII, 332-337.

Akiba, Y., K. Imamura, and E. Sumita. 2001. Using Multiple Edit Distances to Automatically Rank Machine Translation Output. In Proceedings of the MT Summit VIII, Santiago de Compostela, Spain.

Akiba, Yasuhiro, Kenji Imamura, Eiichiro Sumita, Hiromi Nakaiwa, Seiichi Yamamoto, and Hiroshi G. Okuno. 2006. Using Multiple Edit Distances to Automatically Grade Outputs From Machine Translation Systems, IEEE TRANSACTIONS ON AUDIO, SPEECH, AND LANGUAGE PROCESSING, VOL. 14, NO. 2, MARCH 2006 393-402.

Albrecht, Joshua, and Rebecca Hwa.2007.Regression for sentence-level MT evaluationwith pseudo references.ACL. Vol. 45. No. 1.

Alqudsi, Arwa, Nazlia Omar and Khalid Shaker. 2012. Arabic machine translation: a survey, July,2012, J. Artificial Intelligence Review, Springer.

Androutsopoulos, Ion and Prodromos Malakasiotis. 2010. A Survey of Paraphrasing and Textual Entailment Methods. Journal of Artificial Intelligence Research 38(2010), pp. 135-187.

Arnold, D. 2003. Why translation is difficultfor computers. In Computers and Translation:A translator's guide. Benjamins Translation Library.

Avramidis, E., Popovic, M., Vilar, D., Burchardt, A. 2011. Evaluate with Confidence Estimation: Machine ranking of translation outputs using grammatical features. In Proceedings of the Sixth Workshop on Statistical Machine Translation, Association for Computational Linguistics (ACL-WMT), pages 65-70, Edinburgh, Scotland, UK.

Avramidis, Eleftherios.2012. Comparative Quality Estimation: Automatic Sentence-Level Ranking of Multiple Machine Translation Outputs, Proceedings of 24th International Conference on Computational Linguistics. International Conference on Computational Linguistics (COLING-12), December 8-15, Mumbai, India, pages 115-132, publisher The COLING 2012 Organizing Committee

Avramidis, Eleftherios. 2012. Quality estimation for Machine Translation output using linguistic analysis and decoding features, Proceedings of the 7th Workshop on Statistical Machine Translation, pages 84–90.

Aziz, Wilker, Marc Dymetmany, Shachar Mirkinx, Lucia Speciaz, Nicola Cancedday, Ido Dagan. 2010. Learning an Expert from Human Annotations in Statistical Machine Translation: the Case of Out-of-VocabularyWords, EAMT2010.

Babych, B, Hartley A, Atwell E. 2003. Statistical Modelling of MT output corpora for Information Extraction. In: Proceedings of the Corpus Linguistics 2003 conference, edited by Dawn Archer, Paul Rayson, Andrew Wilson and Tony McEnery. Lancaster University (UK), 28 - 31 March 2003. pp. 62-70.

Babych, B, Hartley A. 2004a. Modelling legitimate translation variation for automatic evaluation of MT quality, LREC 2004.

Babych, B. 2004. Weighted N-gram model for evaluating Machine Translation output. CLUK `04. Proceedings of the 7th Annual Colloquium for the UK Special Interest Group for Computational Linguistics. Unviersity of Birmingham 6-7 January, 2004. pp. 15-22.

Babych, Bogdan and Anthony Hartley. 2004b. Extending the BLEU MT evaluation method with frequency weightings. In Proceedings of the 42nd Annual Meeting on Association for Computational Linguistics (ACL '04). Association for Computational Linguistics, Stroudsburg, PA, USA, Article 621.

Banerjee, Satanjeev and Alon Lavie. 2005. METEOR: An Automatic Metric for MT Evaluation with Improved Correlation with Human Judgments. In Proceedings of the 43th Annual Meeting of the Association of Computational Linguistics (ACL- 05), pages 65–72, 2005.

Bangalore, Srinivas, Owen Rambow, and Steven Whittaker.2000. Evaluation metrics for generation.In In Proceedings of the First International Natural Language Generation Conference (INLG2000), pages 1–8, Mitzpe Ramon, Israel.

Barrón-Cedeño, Alberto, Lluís Màrquez, Maria Fuentes, Horacio Rodriguez, Jordi Turmo. 2013. UPC-CORE: What Can Machine Translation Evaluation Metrics and Wikipedia Do for Estimating Semantic Textual Similarity?Second Joint Conference on Lexical and Computational Semantics (SEM), Volume 1: Proceedings of the Main Conferenceand the Shared Task, pages 143-147, Atlanta, Georgia, June 13-14, 2013.2013 Association for Computational Linguistics.

Barzilay, Regina and Lillian Lee. 2003. Learning to paraphrase: an unsupervised approach using multiple-sequence alignment. In Proceedings of the 2003 Conference of the North American Chapter of the Association for Computational Linguistics on Human Language Technology - Volume 1 (NAACL '03), Vol. 1. Association for Computational Linguistics, Stroudsburg, PA, USA, 16-23. DOI=10.3115/1073445.1073448 http://dx.doi.org/10.3115/1073445.1073448

Bech, A. 1997. MT from an Everyday User's Point ofView. In Proceedings of MT Summit VI. pp. 98-105. San Diego.

Bentivogli, Luisa, Marcello Federico, Giovanni Moretti and Michael Paul. 2011. Getting Expert Quality from the Crowd for Machine Translation Evaluation, in MT Summit 2011.

Blatz, J., E. Fitzgerald, G. Foster, S. Gandrabur, C. Goutte,A. Kulesza, A. Sanchis, and N. Ueffing. 2004. ConfidenceEstimation for Machine Translation. In Proceedingsof the 20th international conference on ComputationalLinguistics (COLING '04). Association forComputational Linguistics.

Bosma, W., P. Vossen, A. Soroa, G. Rigau, M.Tesconi, A. Marchetti, M. Monachini, and C.Aliprandi. 2009. KAF: a generic semantic annotationformat. Proceedings of the 5th InternationalConference on Generative Approaches to theLexicon (GL2009), Workshop on Semantic Annotation.Pisa, Italy, pp. 145-152.

Brown, P. F., Pietra, S. A. D., Pietra, V. J. D., & Mercer, R. L. 1993. The mathematicsof statistical machine translation: Parameter estimation. Computational Linguistics,19 (2), 263-311.

Buck, Christian. 2012. Black Box Features for the WMT 2012 Quality Estimation Shared Task, Proceedings of the 7th Workshop on Statistical Machine Translation, pages 91–95.

Callison-Burch, C., Koehn, P., Monz, C. and Zaidan, O. F. 2011. Findings of the 2011 Workshop on Statistical Machine Translation. In Proceedings of the Sixth Workshop on Statistical Machine translation of the Association for Computational Linguistics(ACL-WMT), pages 22-64, Edinburgh, Scotland, UK.

Callison-Burch, C., Koehn, P., Monz, C., Peterson, K., Przybocki, M. and Zaidan, O. F. 2010. Findings of the 2010 Joint Workshop on Statistical Machine Translation and Metrics for Machine Translation. In Proceedings of the 5th Workshop on Statistical Machine Translation, Stroudsburg, Association for Computational Linguistics(ACL-WMT), pages 17-53, PA, USA.

Callison-Burch, C., Koehn, P., Monz,C. and Schroeder, J. 2008. Further meta-evaluation of machine translation. In Proceedings of the Third Workshop on Statistical Machine Translation, Association for Computational Linguistics (ACL-WMT), pages 70-106, Columbus, Ohio, USA.

Callison-Burch, C., Koehn, P., Monz,C. and Schroeder, J. 2009. Findings of the 2009 Workshop on Statistical Machine Translation. In Proceedings of the 4th Workshop on Statistical Machine Translation, the European Chapter of Association for Computational Linguistics (EACL-WMT), pages 1-28, Athens, Greece.

Callison-Burch, Chris, Cameron Fordyce, Philipp Koehn, Christof Monz and Josh Schroeder. 2007. (Meta-) Evaluation of Machine Translation, Proceedings of the Second Workshop on Statistical Machine Translation, pages 136-158, Prague, June 2007.

Callison-Burch, Chris, Philipp Koehn, Christof Monz, Matt Post, Radu Soricut, and Lucia Specia. 2012. Findings of the 2012Workshop on Statistical Machine Translation. In Proceedings of the Seventh Workshop on Statistical Machine Translation, Montreal, Canada, June. Association for Computational Linguistics. pages 10–51.

Carl, M. and A. Way. 2003. Recent Advances in Example-Based Machine Translation. Kluwer Academic Publishers, Dordrecht, The Netherlands.

Carroll, J. B. 1966b. Anexperiment in evaluating the quality of translation. Pierce, J. (Chair), Languages and machines: computers in translation and linguistics. A report by the Automatic Language Processing Advisory Committee (ALPAC), Publication 1416, Division of Behavioral Sciences, National Academy of Sciences, National Research Council, page 67-75, http://www.nap.edu/catalog.php?record_id=9547

Carroll, John B. 1966a. An Experiment in Evaluating the Quality of Translations, Mechanical Translation and Computational Linguistics, vol.9, nos.3 and 4, September and December 1966, page 55-66, Graduate School of Education, Harvard University, file:///D:/down_papar_NLP/An%20experiment%20in%20evaluating%20the%20quality%20of%20translations.htm

Castillo, Julio and Paula Estrella. 2012. Semantic Textual Similarity for MT evaluation, Proceedings of the 7th Workshop on Statistical Machine Translation (WMT2012), pages 52–58, Montre´al, Canada, June 7-8, 2012. Association for Computational Linguistics.

Chan, Y. S. and Ng, H. T. 2008. MAXSIM: A maximum similarity metric for machine translation evaluation. In Proceedings of ACL 2008: HLT, pages 55–62. Association for Computational Linguistics.

Chen, B. and Kuhn, R. 2011. Amber: A modified bleu, enhanced ranking metric. In Proceedings of the Sixth Workshop on Statistical Machine translation of the Association for Computational Linguistics(ACL-WMT), pages 71-77, Edinburgh, Scotland, UK.

Chen, Boxing, Roland Kuhn and George Foster. 2012a. Improving AMBER, an MT Evaluation Metric. Proceedings of the 7th Workshop on Statistical Machine Translation, pages 59–63,Montre´al, Canada, June 7-8, 2012. 2012 Association for Computational Linguistics

Chen, Boxing, Roland Kuhn and Samuel Larkin. 2012b. PORT: a Precision-Order-Recall MT Evaluation Metric for Tuning, Proceedings of the 50th Annual Meeting of the Association for Computational Linguistics, pages 930–939, Jeju, Republic of Korea, 8-14 July 2012.

Chinchor, Nancy. 1992. MUC-4 Evaluation Metrics, in Proc. of the Fourth Message Understanding Conference, pp. 22–29, 1992.

Church, Kenneth, and Eduard Hovy. 1991. Good Applications for Crummy Machine Translation. In Jeannette G. Neal and Sharon M. Walter (eds.) Proceedings of the 1991 Natural Language Processing Systems Evaluation Workshop. Rome Laboratory Final Technical Report RLTR- 91-362.

Cohen, Jacob. 1960. A coefficient of agreement for nominal scales. Educational and Psychological Measurement, 20(1):37–46.

Connell, J. & Shaffer R. 1995. Object-Oriented Rapid Prototyping, Prentice-Hall, Englewood Cliffs, NJ.

Costa-jussà, M. R., C. A. Henríquez and R. E. Banchs. 2012. Evaluating Indirect Strategies for Chinese-Spanish Statistical Machine Translation, Journal of artificial intelligence research, Volume 45, pages 761-780.

Dagan, I., Glickman, O., & Magnini, B. 2006. The PASCAL recognising textual entailment challenge.In Quinonero-Candela, J., Dagan, I., Magnini, B., & d'Alche' Buc, F. (Eds.), MachineLearning Challenges. Lecture Notes in Computer Science, Vol. 3944, pp. 177–190. SpringerVerlag.

Dagan, Ido, and Oren Glickman. 2004. Probabilistic Textual Entailment: Generic Applied Modeling of Language Variability, In: Learning Methods for Text Understanding and Mining workshop, 26-29 January 2004, Grenoble, France.

Dagan, Ido, Oren Glickman, and Bernardo Magnini. 2005. The PASCAL recognising textual entailment challenge. In Proceedings of the First international conference on Machine Learning Challenges: evaluating Predictive Uncertainty Visual Object Classification, and Recognizing Textual Entailment(MLCW'05), Joaquin Quiñonero-Candela, Ido Dagan, Bernardo Magnini, and Florence d'Alché-Buc (Eds.). Springer-Verlag, Berlin, Heidelberg, 177-190. DOI=10.1007/11736790_9 http://dx.doi.org/10.1007/11736790_9

Dahlmeier, D., Liu, C. and Ng, H. T. 2011. TESLA at WMT2011: Translation evaluation and tunable metric. In Proceedings of the Sixth Workshop on Statistical Machine Translation, Association for Computational Linguistics (ACL-WMT), pages 78-84, Edinburgh, Scotland, UK.

de Souza, Jose Guilherme Camargo, Matteo Negri and Yashar Mehdad. 2012. FBK: Machine Translation Evaluation and Word Similarity metrics for Semantic Textual Similarity, First Joint Conference on Lexical and Computational Semantics (SEM), pp 624-630, Montreal, Canada, June 7-8, 2012. Association for Computational Linguistics.

Denkowski, M. and Lavie, A. 2011. Meteor 1.3: Automatic metric for reliable optimization and evaluation of machine translation systems. In Proceedings of the Sixth Workshop on Statistical Machine translation of the Association for Computational Linguistics(ACL-WMT), pages 85-91, Edinburgh, Scotland, UK.

Doddington, G. 2002. Automatic Evaluation of Machine Translation Quality using N-gram Co-occurrence Statistics. In Proceedings of the 2nd International Conference on Human Language Technology Research (HLT-02), pages 138–145, San Francisco, CA, USA, 2002. Morgan Kaufmann Publishers Inc.

Dras, Mark. 1999. Tree Adjoining Grammar and the ReluctantParaphrasing of Text. Ph.D. thesis, Macquarie University.

Dreyer, Markus and Daniel Marcu. 2012. HyTER: Meaning-Equivalent Semantics for Translation Evaluation, Conference of the North American Chapter of the Association for Computational Linguistics: Human Language Technologies, pages 162–171, Montre´al, Canada, June 3-8, 2012. Association for Computational Linguistics.

EAGLES. 1996. EAGLES Evaluation of Natural Language Processing Systems. Final Report. EAGLES Document EAG-EWG-PR.2, Center for Sprogteknologi, Copenhagen.

Echizen-ya, H. and Araki, K. 2010. Automatic evaluation method for machine translation using noun-phrase chunking. In Proceedings of ACL 2010, pages 108–117. Association for Computational Linguistics.

Eck, Matthias and Chiori Hori. 2005. Overview of the IWSLT 2005 Evaluation Campaign, IWSLT 2005.

Federico, Marcello, Luisa Bentivogli, Michael Paul, and Sebastian Stiiker. 2011. Overview of the IWSLT 2011 Evaluation Campaign, 11-27.

Federmann, Christian. 2011. How can we measure machine translation quality?, Tralogy [En ligne], Session 5 - Quality in Translation / La qualité en traduction, mis à jour le : 18/11/2011,URL : http://lodel.irevues.inist.fr/tralogy/index.php?id=76.

Federmann, Christian. 2012. Appraise: an Open-Source Toolkit for Manual Evaluation of MT Output. The Prague Bulletin of Mathematical Linguistics No. 98, 2012, pp. 25–35. doi: 10.2478/v10108-012-0006-9.

Felice, Mariano and Lucia Specia. 2012. Linguistic Features for Quality Estimation. Proceedings of the 7th Workshop on Statistical Machine Translation, pages 96–103.

Fellbaum, C. 1998. WordNet: An Electronic Lexical Database. MIT Press. http://www.cogsci.princeton.edu/˜wn [2000, September 7].

Finkel, Jenny R. and Christopher D. Manning. 2010. Hierarchicaljoint learning: improving joint parsing andnamed entity recognition with non-jointly labeled data.In 48th

Annual Meeting of the Association for ComputationalLinguistics, pages 720–728, Uppsala, Sweden.

Finkel, Jenny Rose, Trond Grenager, and Christopher Manning. 2005. Incorporating non-local information into information extraction systems by gibbs sampling. In Proceedings of the 43rd Annual Meeting on Association for Computational Linguistics, ACL 2005, pages 363–370, Stroudsburg, PA, USA.

Fishel, Mark, Rico Sennrich, Maja Popovic and Ondrej Bojar. 2012. TerrorCat: a Translation Error Categorization-based MT Quality Metric. Proceedings of the 7th Workshop on Statistical Machine Translation, pages 64–70, Montreal, Canada, June 7-8, 2012. Association for Computational Linguistics.

Fraser, Alexander, Daniel Marcu. 2007. Measuring Word Alignment Quality for Statistical Machine Translation, J. Computational Linguistics, September 2007, Vol. 33, No. 3, pages 293-303.

Gamon, Michael, Anthony Aue, Martine Smets. 2005. Sentence-level MT evaluation without reference translations Beyond language modelling, Proceedings of EAMT.2005.pp 103-112.

Gimenez, Jesus and Enrique Amigo. 2008. IQMT: A Framework for Automatic Machine Translation Evaluation, LREC 2006.

Giménez, Jesús and Lluís Márquez. 2007. Linguistic Features for Automatic Evaluation of Heterogenous MT Systems. In Proceedings of the Second Workshop on Statistical Machine Translation, pages 256–264, Prague, June 2007. Association for Computational Linguistics.

Gimenez, Jesus and Lluıs Marquez. 2008. A Smorgasbord of Features for Automatic MT Evaluation. In Proceedings of the 3rd Workshop on Statistical Machine Translation, pages 195–198, Columbus, OH, June 2008. Association for Computational Linguistics.

Gonzalez, Meritxell, Jesus Gimenez and Lluıs Marquez. 2012. A Graphical Interface for MT Evaluation and Error Analysis, Proceedings of the 50th Annual Meeting of the Association for Computational Linguistics, Demo paper, pages 139–144, Jeju, Republic of Korea, 8-14 July 2012. Association for Computational Linguistics.

Gow, Francie.2003. Metrics for Evaluating Translation Memory Software, School of Translation and Interpretation, University of Ottawa. thesis 2003.

Guo, Jiafeng, Gu Xu, Xueqi Cheng, and Hang Li. 2009. Named entity recognition in query. InProceedings of the 32nd international ACM SIGIR conference on Research and development in information retrieval(SIGIR '09). ACM, New York, NY, USA, 267-274. DOI=10.1145/1571941.1571989 http://doi.acm.org/10.1145/1571941.1571989

Hald, Anders. 1998. A History of Mathematical Statistics. New York: Wiley. ISBN 0-471-17912-4.

Hamon, Olivier and Khalid Choukri, Evaluation Methodology and Results for English-to-Arabic MT, MT Summit 2011, XiaMen, China.

Han, Aaron Li-Feng, Derek F. Wong, Lidia S. Chao. 2012. LEPOR: A Robust Evaluation Metric for Machine Translation with Augmented Factors. Proceedings of the 24th International Conference on Computational Linguistics (COLING): Posters, pages 441–450, Mumbai, December 2012. Association for Computational Linguistics. http://aclweb.org/anthology//C/C12/C12-2044.pdf

Han, Aaron Li-Feng, Derek F. Wong, Lidia S. Chao, Liangye He, Yi Lu, Junwen Xing and Xiaodong Zeng. 2013a. Language-independent Model for Machine Translation Evaluation with Reinforced Factors. Proceedings of the 14th International Conference of Machine Translation Summit (MT Summit), pp. 215-222. Nice, France. 2 - 6 September 2013. International Association for Machine Translation. http://www.mt-archive.info/10/MTS-2013-Han.pdf

Han, Aaron Li-Feng, Derek Wong, Lidia S. Chao, Yi Lu, Liangye He, Yiming Wang, Jiaji Zhou. 2013b. A Description of Tunable Machine Translation Evaluation Systems in WMT13 Metrics Task. Proceedings of the ACL 2013 EIGHTH WORKSHOP ON STATISTICAL MACHINE TRANSLATION (ACL-WMT), pp. 414-421, 8-9 August 2013. Sofia, Bulgaria. Association for Computational Linguistics. http://www.aclweb.org/anthology/W13-2253

Han, Aaron Li-Feng, Yi Lu, Derek F. Wong, Lidia S. Chao, Liangye He, Junwen Xing. 2013c. Quality Estimation for Machine Translation Using the Joint Method of Evaluation Criteria and Statistical Modeling. Proceedings of the ACL 2013 EIGHTH WORKSHOP ON STATISTICAL MACHINE TRANSLATION (ACL-WMT), pp. 365-372. 8-9 August 2013. Sofia, Bulgaria. Association for Computational Linguistics. http://www.aclweb.org/anthology/W13-2245

Han, Aaron Li-Feng, Derek F. Wong, Lidia S. Chao, Liangye He, Shuo Li and Ling Zhu. 2013d. Phrase Tagset Mapping for French and English Treebanks and Its Application in Machine Translation Evaluation. Language Processing and Knowledge in the Web. Lecture Notes in Computer Science Volume 8105, 2013, pp 119-131. Volume Editors: Iryna Gurevych, Chris Biemann and Torsten Zesch. Springer-Verlag Berlin Heidelberg. http://dx.doi.org/10.1007/978-3-642-40722-2_13

Han, Aaron Li-Feng, Derek F. Wong, Lidia S. Chao, Liangye He, Ling Zhu and Shuo Li. 2013e. A Study of Chinese Word Segmentation Based on the Characteristics of Chinese. Language Processing and Knowledge in the Web. Lecture Notes in Computer Science Volume 8105, 2013, pp 111-118. Volume Editors: Iryna Gurevych, Chris Biemann and Torsten Zesch. Springer-Verlag Berlin Heidelberg. http://dx.doi.org/10.1007/978-3-642-40722-2_12

Han, Aaron Li-Feng, Derek F. Wong, Lidia S. Chao, Liangye He. 2013f. Automatic Machine Translation Evaluation with Part-of-Speech Information. Text, Speech, and Dialogue. Lecture Notes in Computer Science Volume 8082, 2013, pp 121-128. Volume Editors: I. Habernal and V. Matousek. Springer-Verlag Berlin Heidelberg. http://dx.doi.org/10.1007/978-3-642-40585-3_16

Han, Aaron Li-Feng, Derek Fai Wong and Lidia Sam Chao. 2013g. Chinese Named Entity Recognition with Conditional Random Fields in the Light of Chinese Characteristics. Language Processing and Intelligent Information Systems. Lecture Notes in Computer Science Volume 7912, 2013, pp 57-68. M.A. Klopotek et al. (Eds.): IIS 2013. Springer-Verlag Berlin Heidelberg. http://dx.doi.org/10.1007/978-3-642-38634-3_8

Han, Aaron Li-Feng, Derek F. Wong, Lidia S. Chao, Liangye He and Yi Lu. 2014. Unsupervised Quality Estimation Model for English to German Translation and Its Application in Extensive Supervised Evaluation. The Scientific World Journal, Issue:

Recent Advances in Information Technology. Page 1-12, April 2014. Hindawi Publishing Corporation. ISSN:1537-744X. http://www.hindawi.com/journals/tswj/aip/760301/

Imamura, Kenji, Eiichiro Sumita, and Yuji Matsumoto. 2003. Feedback cleaning of machine translation rules using automatic evaluation. In Proceedings of the 41st Annual Meeting on Association for Computational Linguistics - Volume 1 (ACL '03), Vol. 1. Association for Computational Linguistics, Stroudsburg, PA, USA, 447-454. DOI=10.3115/1075096.1075153 http://dx.doi.org/10.3115/1075096.1075153.

Isozaki, H., T. Hirao, K. Duh, K. Sudoh, H. Tsukada. 2010. Automatic Evaluation of Translation Quality for Distant Language Pairs. In Proceedings of EMNLP.

Jagarlamudi, Jagadeesh and Hal Daume III.2012. Regularized Interlingual Projections: Evaluation on Multilingual Transliteration, Proceedings of the 2012 Joint Conference on Empirical Methods in Natural Language Processing and Computational Natural Language Learning, pages 12–23, Jeju Island, Korea, 12–14 July 2012. Association for Computational Linguistics.

Joachinms, T. 2006. Training linear SVMs in linear time. In Proceedings of the ACM Conference on Knowledge Discovery and Data Mining. page 217-226, Philadelphia, PA, USA.

Kauchak, D. and R. Barzilay. 2006. Paraphrasing for Automatic Evaluation. Proceedings of the Human Language Technology Conference of the North American Chapter of the ACL, pages 455–462.

Kauchak, David, and Regina Barzilay. 2006. Paraphrasing for automatic evaluation. InProceedings of the main conference on Human Language Technology Conference of the North American Chapter of the Association of Computational Linguistics (HLT-NAACL '06). Association for Computational Linguistics, Stroudsburg, PA, USA, 455-462.

Kendall, Maurice G.1938.A new measure of rank correlation. Biometrika, 30:81–93.

Kendall, Maurice G.and Jean Dickinson Gibbons.1990. Rank Correlation Methods.Oxford University Press, New York.

Kerridge, D. 1975. The interpretation of rank correlations. Applied Statistics, 24(2): 257–258.

King, M. 1990. A Workshop on Evaluation: Background paper. Proceedings of the Third International Conference on Theoretical and Methodological Issues in Machine Translation of Natural Languages, Linguistic Research Center, University of Texas, Austin, TX. 255-259

King, Margaret, Andrei Popescu-Belis and Eduard Hovy. 2003. FEMTI: Creating and Using a Framework for MT Evaluation. In Proceedings of MT Summit IX, New Orleans, LA. Sept. 2003. pp. 224-231.

Koehn, P. 2004. Statistical significance tests for machine translation evaluation. In Lin, D. and Wu, D., editors, Proceedings of EMNLP 2004, pages 388–395, Barcelona, Spain. Association for Computational Linguistics.

Koehn, P. and Monz, C. 2005. Shared task: Statistical machine translation between European languages. In Proceedings of the ACL Workshop on Building and Using Parallel Texts, pages 119–124, Ann Arbor, Michigan. Association for Computational Linguistics.

Koehn, P., Hoang, H., Birch, A., Callison-Burch, C.,Federico, M., Bertordi, N., Cowan, B., Shen, W.,Moran, C., Zens, R., Dyer, C., Bojar, O., Constantin,A. and Herbst, E. 2007.

Moses: Open Source Toolkitfor Statistical Machine Translation. In Proceedings ofthe ACL 2007, pp. 177-180.

Koehn, Philipp and Christof Monz. 2006. Manual and Automatic Evaluation of Machine Translation between European Languages, Proceedings of the ACLWorkshop on Statistical Machine Translation, pages 102–121,New York City, June 2006.

Koehn, Philipp. 2010. Statistical Machine Translation, (University of Edinburgh), Cambridge University Press.

Koehn, Philipp. 2011.What is a better translation-reflections on six years of running evaluation campaigns. Tralogy, 2011.

Kos, K. and O. Bojar. 2009. Evaluation of Machine Translation Metrics for Czech as the Target Language, 2009, The Prague Bulletin of Mathematical Linguistics, Number 92, 135-147.

Kuhn, H. W. 1955. The Hungarian method for the assignment problem. Naval research logistics quarterly, 2:83–97.

Landis, J. Richard. and Gary G. Koch. 1977. The measurement of observer agreement for categorical data. Biometrics, 33:159–174.

Lapata, Mirella. 2003. Probabilistic text structuring: Experiments with sentence ordering.In Proceedings of the 41st Annual Meeting of the Association for Computational Linguistics, pages 545–552, Sapporo, Japan.

Lapata, Mirella. 2006. Automatic Evaluation of Information Ordering: Kendall's Tau, J. Computational Linguistics, December 2006, Vol. 32, No. 4, Pages 471-484.

Lavie, Alon, and Abhaya Agarwal. 2007. METEOR: An Automatic Metric for MT Evaluation with High Levels of Correlation with Human Judgments, Proceedings of the ACL Second Workshop on Statistical Machine Translation, pages 228-231, Prague, June 2007.

Lavie, Alon, Kenji Sagae, Shyamsundar Jayaraman. 2004. The Significance of Recall in Automatic Metrics for MT Evaluation, pp 134-143. The 6th conference of the association for machine translation in the Americas, AMTA2004, Washington, DC, USA.

LDC. 2005. Linguistic data annotation specification:Assessment of fluency and adequacy in translations.Revision 1.5.

Lebanon, Guy and John Lafferty. 2002. Combining rankings using conditional probability models on permutations.In Proceedings of the 19th International Conference on Machine Learning.San Francisco, CA: Morgan Kaufmann Publishers, pages 363–370.

Lembersky, Gennadi, Noam Ordan, Shuly Wintner. 2012. Language Models for Machine Translation- Original vs. Translated Texts.2012, J. Computational Linguistics, Volume 38, Number 4.

Leusch, Gregor and Nicola Ueffing and David Vilar and Hermann Ney. 2005. Preprocessing and Normalization for Automatic Evaluation of Machine Translation. Proceedings of the ACL Workshop on Intrinsic and Extrinsic Evaluation Measures for Machine Translation and/or Summarization, pages 17–24, Ann Arbor, June 2005.

Leusch, Gregor, Nicola Ueffing, and Hermann Ney. 2006. CDer: Efficient MT Evaluation Using Block Movements. In Proceedings of the 13th Conference of the European Chapter of the Association for Computational Linguistics (EACL-06), 241—248, 2006.

Li, A. 2005. Results of the 2005 NIST machine translation evaluation. In Machine Translation Workshop.

LI, LiangYou, GONG ZhengXian and ZHOU GuoDong. 2012. Phrase-Based Evaluation for Machine Translation, Proceedings of COLING 2012: Posters, pages 663–672,COLING 2012, Mumbai, December 2012.

Li, Xianhua, Yajuan Lü, YaoMeng, Qun Liu, Hao Yu. 2011. Feedback Selecting of Manually Acquired Rules Using Automatic Evaluation, Proceedings of the 4th Workshop on Patent Translation, pages 52-59,MT Summit XIII, Xiamen, China, September 2011.

Lin, Chin-Yew and E.H. Hovy, 2003. Automatic Evaluation of Summaries Using N-gram Co-occurrence Statistics. In Proceedings of 2003 Language Technology Conference (HLT-NAACL 2003), Edmonton, Canada, May 27 - June 1, 2003.

Lin, Chin-Yew and Franz Josef Och. 2004a. Automatic Evaluation of Machine Translation Quality Using Longest Common Subsequence and Skip-Bigram Statistics. In Proceedings of the 42nd Annual Meeting of the Association for Computational Linguistics (ACL 2004), Barcelona, Spain, July 21 - 26, 2004.

Lin, Chin-Yew and Franz Josef Och. 2004b. ORANGE: a Method for Evaluating Automatic Evaluation Metrics for Machine Translation. In Proceedings of the 20th International Conference on Computational Linguistics (COLING 2004), Geneva, Switzerland, August 23 - August 27, 2004.

Lin, Chin-Yew. 2004a. ROUGE: a Package for Automatic Evaluation of Summaries. In Proceedings of the Workshop on Text Summarization Branches Out (WAS 2004), Barcelona, Spain, July 25 - 26, 2004.

Lin, Chin-Yew. 2004b. Looking for a Few Good Metrics: Automatic Summarization Evaluation - How Many Samples Are Enough?. In Proceedings of the NTCIR Workshop 4, Tokyo, Japan, June 2 - June 4, 2004. McKeown, K., R. Barzilay, D. Evans, V.

Lita, Lucian Vlad, Monica Rogati and Alon Lavie. 2005. BLANC: Learning Evaluation Metrics for MT, Proceedings of Human Language Technology Conference and Conference on Empirical Methods in Natural Language Processing (HLT/EMNLP), pages 740–747, Vancouver, October 2005. Association for Computational Linguistics.

Liu, Chang and Hwee Tou Ng. 2012. Character-Level Machine Translation Evaluation for Languages with AmbiguousWord Boundaries, Proceedings of the 50th Annual Meeting of the Association for Computational Linguistics, pages 921–929, Jeju, Republic of Korea, 8-14 July 2012.

Liu, Chang, Daniel Dahlmeier, and Hwee Tou Ng. 2010. TESLA: Translation evaluation of sentences with linear-programming-based analysis. In Proceedings of the Joint Fifth Workshop on Statistical Machine Translation and MetricsMATR.

Liu, Chang, Daniel Dahlmeier, and Hwee Tou Ng. 2011. Better evaluation metrics lead to better machine translation. InProceedings of EMNLP.

Liu, Ding and Daniel Gildea. 2005. Syntactic features for evaluation of machine translation. In Proceedings of the ACL Workshop on Intrinsic and Extrinsic Evaluation Measures for Machine Translation and/or Summarization, pages 25-32.

Liu, Ding and Daniel Gildea. 2006. Stochastic iterative alignment for machine translation evaluation. Procedding of ACL-06. Sydney.

Liu, Ding and Daniel Gildea. 2007. Maximum Correlation Training for Machine Translation Evaluation, 2007 NAACL.

Lo, Chi-kiu and Dekai Wu. 2011a. MEANT: An Inexpensive, High- Accuracy, Semi-Automatic Metric for Evaluating Translation Utility based on Semantic Roles. Proceedings of the 49th Annual Meeting of the Association for Computational Linguistics, pages 220–229, Portland, Oregon, June 19-24.

Lo, Chi-kiu and Dekai Wu. 2011b. SMT vs. AI redux: How semantic frames evaluate MT more accurately. In Proceedings of the 22nd International Joint Conference on Artificial Intelligence (IJCAI-11).

Lo, Chi-kiu and Dekai Wu. 2011c. Structured vs. Flat Semantic Role Representations for Machine Translation Evaluation. In Proceedings of the 5th Workshop on Syntax and Structure in Statistical Translation (SSST-5).

Lo, Chi-kiu, Anand Karthik Tumuluru and Dekai Wu. 2012. Fully Automatic Semantic MT Evaluation. Proceedings of the 7th Workshop on Statistical Machine Translation, pages 243–252, Montreal, Canada, June 7-8. Association for Computational Linguistics.

Ma, Xiaoyi. 2006. Multiple-Translation Chinese (MTC)part 4. Linguistic Data Consortium.

Mariani, Joseph. 2011. Report of the Session 5 of Tralogy: Quality in Translation, Tralogy [En ligne], Session 5 - Quality in Translation / La qualité en traduction, mis à jour le : 17/10/2011,URL: http://lodel.irevues.inist.fr/tralogy/index.php?id=160&format=print.

Mariño，José B., Rafael E. Banchs, Josep M. Crego, Adrià de Gispert, Patrik Lambert, José A. Fonollosa, Marta R. Costa-jussà. 2006. N-gram based machine translation. Computational Linguistics, Vol. 32, No. 4. pp. 527-549, MIT Press.

Marsh, Elaine,and Dennis Perzanowski.1998. MUC-7 Evaluation of IE Technology: Overview of Results. Proceedings of Message Understanding Conference (MUC-7), 29th April 1998.

Matusov, E., Leusch, G., Banchs, R.E., Bertoldi, N., Dechelotte, D., Federico, M., Kolss, M., Young-Suk Lee, Marino, J.B., Paulik, M., Roukos, S., Schwenk, H., Ney, H. 2008. System Combination for Machine Translation of Spoken and Written Language, Audio, Speech, and Language Processing, IEEE Transactions on, Volume: 16, Issue: 7, 2008, Page(s): 1222-1237.

McKeown, Kathleen R. 1979. Paraphrasing using given andnew information in a question-answer system. In Proc. ofthe ACL, pages 67–72.

Mehdad, Yashar, Matteo Negri and Marcello Federico. 2012. Match without a Referee: Evaluating MT Adequacy without Reference Translations, Proceedings of the 7th Workshop on Statistical Machine Translation, pages 171–180, Montreal, Canada, June 7-8, 2012. Association for Computational Linguistic.

Melamed, I. Dan, Ryan Green and Joseph P. Turian. 2003. Precision and Recall of Machine Translation. In proceedings of HLT-2003.

Menezes, Arul, Kristina Toutanova and Chris Quirk. 2006. Microsoft Research Treelet Translation System: NAACL 2006 Europarl Evaluation, Proceedings of the ACLWorkshop on Statistical Machine Translation, pages 158-161, New York City, June.

Meteer, Marie and Varda Shaked. 1988. Strategies for effectiveparaphrasing. In Proc. of COLING, pages 431–436.

Miller, G. A., R. Beckwith, C. Fellbaum, D. Gross, and K. J.Miller. 1990. WordNet: an on-line lexical database.International Journal of Lexicography, 3(4).pp. 235–244.

Miller, K. J. and Vanni, M. 2005. Inter-rater agreement measures, and the refinement of metrics in the PLATO MT evaluation paradigm. In Proceedings of the Tenth Machine Translation Summit (MT Summit X), Phuket, Thailand.

Mirkin, Shachar, Lucia Specia, Nicola Cancedda, Ido Dagan, Marc Dymetman, Idan Szpektor. 2009. Source-Language Entailment Modeling for Translating Unknown Terms, Proceedings of the Joint Conference of the 47th Annual Meeting of the ACL and the 4th International Joint Conference on Natural Language Processing of the AFNLP, pages 791–799, Suntec, Singapore, 2-7 August 2009. ACL and AFNLP.

Mladenić, Dunja and Marko Grobelnik. 1998. Featureselection for classification based on text hierarchy. InConference on Automated Learning and Discovery(CONALD-98).

Montgomery, D.C. and Runger, G.C. 1994. Applied statistics and probability for engineers, page. John Wiley & Sons New York.

Montgomery, D.C. and Runger, G.C. 2003. Applied statistics and probability for engineers(third edition), page. John Wiley & Sons New York.

Moran，John and David Lewis. 2012. Unobtrusive methods for low-cost manual assessment of machine translation. Tralogy [Online], Session 5 - Quality in Translation / quality in translation, updated on: 16/07/2012, URL: http://lodel.irevues.inist.fr/tralogy/index.php?id=141

Munkres, J. 1957. Algorithms for the Assignment and Transportation Problems. Journal of the Society for Industrial and Applied Mathematics, 5:32–38.

Mutton, Andrew, Mark Dras, Stephen Wan and Robert Dale. 2007. GLEU: Automatic Evaluation of Sentence-Level Fluency, Proceedings of the 45th Annual Meeting of the Association of Computational Linguistics, pages 344–351, Prague, Czech Republic, June 2007. 2007 Association for Computational Linguistics.

Nagao, M. 1984. A framework of a mechanical translation between English and Japaneseby analogy principle. In Elithorn, A., & Banerji, R. (Eds.), Artificialand HumanIntelligence, pp. 173-180. North-Holland.

Nakov, P. and H. T. Ng. 2012. Improving Statistical Machine Translation for a Resource-Poor Language Using Related Resource-Rich Languages, Journal of Artificial Intelligence Research Vol.44 (2012) 179-222.

Naskar, Sudip Kumar, Antonio Toral, Federico Gaspari and Andy Way. 2011. Framework for Diagnostic Evaluation of MT Based on Linguistic Checkpoints, in MT Summit 2011.

Nießen, Sonja, Franz Josef Och, Gregor Leusch, and Hermann Ney. 2000. A Evaluation Tool for Machine Translation: Fast Evaluation for MT Research. In Proceedings of the 2nd International Conference on Language Resources and Evaluation (LREC-2000).

NIST. 2002. The NIST 2002 Machine TranslationEvaluation Plan (MT-02),NIST Open MT Evaluation workshop, Santa Monica, 22-23 July 2002.

NIST. 2009. The 2009 NIST Open Machine TranslationEvaluation Plan (MT09),NIST Open MTEvaluation workshop, Ottawa, ON,Canada, 2009.

Och, F. J. 2003. Minimum Error Rate Training for Statistical Machine Translation. In Proceedings of the 41st Annual Meeting of the Association for Computational Linguistics (ACL-2003). pp. 160-167.

Och, Franz J. and Hermann Ney. 2003. A systematic comparison of various statistical alignment models. J. Computational Linguistics, 29(1):19–51.

Olive, Joseph, Caitlin Christianson, John McCary. 2011. Handbook of Natural LanguageProcessing and Machine Translation, DOI 10.1007/978-1-4419-7713-7_1, Springer SciencetBusiness Media, LLC 2011. (Chapter 5: Machine Translation Evaluation and Optimization).

Owczarzak, Karolina, Declan Groves, Josef Van Genabith and Andy Way. 2006. Contextual Bitext-Derived Paraphrases in Automatic MT Evaluation, Proceedings of the ACLWorkshop on Statistical Machine Translation, pages 86-93,New York City, June 2006.

Owczarzak, Karolina, Josef van Genabith and Andy Way. 2007. Labelled Dependencies in Machine Translation Evaluation, Proceedings of the ACL Second Workshop on Statistical Machine Translation, pages 104-111, Prague, June 2007.

Pado, Sebastian, Michel Galley, Dan Jurafsky and Christopher D. Manning. 2009b. Robust Machine Translation Evaluation with Entailment Features. Proceedings of the 47th Annual Meeting of the ACL and the 4th IJCNLP of the AFNLP, pages 297–305,Suntec, Singapore, 2-7 August 2009.

Pado, Sebastian, Michel Galley, Dan Jurafsky, and Christopher D. Manning. 2009a. Textual entailment features for machine translation evaluation. In Proceedings of the Fourth Workshop on Statistical Machine Translation (StatMT '09). Association for Computational Linguistics, Stroudsburg, PA, USA, 37-41.

Papineni, Kishore, Salim Roukos, Todd Ward, and Wei-Jing Zhu. 2002. BLEU: A Method for Automatic Evaluation of Machine Translation. In Proceedings of the 40th Annual Meeting of the Association for Computational Linguistics (ACL-02), pages 311–318, 2002.

Parton, K., Tetreault, J., Madnani, N. and Chodorow, M. 2011. E-rating Machine Translation. In Proceedings of the Sixth Workshop on Statistical Machine translation of the Association for Computational Linguistics(ACL-WMT), pages 108-115, Edinburgh, Scotland, UK.

Paul, M. 2009. Overview of the IWSLT 2009 Evaluation Campaign. InProc. of IWSLT, Tokyo, Japan, 2009, pp. 1–18.

Paul, M., E. Sumita, L. Bentivogli, M. Federico. 2012. Crowd-based MT Evaluation for non-English Target Languages, Proceedings of the 16th EAMT Conference, 229-237,28-30 May 2012, Trento, Italy.

Paul, Michael, Marcello Federico and Sebastian Stiiker. 2010. Overview of the IWSLT 2010 Evaluation Campaign, Proceedings of the 7th International Workshop on Spoken Language Translation, Paris, December 2nd and 3rd, 2010, page 1-25.

Paul, Michael. 2008. Overview of the IWSLT 2008 Evaluation Campaign, 2008, IWSLT, Proceeding of IWLST 2008, Hawaii, USA.

Pearson, Karl. 1900. On the criterion that a given systemof deviations from the probable in the case of a correlatedsystem of variables is such that it can be reasonablysupposed to have arisen from random sampling.Philosophical Magazine, 50(5):157-175.

Peral, Jesus and Antonio Ferrandez. 2003. Translation of Pronominal Anaphora between English and Spanish: Discrepancies and Evaluation, Journal of Artificial Intelligence Research 18 (2003), 117-147.

Petrov, S., Barrett, L., Thibaux, R., and Klein, D. 2006. Learning Accurate, Compact, and Interpretable Tree Annotation. In Proceedings of the 21st International Conference on ComputationalLinguistics and 44th Annual Meeting of the Association for Computational Linguistics, pages 433–440, Sydney, Australia. Association for Computational Linguistics.

Popovic, M. 2011. Morphemes and POS tags for n-gram based evaluation metrics. In Proceedings of the Sixth Workshop on Statistical Machine Translation, Association for Computational Linguistics (ACL-WMT), pages 104-107, Edinburgh, Scotland, UK.

Popovic, M. and Hermann Ney. 2007. Word Error Rates: Decomposition over POS Classes and Applications for Error Analysis, Proceedings of the Second Workshop on Statistical Machine Translation, pages 48–55, Prague, June 2007. Association for Computational Linguistics.

Popovic, M., Vilar, D., Avramidis, E. and Burchardt, A. 2011. Evaluation without references: IBM1 scores as evaluation metrics. In Proceedings of the Sixth Workshop on Statistical Machine Translation, Association for Computational Linguistics (ACL-WMT), pages 99-103, Edinburgh, Scotland, UK.

Popovic, Maja. 2012. Class error rates for evaluation of machine translation output. Proceedings of the 7th Workshop on Statistical Machine Translation, pages 71–75, Canada, June 7-8. Association for Computational Linguistics.

Porter, M.F. 1980. An algorithm for suffic stripping. Program, 14(3):130–137.

Povlsen, Claus, Nancy Underwood, Bradley Music, and Anne Neville. 1998. Evaluating Text-Type Suitability for Machine Translation a Case Study on an English-Danish System.Proceedings of the First Language Resources and Evaluation Conference, LREC-98, Volume I. 27-31. Granada, Spain.

Przybocki, Mark, Kay Peterson, and S´ebastien Bronsart.2009. NIST 2008 metrics for machine translation(MetricsMATR08) development data. Linguistic DataConsortium.

Przybocki, Mark, Kay Peterson, Sebastien Bronsart. 2008. Translation Adequacy and Preference Evaluation Tool (TAP-ET), Proceedings of the 6th International Conference on Language Resources and Evaluation (LREC 2008)

Quirk, C.B. 2004. Training a Sentence-Level MachineTranslation Confidence Measure. In Proceedings ofLREC 2004.

Raybaud, Sylvain, David Langlois, and Kamel Smaili. 2011. ”this sentence is wrong.” detecting errors inmachine-translated sentences. J. Machine Translation,25(1):1–34, March.

Reeder, F. 2004. Investigation of intelligibility judgments. In Proceedings of the Conference of the Association for Machine Translation in the Americas (AMTA 2004), LNAI 3265, pp. 227–235.

Reeder, F. 2006a. Direct application of a language learner test to MT evaluation. Proceedings of the 7th Conference of the Association for Machine Translation in the Americas, pages 166-175, Cambridge, August 2006.

Reeder, F. 2006b. Measuring MT adequacy using latent semantic analysis. In Proceedings of the 7th Conference of the Association for Machine Translation in the Americas, pages 176-184, Cambridge, August 2006.

Riezler, Stefan and John T. Maxwell III. 2005. On Some Pitfalls in Automatic Evaluation and Significance Testing for MT. Proceedings of the ACL Workshop on Intrinsic and Extrinsic Evaluation Measures for Machine Translationand/or Summarization, pages 57–64, Ann Arbor, June 2005.

Roturier, Johann and Anthony Bensadoun. 2011. Evaluation of MT Systems to Translate User Generated Content, MT Summit XIII.

Sager, J. 1978. Criteria for Machine Translation Evaluation. Proceedings of Workshop on Evaluation Problems in Machine Translation. Luxembourg. February, 1978.

Salton, G. and M.E. Lesk. 1968. Computer evaluation of indexing and text processing. Journal of the ACM, 15(1) , 8-36.

Sanchez-Martınez, F. and Forcada, M. L. 2009.Inferring shallow-transfer machine translation rules from small parallel corpora. Journal of Artificial Intelligence Research , 34:605–635.

Sasaki, Y. and R. Fellow. 2007. The truth of the F-measure. MIB-School of Computer Science, University of Manchester, pp. 1-5, 2007.

Siegel, Sidney and N.John Castellan.1988. Non Parametric Statistics for the Behavioral Sciences.McGraw-Hill, New York.

Snover, Matthew G., Nitin Madnani, Bonnie Dorr, and Richard Schwartz. 2009. TER-Plus: paraphrase, semantic, and alignment enhancements to Translation Edit Rate. J. Machine Tranlslation (2009a), 23: 117-127.

Snover, Matthew G., Nitin Madnani, Bonnie Dorr, and Richard Schwartz. 2009. Fluency, Adequacy, or HTER? Exploring Different Human Judgments with a Tunable MT Metric. WMT2009b, Proceedings of the 4th EACL Workshop on Statistical Machine Translation, pages 259–268,Athens, Greece, 30 March – 31 March 2009.

Snover, Matthew, Bonnie J. Dorr, Richard Schwartz, Linnea Micciulla, and John Makhoul. 2006. A Study of Translation Edit Rate with Targeted Human Annotation. In Proceedings of the 7th Conference of the Association for Machine Translation in the Americas (AMTA-06), pages 223–231, 2006.

Song, X. and Cohn, T. 2011. Regression and ranking based optimisation for sentence level MT evaluation. In Proceedings of the (ACL-WMT), pages 123-129, Edinburgh, Scotland, UK.

Specia, L. and Gimenez, J. 2010. Combining Confidence Estimation andReference-based Metrics for Segment-level MT Evaluation. The Ninth Conferenceof the Association for Machine Translation in the Americas (AMTA 2010).

Specia, Lucia, Dhwaj Raj, Marco Turchi. 2010. Machine translation evaluation versus quality estimation, J. Machine Translation (2010), 24:39-50. DOI 10.1007/s10590-010-9077-2

Specia, Lucia, Najeh Hajlaoui, Catalina Hallett and Wilker Aziz. 2011. Predicting Machine Translation Adequacy, MT Summit XIII.

Specia, Lucia. 2011. Exploiting Objective Annotations for Measuring Translation Post-editing Effort. In Proceedings of the 15th Conference of the European Association for Machine Translation, pages 73–80, Leuven.

Stockle, A. 2002. SRILM - An Extensible LanguageModeling Toolkit, In Proceedings of InternationalConference on Spoken Language Processing

Su, Hung-Yu and Chung-Hsien Wu. 2009. Improving Structural Statistical Machine Translationfor Sign Language With Small Corpus UsingThematic Role Templates as Translation Memory, IEEE TRANSACTIONS ON AUDIO, SPEECH, AND LANGUAGE PROCESSING, VOL. 17, NO. 7, SEPTEMBER 2009.

Su, Hung-Yu, Chung-Hsien Wu. 2009. Improving Structural Statistical Machine Translation for Sign Language With Small Corpus Using Thematic Role Templates as Translation Memory, Audio, Speech, and Language Processing, IEEE Transactions on, Volume:17, Issue: 7, 2009, Page(s): 1305- 1315.

Su, Keh-Yih, Wu Ming-Wen and Chang Jing-Shin. 1992. A New Quantitative Quality Measure for Machine Translation Systems. In Proceedings of the 14th International Conference on Computational Linguistics, pages 433–439, Nantes, France, July 1992.

Suzuki, Hirokazu. 2011. Automatic post-editing based on SMT and its selective application by sentence-level automatic quality evaluation. MT Summit XIII.

Tillmann, Christoph, Stephan Vogel, Hermann Ney, Arkaitz Zubiaga, and Hassan Sawaf. 1997. Accelerated DP Based Search For Statistical Translation. In Proceedings of the 5th European Conference on Speech Communication and Technology (EUROSPEECH-97), 1997.

Toral, Antonio; Sudip Kumar Naskar; Joris Vreeke; Federico Gaspari; Declan Groves. 2013. A Web Application for the Diagnostic Evaluation of Machine Translation over Specific Linguistic Phenomena. Proceedings of the NAACL HLT 2013 Demonstration Session, pages 20-23,Atlanta, Georgia, 10-12 June 2013. Association for Computational Linguistics.

Turian, Joseph P., Luke Shen, and I. Dan Melamed. 2003. Evaluation of Machine Translation and its Evaluation. In Machine Translation Summit IX, pages 386–393. International Association for Machine Translation, September 2003.

Van Rijsbergen, C. J. 1979. Information Retrieval (2nd ed.). Butterworth-Heinemann, Newton, MA, USA.

Vanni, Michelle and Florence Reeder. 2000. How Are You Doing? A Look at MT Evaluation. InProceedings of the 4th Conference of the Association for Machine Translation in the Americas on Envisioning Machine Translation in the Information Future (AMTA '00), John S. White (Ed.). Springer-Verlag, London, UK, 109-116.

Vilar, David, Gregor Leusch, Hermann Ney and Rafael E. Banchs. 2007. Human Evaluation of Machine Translation Through Binary System Comparisons, Proceedings of the ACL Second Workshop on Statistical Machine Translation, pages 96–103, Prague, June 2007.

Vossen, P., G. Rigau, E. Agirre, A. Soroa, M.Monachini, and R. Bartolini. 2010. KYOTO: anopen platform for mining facts. COLING-2010:Proceedings of the 6th international workshop onOntologies and Lexical Resources (Ontolex 2010),Beijing, China, pp. 1-10.

Voutilainen, A., Heikkilä, J. & Anttila, A. 1992.Constraint Grammar of English. A Performance-Oriented Introduction. University of Helsinki:Department of General Linguistics.

Wang, Mengqiu and Christopher D. Manning. 2012a. SPEDE: Probabilistic Edit Distance Metrics for MT Evaluation, WMT2012, 76-83.

Wang, Mengqiu and Christopher D. Manning.2012b. Probabilistic Finite State Machines for Regression-based MT Evaluation. EMNLP 2012.

Weaver, Warren. 1955. Translation. In William Locke and A. Donald Booth, editors, Machine Translation of Languages: Fourteen Essays. John Wiley & Sons, New York, pages 15–23.

White, J. 1995. Approaches to Black Box MT Evaluation. Proceedings of MT Summit V.

White, J. 2000. Toward an Automated, Task-Based MT Evaluation Strategy. In Maegaard, B., ed., Proceedings of the Workshop on Machine Translation Evaluation at LREC-2000. Athens, Greece.

White, J. S., O'Connell, T. A., and O'Mara, F. E. 1994. The ARPA MT evaluation methodologies: Evolution, lessons, and future approaches. In Proceedings of the Conference of the Association for Machine Translation in the Americas (AMTA 1994). pp193-205.

White, John S. and Kathryn B. Taylor. 1998. A Task-Oriented Evaluation Metric for MachineTranslation. First International Conference on Language Resources & Evaluation, Granada, Spain, 28-30 May 1998.

Wong, B. T-M and Kit, C. 2008. Word choice and word position for automatic MT evaluation. In Workshop: MetricsMATR of the Association for Machine Translation in the Americas (AMTA), short paper, 3 pages, Waikiki, Hawai'I, USA.

Wong, Billy and Chunyu Kit. 2009. ATEC: automatic evaluation of machine translation via word choice and word order. Mach Translat (2009) 23:141–155.

Wong, Billy and Chunyu Kit.2010. The Parameter-optimized ATEC Metric for MT Evaluation. WMT2010. Proceedings of the Joint 5th Workshop on Statistical Machine Translation and MetricsMATR, pages 360–364,Uppsala, Sweden, 11-16 July 2010. Association for Computational Linguistics

Wong, Billy and Chunyu Kit. 2011. Comparative evaluation of term informativeness measures for machine translation evaluation metrics, in MT Summit 2011, pp.537-544. Sep 19-23, 2011, Xiamen, China.

Wong, Billy T. M. and Chunyu Kit. 2012. Extending Machine Translation Evaluation Metrics with Lexical Cohesion to Document Level. Proceedings of the 2012 Joint Conference on Empirical Methods in Natural Language Processing and Computational NaturalLanguage Learning, pages 1060–1068, Jeju Island, Korea, 12–14 July 2012. c 2012 Association for Computational Linguistics.

Wu, Zhibiao and Martha Palmer. 1994. Verb semanticsand lexical selection. In ACL 1994, pages 133–138,Las Cruces.

Xiong, Deyi, Min Zhang, Haizhou Li. 2011. A Maximum-Entropy Segmentation Model for Statistical Machine Translation, Audio, Speech, and Language Processing, IEEE Transactions on, Volume:19, Issue: 8, 2011 , Page(s): 2494- 2505.

Ye, Yang, Ming Zhou and Chin-Yew Lin. 2007. Sentence Level Machine Translation Evaluation as a Ranking Problem: one step aside from BLEU, Proceedings of the ACL Second Workshop on Statistical Machine Translation, pages 240-247, Prague.

Zhou, L., C.Y. Lin and E. Hovy. 2006. Re-evaluating Machine Translation Results with Paraphrase Support. Proceedings of the 2006 Conference on Empirical Methods in Natural Language Processing (EMNLP 2006), pages 77–84.

Zhou, Liang, Chin-Yew Lin, and Eduard Hovy. 2006. Re-evaluating machine translation results with paraphrase support. In Proceedings of the 2006 Conference on Empirical Methods in Natural Language Processing (EMNLP '06). Association for Computational Linguistics, Stroudsburg, PA, USA, 77-84.

Zhou, M., Wang, B., Liu, S., Li, M., Zhang, D., and Zhao, T. 2008. Diagnostic evaluationof machine translation systems using automatically constructed linguistic check-points. InProceedings of COLING 2008, pages 1121–1128.

# APPENDIX A: Proposed Universal Phrase Tagset and Mappings

## Phrase Tagset Mapping between Universal Tagset and Existing Treebanks (part1)

| Universal Phrase Tag | English PennTreebank I (Marcus et al., 1993) | English PennTreebank II (Bies et al., 1995) | Chinese PennTreebank (Xue and Jiang, 2010) | Portuguese Floresta Treebank (Afonso et al., 2002) | FrenchTreebank (Abeille, 2003) | Japanese Treebank Tüba-J/S (Kawata and Bartels, 2000) |
|---|---|---|---|---|---|---|
| NP | NP, WHNP | NP, NAC, NX, QP, WHNP | NP, CLP, QP, LCP, WHNP | np | NP | NPper, NPloc, NPtmp, NP, NP.foc |
| VP | VP | VP | VP, VCD, VCP, VNV, VPT, VRD, VSB | vp | VN, VP, VPpart, VPinf | VP.foc, VP, VPcnd, VPfin |
| AJP | ADJP | ADJP, WHADJP | ADJP, DP, DNP | ap, adjp | AP | AP.foc, AP, APcnd |
| AVP | ADVP, WHADVP | ADVP, WHAVP, PRT, WHADVP | ADVP, DVP | advp | AdP | ADVP.foc, ADVP |
| PP | PP, WHPP | PP, WHPP | PP | pp | PP | PP, PP.foc, PPnom, PPgen, PPacc |
| S | S, SBAR, SBARQ, SINV, SQ | S, SBAR, SBARQ, SINV, SQ, PRN, FRAG, RRC | IP, CP, PRN, FRAG, INC | fcl, icl, acl, cu, x, sq | SENT, Ssub, Sint, Srel, S | S, SS |
| CONJP | | CONJP | | | | |
| COP | | UCP | UCP | | COORD | |
| X | X | X, INTJ, LST | LST, FLR, DFL, INTJ, URL, X | | | ITJ, GR, err |
| Universal Phrase Tag | Danish Arboretum Treebank | German NegraTreebank (Skut et al., 1997) | Spanish UAM Treebank (Moreno et al., 1999) | Hungarian Szeged Treebank | Spanish Treebank (Volk, 2009) | Swedish Talbanken05 (Nivre et al., 2006) |
| NP | np | NP, CNP, MPN, NM | HOUR, NP, QP, SCORE, TITLE | NP, QP | NP, MPN | CNP, NP |
| VP | vp, acl | VP, CVP, VZ, CVZ | VP | VP, INF_, INF0 | SVC | CVP, VP |
| AJP | ajp | AP, CAP, MTA | ADJP | ADJP | AP | AP, CAP, |
| AVP | dvp | AVP, CAVP, AA | ADVP, PRED-COMPL | ADVP, PA_, PA0 | AVP | AVP, CAVP, |
| PP | pp | PP, CAC, CPP, CCP | PP | PP | PP, MTP | CPP, PP |
| S | fcl, icl | S, CS, CH, DL, PSEUDO | CL, S | S | S, INC | CS, S |
| CONJP | cp | | | C0 | MTC | |
| COP | | CO | | CP | CS, CNP, CPP, CAP, CAVP, CAC, CCP, CO | CONJP, CXP |
| X | par | ISU, QL | | FP, XP | | NAC, XP |

Phrase Tagset Mapping between Universal Tagset and Existing Treebanks (part2)

| Universal Phrase Tag | Arabic PENN Treebank (Bies and Maamouri, 2003.) | Korean Penn Treebank (Han et al., 2001; 2002) | Estonian Arborest Treebank | Icelandic IcePaHC Treebank (Wallenberg et al., 2011) | Italian ISST Treebank (Montemagni et al., 2000; 2003) | Portuguese Tycho Brahe Treebank (Galves & Faria, 2010) |
|---|---|---|---|---|---|---|
| NP | NP, NX, QP, WHNP | NP | AN>, <AN, NN>, <NN, | NP, QP, WNP | SN | NP, NP-ACC, NP-DAT, NP-GEN, NP-SBJ, IP-SMC, NP-LFD, NP-ADV, NP-VOC, NP-PRN |
| VP | VP | VP | VN>, <VN, INF_N>, <INF_N | VP | IBAR | VB, VB-P |
| AJP | ADJP, WHADJP | ADJP, DANP | | ADJP, ADJP-SPR | SA | ADJP, ADJP-SPR |
| AVP | ADVP, WHADVP | ADVP, ADCP | AD>, <AD | ADVP, ADVP-DIR, ADVP-LOC, ADVP-TMP, RP | SAVV | ADVP, WADVP |
| PP | PP, WHPP | | | PP, WPP, PP-BY, PP-PRN | SP, SPD, SPDA | PP, PP-ACC, PP-SBJ, PP-LFD, PP-PRN, PP-LOC |
| S | S, SBAR, SBARQ, SQ | S | | | F, SV2, SV3, SV5, FAC, FS, FINT, F2 | RRC, CP, CP-REL, IP-MAT, IP-INF, IP-SUB, CP-ADV, CP-THT |
| CONJP | CONJP, NAC | | | CONJP | CP, COMPC | CONJP |
| COP | UCP | | PN>, <PN | | FC, COORD | |
| X | PRN, PRT, FRAG, INTJ, X | INTJ, PRN, X, LST, XP | <P, P>, <Q, Q> | LATIN | FP, COMPT, COMPIN | |
| | | | | | | |
| Universal Phrase Tag | Hindi-Urdu Treebank (Bhatt et al., 2012) | Catalan AnCora Treebank (Civit and Marti, 2004; Taulé et al., 2008) | Swedish Treebank () | Vietnamese Treebank (Nguyen et al., 2009) | Thai CG Treebank (Ruangrajitpakorn et al., 2009) | Hebrew (Sima'an et a., 2001) |
| NP | NP, NP-P, NP-NST, SC-A, SC-P, NP-P-Pred | Sn | NP | NP, WHNP, QP | np, num, spnum | NP-gn-(H) |
| VP | VP, VP-Pred, V' | Gv | VP | VP | | PREDP, VP, VP-MD, VP-INF |
| AJP | AP, AP-Pred | Sa | AP | AP, WHAP | | ADJP-gn-(H) |
| AVP | DegP | sadv, neg | AVP | RP, WHRP | | ADVP |
| PP | | Sp | PP | PP, WHPP | pp | PP |
| S | | S, S*, S.NF.C, S.NF.A, S.NF.P, S.F.C, S.F.AComp, S.F.AConc, S.F.Acons, S.F.Acond, S.F.R, | ROOT, S | S, SQ, SBAR | s, ws, root | FRAG, FRAGQ, S, SBAR, SQ |
| CONJP | | conj.subord, coord | | | | |
| COP | CCP, XP-CC | | | | | |
| X | CP | interjeccio, morfema.verbal, morf.pron | XP | XP, YP, MDP | | INTJ, PRN |

VITA

(Aaron) Li-Feng Han

University of Macau

2014

**Competition Award**

Second Prize in National Post-Graduate Mathematical Contest in Modeling (NPGMCM2011)
2011. National Postgraduate Mathematical Contest In Modeling Committee
The first prizes and second prizes occupy 20.45% of total 2,245 teams from 242 universities and research institutes all over the P.R.C. country, including more than 300 doctoral students.(paper).http://www.shumo.com/home/html/1317.html

**Open-source Toolkits Developed** and **Google-scholar page**

Open source tools:
Homepage: https://github.com/aaronlifenghan (LEPOR, HPPR, EBLEU, n/hLEPOR)

Google scholar citation:
Homepage: http://scholar.google.com/citations?hl=en&user=_vf3E2QAAAAJ

**Selected Publications:**

1. Aaron Li-Feng Han, Derek F. Wong, Lidia S. Chao, Liangye He and Yi Lu. Unsupervised Quality Estimation Model for English to German Translation and Its Application in Extensive Supervised Evaluation. *The Scientific World Journal, Issue: Recent Advances in Information Technology*. Page 1-12, April 2014.Hindawi Publishing Corporation. ISSN:1537-744X. http://www.hindawi.com/journals/tswj/aip/760301/

2. Aaron Li-Feng Han, Derek F. Wong, Lidia S. Chao, Liangye He, Yi Lu, Junwen Xing and Xiaodong Zeng. Language-independent Model for Machine Translation Evaluation with Reinforced Factors. *Proceedings of the 14th International Conference of Machine Translation Summit (MT Summit)*, pp. 215-222. Nice, France. 2 - 6 September 2013. International Association for Machine Translation. http://www.mt-archive.info/10/MTS-2013-Han.pdf

3. Aaron Li-Feng Han, Derek Wong, Lidia S. Chao, Yi Lu, Liangye He, Yiming Wang, Jiaji Zhou. A Description of Tunable Machine Translation Evaluation Systems in WMT13 Metrics Task. *Proceedings of the ACL 2013 EIGHTH WORKSHOP ON STATISTICAL MACHINE TRANSLATION (ACL-WMT)*, pp. 414-421, 8-9 August 2013. Sofia, Bulgaria. Association for Computational Linguistics. http://www.aclweb.org/anthology/W13-2253

4. Aaron Li-Feng Han, Yi Lu, Derek F. Wong, Lidia S. Chao, Liangye He, Junwen Xing. Quality Estimation for Machine Translation Using the Joint Method of Evaluation Criteria and Statistical Modeling. *Proceedings of the ACL 2013 EIGHTH WORKSHOP ON STATISTICAL MACHINE TRANSLATION (ACL-WMT)*, pp. 365-372. 8-9 August 2013. Sofia, Bulgaria. Association for Computational Linguistics. http://www.aclweb.org/anthology/W13-2245

5. Aaron Li-Feng Han, Derek F. Wong, Lidia S. Chao, Liangye He, Shuo Li and Ling Zhu. Phrase Tagset Mapping for French and English Treebanks and Its Application in Machine Translation Evaluation. *Language Processing and Knowledge in the Web. Lecture Notes in Computer Science Volume 8105*, 2013, pp 119-131. Volume Editors: Iryna Gurevych, Chris Biemann and Torsten Zesch. Springer-Verlag Berlin Heidelberg. http://dx.doi.org/10.1007/978-3-642-40722-2_13

6. Aaron Li-Feng Han, Derek F. Wong, Lidia S. Chao, Liangye He, Ling Zhu and Shuo Li. A Study of Chinese Word Segmentation Based on the Characteristics of Chinese. *Language Processing and Knowledge in the Web. Lecture Notes in Computer Science Volume 8105*, 2013, pp 111-118. Volume Editors: Iryna Gurevych, Chris Biemann and Torsten Zesch. Springer-Verlag Berlin Heidelberg. http://dx.doi.org/10.1007/978-3-642-40722-2_12

7. Aaron Li-Feng Han, Derek F. Wong, Lidia S. Chao, Liangye He. Automatic Machine Translation Evaluation with Part-of-Speech Information. *Text, Speech, and Dialogue. Lecture Notes in Computer Science Volume 8082*, 2013, pp 121-128. Volume Editors: I. Habernal and V. Matousek. Springer-Verlag Berlin Heidelberg. http://dx.doi.org/10.1007/978-3-642-40585-3_16

8. Aaron Li-Feng Han, Derek Fai Wong and Lidia Sam Chao. Chinese Named Entity Recognition with Conditional Random Fields in the Light of Chinese Characteristics. *Language Processing and Intelligent Information Systems. Lecture Notes in Computer Science Volume 7912*, 2013, pp 57-68. M.A. Klopotek et al. (Eds.): IIS 2013. Springer-Verlag Berlin Heidelberg. http://dx.doi.org/10.1007/978-3-642-38634-3_8

9. Aaron Li-Feng Han, Derek F. Wong and Lidia S. Chao. LEPOR: A Robust Evaluation Metric for Machine Translation with Augmented Factors. *Proceedings of the 24th International Conference on Computational Linguistics (COLING): Posters*, pages 441–450, Mumbai, December 2012. Association for Computational Linguistics. http://aclweb.org/anthology//C/C12/C12-2044.pdf